\documentclass[10pt,journal,compsoc]{IEEEtran}

\usepackage{booktabs}
\usepackage{amssymb}
\usepackage{amsfonts}
\usepackage{amsmath}
\usepackage{array}
\usepackage{threeparttable}
\usepackage{booktabs}
\usepackage{cite}
\usepackage[dvips]{graphicx}
\usepackage{xcolor}
\usepackage{comment}
\usepackage{makecell}
\usepackage{multirow}
\usepackage{textcomp,mathcomp}
\usepackage[linesnumbered, ruled, vlined]{algorithm2e}
\usepackage{subcaption}
\usepackage{caption}
\usepackage{pifont}
\usepackage[colorlinks=true, linkcolor=red]{hyperref}

\begin{document}
\newtheorem{definition}{\it Definition}
\newtheorem{theorem}{\bf Theorem}
\newtheorem{lemma}{\it Lemma}
\newtheorem{corollary}{\bf Corollary}
\newtheorem{remark}{\it Remark}
\newtheorem{example}{\it Example}
\newtheorem{case}{\bf Case Study}
\newtheorem{assumption}{\it Assumption}
\newtheorem{property}{\it Property}
\newtheorem{proposition}{\bf Proposition}

\newcommand{\hP}[1]{{\boldsymbol h}_{{#1}{\bullet}}}
\newcommand{\hS}[1]{{\boldsymbol h}_{{\bullet}{#1}}}

\newcommand{\ba}{\boldsymbol{a}}
\newcommand{\baq}{\overline{q}}
\newcommand{\bA}{\boldsymbol{A}}
\newcommand{\bb}{\boldsymbol{b}}
\newcommand{\bB}{\boldsymbol{B}}
\newcommand{\bc}{\boldsymbol{c}}
\newcommand{\bd}{\boldsymbol{d}}
\newcommand{\bcD}{\boldsymbol{\cal D}}
\newcommand{\bcO}{\boldsymbol{\cal O}}
\newcommand{\bh}{\boldsymbol{h}}
\newcommand{\bH}{\boldsymbol{H}}
\newcommand{\bl}{\boldsymbol{l}}
\newcommand{\bm}{\boldsymbol{m}}
\newcommand{\bn}{\boldsymbol{n}}
\newcommand{\bo}{\boldsymbol{o}}
\newcommand{\bO}{\boldsymbol{O}}
\newcommand{\bp}{\boldsymbol{p}}
\newcommand{\bq}{\boldsymbol{q}}
\newcommand{\bR}{\boldsymbol{R}}
\newcommand{\bs}{\boldsymbol{s}}
\newcommand{\bS}{\boldsymbol{S}}
\newcommand{\bT}{\boldsymbol{T}}
\newcommand{\bw}{\boldsymbol{w}}
\newcommand{\bz}{\boldsymbol{z}}

\newcommand{\balpha}{\boldsymbol{\alpha}}
\newcommand{\bbeta}{\boldsymbol{\beta}}
\newcommand{\bgamma}{\boldsymbol{\gamma}}
\newcommand{\bomega}{\boldsymbol{\omega}}
\newcommand{\btomega}{\boldsymbol{\tilde \omega}}

\newcommand{\bOmega}{\boldsymbol{\Omega}}

\newcommand{\bTheta}{\boldsymbol{\Theta}}
\newcommand{\bphi}{\boldsymbol{\phi}}
\newcommand{\btheta}{\boldsymbol{\theta}}
\newcommand{\bvarpi}{\boldsymbol{\varpi}}
\newcommand{\bpi}{\boldsymbol{\pi}}
\newcommand{\bpsi}{\boldsymbol{\psi}}
\newcommand{\bxi}{\boldsymbol{\xi}}
\newcommand{\bx}{\boldsymbol{x}}
\newcommand{\by}{\boldsymbol{y}}

\newcommand{\cA}{{\cal A}}
\newcommand{\bcA}{\boldsymbol{\cal A}}
\newcommand{\cB}{{\cal B}}
\newcommand{\cD}{{\cal D}}
\newcommand{\cE}{{\cal E}}
\newcommand{\cG}{{\cal G}}
\newcommand{\cH}{{\cal H}}
\newcommand{\cI}{{\cal I}}
\newcommand{\bcH}{\boldsymbol {\cal H}}
\newcommand{\cJ}{{\cal J}}
\newcommand{\cK}{{\cal K}}
\newcommand{\cL}{{\cal L}}
\newcommand{\cM}{{\cal M}}
\newcommand{\cO}{{\cal O}}
\newcommand{\cR}{{\cal R}}
\newcommand{\cS}{{\cal S}}
\newcommand{\dcS}{\ddot{{\cal S}}}
\newcommand{\ds}{\ddot{{s}}}
\newcommand{\cT}{{\cal T}}
\newcommand{\cU}{{\cal U}}
\newcommand{\wt}[1]{\widetilde{#1}}

\newcommand{\bfW}{\mathbf{W}}
\newcommand{\bfw}{\mathbf{w}}

\newcommand{\mA}{\mathbb{A}}
\newcommand{\mE}{\mathbb{E}}
\newcommand{\mG}{\mathbb{G}}
\newcommand{\mR}{\mathbb{R}}
\newcommand{\mS}{\mathbb{S}}
\newcommand{\mU}{\mathbb{U}}
\newcommand{\mV}{\mathbb{V}}
\newcommand{\mW}{\mathbb{W}}

\newcommand{\uq}{\underline{q}}
\newcommand{\ubq}{\underline{\boldsymbol q}}

\newcommand{\red}[1]{\textcolor[rgb]{1,0,0}{#1}}
\newcommand{\gre}[1]{\textcolor[rgb]{0,1,0}{#1}}
\newcommand{\blu}[1]{\textcolor[rgb]{0,0,0}{#1}}

\newcommand{\ltgr}[1]{\textcolor[rgb]{0.6,0.6,0.6}{#1}}

\newcommand{\best}{$\uparrow$}
\newcommand{\worst}{$\downarrow$}

\title{SANet: A Semantic-aware Agentic AI Networking Framework for Cross-layer Optimization in 6G}

\author{Yong~Xiao, \IEEEmembership{Senior~Member, IEEE}, Xubo Li, Haoran Zhou, Yingyu Li, Yayu Gao, Guangming~Shi, \IEEEmembership{Fellow, IEEE}, Ping Zhang, \IEEEmembership{Fellow, IEEE}, and Marwan Krunz, \IEEEmembership{Fellow, IEEE} 

\thanks{*This work is accepted at IEEE Transactions on Mobile Computing. Copyright may be transferred without notice, after which this version may no longer be accessible.

This work was supported in part by the National Natural Science Foundation of China (NSFC) under grants 62525109 and 62571208, the Mobile Information Network National Science and Technology Key Project under grant 2024ZD1300700, and Hubei Natural Science Foundation Innovation Research Group Program under grant 2026AFA044. 
An earlier version of this paper is presented in part at the IEEE GLOBECOM Conference, Taipei, Taiwan, December 2025\cite{XY2025SANNet}. \blu{Code is available at \texttt{https://github.com/WirelessAIatHUST/SANet} (Corresponding author: Yingyu Li.)}

Yong~Xiao, Xubo~Li, Haoran~Zhou, and Yayu~Gao are with the School of Electronic Information and Communications, the Huazhong University of Science and Technology, Wuhan, China 430074. Yong~Xiao is also with the Peng Cheng Laboratory, Shenzhen, China, and Pazhou Laboratory (Huangpu), Guangzhou, China (e-mail: \{yongxiao, yayugao\}@hust.edu.cn). 

Y. Li is with the School of Mechanical Engineering and Electronic Information, China University of Geosciences (Wuhan), China, 430074 (e-mail: liyingyu29@cug.edu.cn).

G. Shi is with the Peng Cheng Laboratory, Shenzhen, China 518055, also with the School of Artificial Intelligence, Xidian University, Xi'an, Shaanxi 710071, China (e-mail: gmshi@xidian.edu.cn). 
(e-mail: gmshi@xidian.edu.cn).

P. Zhang is with the State Key Laboratory of Networking and Switching Technology, Beijing University of Posts and Telecommunications, Beijing, China 100876 (email: pzhang@bupt.edu.cn).


M. Krunz is with the Department of Electrical and Computer Engineering, the University of Arizona, Tucson, AZ 85721 (e-mail: krunz@arizona.edu).
}
}

\maketitle

\begin{abstract}
Agentic AI networking (AgentNet) is a novel AI-native networking paradigm in which a large number of specialized AI agents collaborate to perform autonomous decisions, dynamic environmental adaptation, and complex missions. AgentNet has the potential to facilitate real-time network management and optimization functions, including self-configuration, self-optimization, and self-adaptation across diverse and complex environments, laying the foundation for fully autonomous networking systems. Despite its promise, AgentNet is still in the early stages of development and still lacks an effective networking framework to support automatic goal discovery, multi-agent self-orchestration, and task assignment. This paper proposes SANet, a novel semantic-aware AgentNet architecture for wireless networks. SANet can infer the semantic goal of the user and automatically assign agents associated with different layers of the network stack to fulfill the inferred goal. Motivated by the fact that AgentNet is a decentralized framework in which collaborating agents may generally have different and even conflicting objectives, we formulate the decentralized optimization of SANet as a multi-agent multi-objective problem, and focus on finding the Pareto-optimal solution for agents with distinct and potentially conflicting objectives. We propose three novel metrics for evaluating SANet: (the agents' objective) optimization error, (dynamic environment) generalization error, and (multi-objective) conflicting error. Furthermore, we develop a model partition and sharing (MoPS) framework in which large models, e.g., deep learning models, of different agents can be partitioned into shared and agent-specific parts that are jointly constructed and deployed according to agents' local computational resources. Two decentralized optimization algorithms, static-weighting and dynamic-weighting algorithms, are introduced to optimize the above three metrics. A bandwidth-adaptive compression framework is also proposed to enable different agents to perform in situ compression of their intermediate embeddings, dynamically adjusting to localized resource constraints and task requirements. We derive theoretical bounds for all these performance metrics and prove that there exists a three-way tradeoff among optimization, generalization, and conflicting errors. Finally, to validate our theoretical results, we develop an open-source Radio Access Network (RAN) and core network-based hardware prototype that implements three Transformer-based time-series prediction agents to interact with three different layers of the network. Experimental results show that the proposed MoPS framework achieves performance gains of up to $14.61\%$ while requiring only $44.37\%$ of the Floating-Point Operations (FLOPs) for inference at each agent compared to state-of-the-art algorithms. Also, compared to the static-weighting algorithm, the dynamic-weighting algorithm achieves up to $83.81\%$ reduction in training errors caused by conflicting objectives. 

\end{abstract}

\section{Introduction}
\label{Section_Introduction}

Wireless systems have long utilized a centralized and layered architecture, primarily due to its inherent structural simplicity and ease of management. 
Emerging NextG applications and services, such as mixed reality (XR)-based immersive communication\cite{Nguyen2024MetaverseBrainComputer,XY2018TactileInternet}, semantic communication\cite{shi2020semantic, Deniz2024JSCCProcIEEE}, and ambient intelligence\cite{Albert2020AmbientIntelligence}, create high computation and communication demands on 
the network edge, rendering the existing centralized architecture a major performance and scalability bottleneck. Specifically, the existing layered design requires that any tactical decision made at the edge must be monitored and approved by higher-layer entities, which causes intolerably high communication overhead and latency. Furthermore, the recent trend toward deep integration of AI and communication networking has exacerbated these challenges, as the architecture must now support the generation, transfer, and processing of large volumes of data across layers. Recent works focused on cross-layer optimization, particularly those leveraging distributed optimization techniques, and proposed useful theoretical solutions to mitigate these issues; however, these works often require predefined and carefully designed multi-layer decomposition schemes and coordinated objectives with limited flexibility and scalability. 

This has led to the emergence of agentic AI networking ({\it AgentNet}), a flexible, goal-driven networking framework that focuses on the autonomous interaction and self-adaptation of a diverse group of AI agents to achieve different goals. 
Unlike existing AI-based network optimization solutions that are primarily built on closed-loop and passive learning frameworks and are based on a fixed training dataset, agents in AgentNet can actively seek new information, explore possibilities, and leverage real-world knowledge and experience to achieve a wide range of goals, thereby minimizing the influence of potential bias within the initial training dataset\cite{xiao2025AgentNet, XY2020Selflearning, Yang2022NetMagazine, Morris2024PositionAGI}. 


\begin{figure}[t]
    \begin{minipage}[t]{0.49\linewidth}
        \includegraphics[width=1\textwidth]{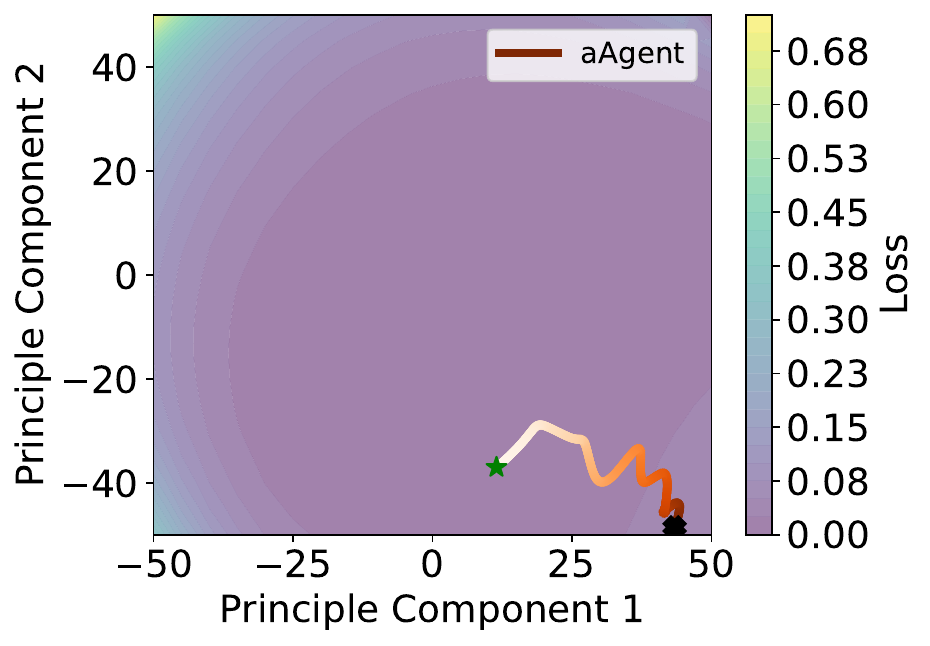}
        \captionsetup{labelformat=empty}
        \vspace{-0.7cm}
        \caption*{(a) aAgent }
    \end{minipage}
    \hfill
    \begin{minipage}[t]{0.49\linewidth}
        \includegraphics[width=1\textwidth]{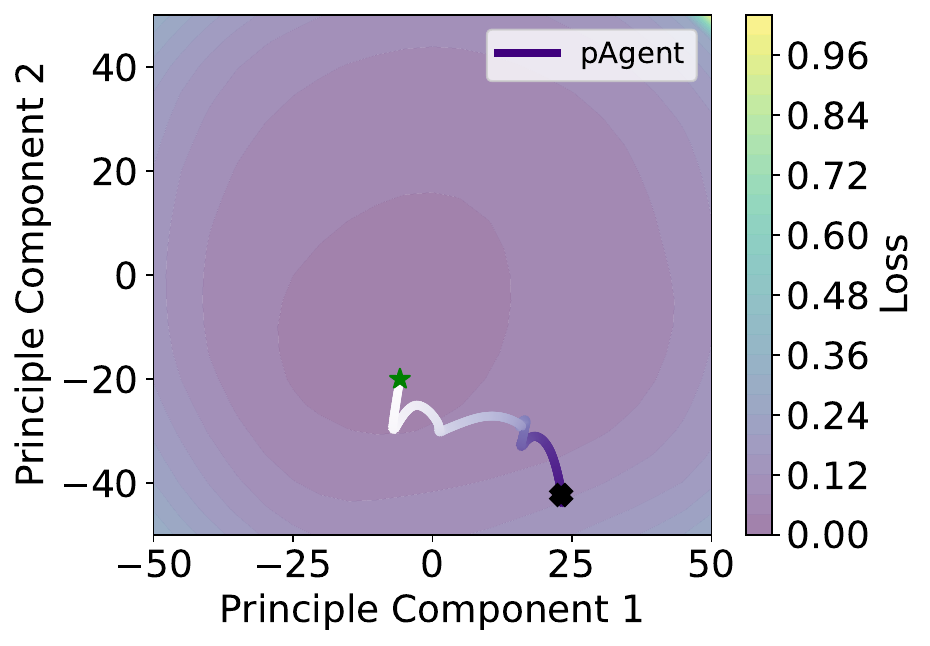}
        \captionsetup{labelformat=empty}
        \vspace{-0.7cm}
        \caption*{(b) pAgent}
    \end{minipage}
    \vfill
    \begin{minipage}[t]{0.49\linewidth}
        \includegraphics[width=1\textwidth]{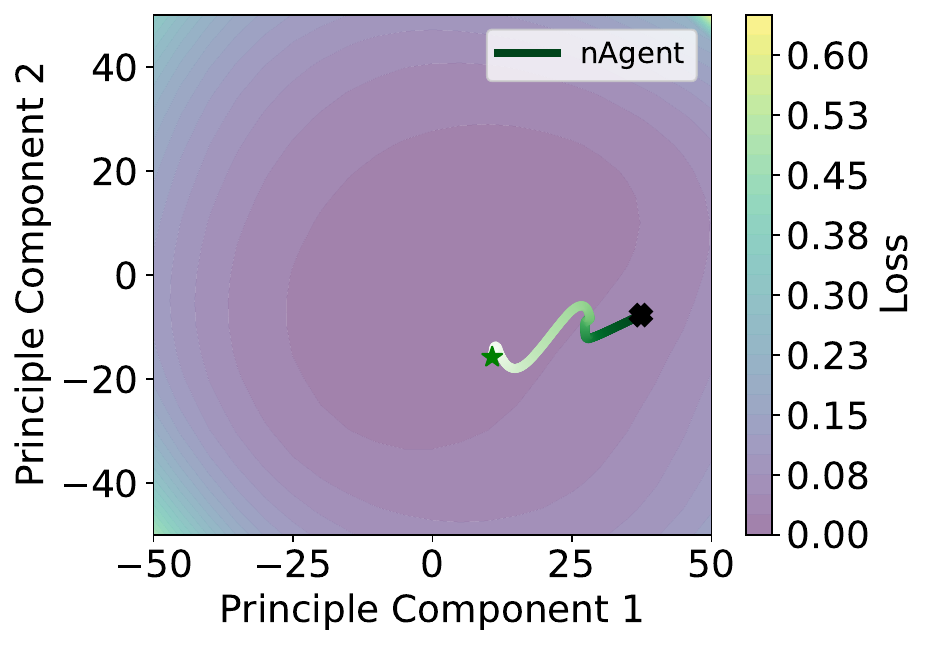}
        \captionsetup{labelformat=empty}
        \vspace{-0.7cm}
        \caption*{(c) nAgent}
    \end{minipage}
    \hfill
    \begin{minipage}[t]{0.49\linewidth}
        \includegraphics[width=1\textwidth]{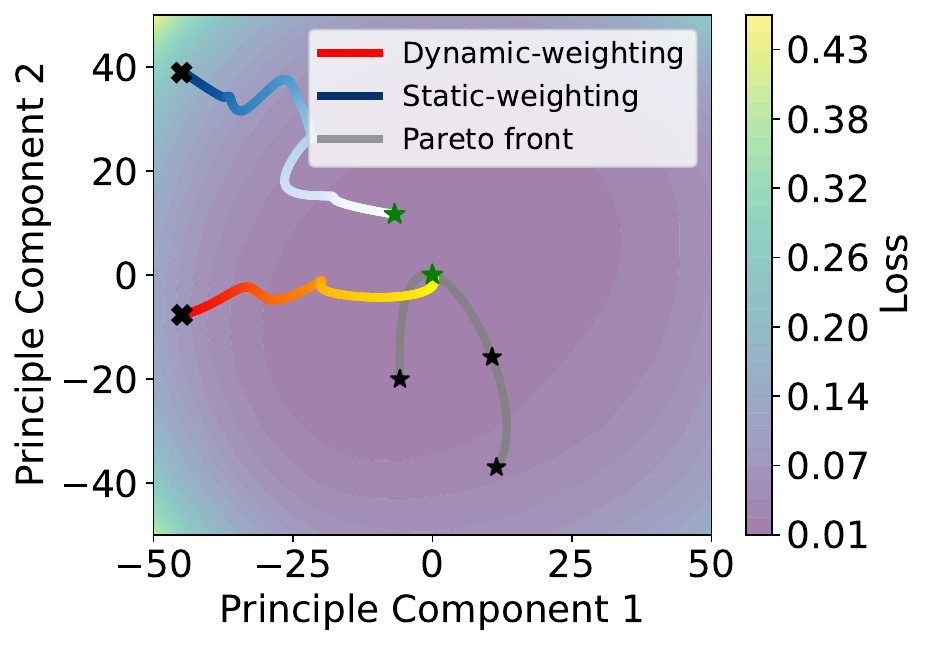}
        \captionsetup{labelformat=empty}
        \vspace{-0.7cm}
        \caption*{(d) SANet}
    \end{minipage}
    \vspace{-0.3cm}
    \caption{\small (a)-(c) Learning trajectories of three agents deployed at the physical layer (pAgent), application layer (aAgent), and network layer (nAgent) of a wireless network ({\bf black} \ding{54} denotes the initialization of the model, \gre{\bf green $\star$} denotes the end point); (d) SANet's collaborative learning trajectories of all three agents using static- and dynamic-weighting algorithms where \ltgr{grey line} denotes the Pareto front, {\bf black $\star$} denotes the local optimum of each agent.}
    \label{Fig_IntroExample}
    \vspace{-0.8cm}
\end{figure}



Despite its promise, AgentNet is still in the early stage of development\cite{Shavit2023PracticesAgenticAI}. In particular, due to the decentralized setting of AgentNet, coordinating and jointly optimizing the actions, goals, and objectives of different agents is notoriously challenging, often leading to new issues such as suboptimal consensus, difficulties in network-wide performance benchmarking, and security risks. 
To shed more light on the challenges of decentralized optimization of different agents, 
we consider an AgentNet consisting of three different agents deployed at the application layer (aAgent), physical layer (pAgent), and network layer (nAgent), with different objective functions for predicting user's semantic demands, channel state information (CSI) of wireless channels, and network layer traffic. Fig. \ref{Fig_IntroExample} (a)-(c) depict the learning trajectories as well as the contours of the objective functions of these three agents. We can observe that different objective functions of agents generally result in different learning trajectories. When all three agents are selected to collaborate in optimizing their model performance and generalization performance, and no conflict-resolving mechanisms are incorporated, e.g., the static-weighting algorithm in Fig. \ref{Fig_IntroExample}(d), agents may get stuck in some initializations, instead of converging to the Pareto front (the set of values in the objective space that includes all Pareto optimal solutions). By employing a conflict-resolving mechanism, such as our proposed dynamic-weighting algorithm (explained later in the paper), the learning trajectories of agents can navigate through these conflicting objectives and converge to the Pareto front.  

So far, most research efforts have focused on multi-agent task planning and adaptation, ignoring the fact that the networking architecture plays a fundamental role in the effectiveness of AgentNet systems. In particular, a recent study has already suggested that AgentNet is in essence a new communication networking paradigm that focuses on effective communication and seamless interaction between users and agents, as well as among diverse agents within a networked system\cite{xiao2025AgentNet}. This is fundamentally different from the existing disjoint computation-communication architecture in which AI model-based modules are deployed as plugins or over-the-top applications to enhance the data transportation and delivery capabilities of a networking system. 
There is still a lack of a comprehensive framework that can automatically detect a user's semantic goal and self-orchestrate and coordinate the networking functions and model development of agents in a decentralized fashion to fulfill the identified goal.

In this paper, we propose a semantic-aware AgentNet framework called SANet for cross-layer optimization. According to SANet, the user's semantic demand can be automatically detected, and an agent controller can directly select a set of agents deployed across multiple layers of a wireless system to collaboratively deploy models that address the detected demand. In particular, the agent controller does not know the datasets, states, and final decisions of the agents, but can convert the recognized goal into a set of subtasks to be distributed to the agents. Each agent corresponds to a physical or logical entity composed of one or more models and action functions designed with certain objectives to meet a specific goal. Different agents can have different computational capabilities and aim to optimize different objectives, depending on their engagement with heterogeneous datasets and data modalities. This is fundamentally different from existing cross-layer optimization solutions, which require a manually crafted multilayer task separation and coordination scheme, as well as a carefully tailored objective for each separate task that aligns with the unique characteristics and goals of each layer. 
%

SANet represents a paradigm shift from the traditional data-delivery-focused networking to semantic goal-driven and solution-finding-focused networking. The performance metrics for evaluating SANet are therefore no longer the same as those traditional data delivery metrics, e.g., data rate or packet delivery error. New metrics, such as the agents' optimization performance and model generalization performance, are needed to access SANet. In addition, because SANet is a decentralized framework in which different agents pursue distinct, often conflicting, objectives, it is generally impossible to find a globally optimized solution that maximizes the performance of all agents. Accordingly, in this paper, we focus on finding the Pareto-optimal solution that targets the best possible trade-offs among agents of conflicting objectives. 

We propose a multi-agent model sharing and partitioning (MoPS) for SANet, in which large models (e.g., deep learning models) of different agents are partitioned into shared and agent-specific parts that are jointly constructed and deployed according to agents' local computational resources. Three novel metrics for evaluating the SANet's performance are introduced: the agents' objective optimization error, dynamic-environment generalization error, and multi-objective conflicting error. 
Two decentralized optimization algorithms, static-weighting and dynamic-weighting, are introduced to optimize the three metrics. We derive theoretical bounds for all three metrics and prove that there exists a three-way tradeoff among optimization, generalization, and conflicting errors. \blu{We also propose a bandwidth-adaptive compression and decompression framework. This ``add-on" module complements the MoPS framework by enabling agents to perform in situ compression of their intermediate embeddings, dynamically adjusting to local resource constraints. }  Finally, we experimentally validate the performance of SANet by developing an open-source Radio Access Network (RAN) and core network-based hardware prototype, in which we deploy agents associated with three different layers. 

The main contributions of the paper are summarized as follows:

\noindent{\bf (1) AgentNet-based framework for cross-layer optimization:} We introduce a general framework for cross-layer optimization in wireless networks. In this framework, an agent controller proactively identifies the user's semantic goal and divides this goal into multiple subtasks associated with agents deployed at three different layers of a wireless system (application layer, network layer, and physical layer). Different agents interact with different environments and have different local datasets and optimization objectives. 
We formulate the decentralized optimization of SANet as a multi-agent multi-objective problem and focus on finding the Pareto-optimal solution among agents with distinct and potentially conflicting objectives. We propose three novel metrics for evaluating the SANet's performance. 

\noindent{\bf (2) Multi-agent model sharing and partitioning: } We propose a multi-agent collaborative learning framework, MoPS, based on model sharing and partitioning in which multiple large models, e.g., deep learning models, deployed at different agents for different subtasks can be partitioned into a shared part and an agent-specific part. The two parts are jointly constructed using two new unified interfaces: the embedding coordination interface (E-interface) and the gradient coordination interface (G-interface). 
We propose two decentralized optimization algorithms: static and dynamic-weighting, to jointly optimize the agents' optimization, generalization, and conflicting errors under different communication and computational resource costs.  

\noindent{\blu{\bf (3) Bandwidth-adaptive compression: }} \blu{ We develop a bandwidth-adaptive compression framework, in which a pair of compression and decompression modules can be learned and deployed as add-ons to complement the MoPS framework by enabling agents to perform in situ compression of intermediate embeddings, dynamically adjusting to their local bandwidth-constraints and task requirements. } 

\noindent{\bf (4) Theoretical analysis of the three-way tradeoff: }  We derive bounds on the optimization, generalization, and conflicting errors under the static and dynamic-weighting algorithms. Based on these bounds, we provide a theoretical characterization of the three-way tradeoff among these errors for SANet. 


\noindent{\bf (5) Prototype and experiments: } We develop a hardware/software platform, consisting of an open-source RAN and softwareized 5G core network, and deploy three types of agents at the application, network, and physical layers to evaluate the SANet performance. Experimental results show that the proposed MoPS provides up to $14.61\%$ performance gain with only $44.37\%$ of the Floating-Point Operations (FLOPs) required for inference at each agent compared to state-of-the-art algorithms. Our proposed dynamic-weighting algorithm reduces training errors caused by conflicting objectives by up to $83.81\%$, compared to the static-weighting algorithm.  


\section{Related Work}
\label{Section_RelatedWork}
\subsection{Cross-layer Optimization}

Recent results demonstrated that the joint optimization of multiple layers in a wireless system has the potential to mitigate the inherent limitations of traditional layered networking architectures and to significantly improve resource utilization, thereby meeting the increasingly stringent QoS demands of emerging applications. Previous works often focused on multi-layer joint optimization of a specific part of a networking system, e.g., the RAN. 
For example, the authors in \cite{Chang2024CrossLayerOpt} showed that the spectrum efficiency of the C-RAN can be improved by jointly optimizing parameters across the physical, data link, and transport layers, while guaranteeing statistical end-to-end QoS. 
The authors in \cite{Shiwen2023CrossLayerOpt} proposed a mixed discrete/continuous variable combinatorial solution to jointly optimize user scheduling at the MAC layer and beamforming at the physical layer in a multiuser multi-antenna system. 
Progress has also been made in specialized wireless systems, such as Reconfigurable Intelligent Surface (RIS) wireless systems and satellite networks. In particular, in \cite{LiWei2025CrossLayerOpt}, the authors proposed a deep reinforcement learning approach to jointly optimize video quality at the application layer and the physical layer environment controlled by RIS\cite{LiWei2025CrossLayerOpt}. 

Despite these promising progresses, existing solutions often require a predefined multi-layer decomposition scheme. The objectives must be manually crafted to fit the specific characteristics and goals of each layer. In contrast, in this paper, we focus on autonomous and decentralized cross-layer optimization based on AgentNet, whereby an agent controller proactively senses the semantic goals of users and autonomously divides the sensed goal into different subtasks to be tackled by agents at different layers. Our proposed framework is general and can preserve the layered architecture of existing systems while facilitating scalable, adaptive, and cooperative decision-making that accommodates dynamic network conditions and heterogeneous requirements. 

\subsection{AI-based Network Optimization}

AI models have been increasingly deployed to optimize various parts of the networking architecture, from smart applications and services to RAN\cite{alzailaa2025review} and core networks \cite{taleb2023ai}. Compared to traditional optimization-based solutions that rely on simplified, static models, AI-based solutions leverage data-driven learning, which can efficiently solve computationally intractable, high-dimensional, and non-convex optimization problems presented by the scale and heterogeneity of modern wireless systems. For example, various AI-based solutions have been adopted for analysis, estimation, and optimization of physical-layer functions. In particular, the authors in \cite{Zhu2024SANe} proposed SANSee, a novel framework that employs a physical layer semantic-aware network (pSAN) to characterize data correlations across different locations. A zero-shot transfer learning-based solution was introduced to allow location-specific models to be directly transferred from a limited number of pre-trained models, achieving accuracy comparable to locally trained, supervised models. 
%
AI capabilities were also incorporated into the core network since the very first version of 5G. More specifically, the Network Data Analytics Function (NWDAF) was introduced as a key functional component for data analysis and processing in 3GPP Release 15. Its scope and capability have been extended significantly since that release. For example, Release 18 has emphasized the integration of AI and ML into NWDAF, enabling more sophisticated analytics and automation capabilities. NWDAF in Release 19 focused on providing the necessary insights and recommendations for supporting self-optimization and self-healing \cite{lin2025bridge}. The ITU-T and IMT-2030 (6G) group have also initiated the work for standardizing semantic-aware networking with the main objective to enable the networking systems to learn and adapt according to the dynamics of users' semantics and network intents\cite{ITU2023SAN, XY2024ACMGetMobileSAN}.

Despite the considerable research in this domain, the majority of AI-based network optimization solutions still employ passive learning frameworks that are trained exclusively on historical datasets. These methods operate under the often-invalid premise that past statistical data patterns will persist across future deployments, thereby limiting their capacity to adapt effectively to the increasingly dynamic, diversified, and personalized service requirements of modern telecommunication environments. In this paper, we propose an AgentNet-based solution in which the user's semantic goal and demand can be predicted and proactively addressed by a carefully selected subset of collaborative agents. 

\subsection{Agentic AI and AgentNet}

Recently, Agentic AI has attracted significant interest due to its potential to fundamentally shift the paradigm from reactive, passive, and monolithic learning to autonomous, goal-driven, multi-agent cooperation based on decentralized learning and solution finding. Critically, agentic AI systems are evolving towards a decentralized structure where heterogeneous AI agents, each with distinct skillsets, specialized knowledge, and optimization objectives, collaborate 
to achieve a complex, shared goal. This approach offers significant advantages over traditional centralized systems, which suffer from a single point of failure, scalability constraints, and high latency. The decentralized learning approach of agentic AI differs from most existing distributed learning systems (such as Federated Learning) in that it enables autonomous decision-making and real-time interaction during execution, rather than solely focusing on the distributed implementation of a single centralized goal. The resulting system is inherently more resilient and fault-tolerant, as the failure of one AI agent does not compromise the operation of the entire system. Furthermore, agentic AI promotes enhanced privacy by keeping sensitive data local to each agent\cite{Sapkota2025AIagentsandAgenticAI}. 
Most existing research in this topic focuses on exploring the reasoning capacities for multi-agent planning, tool use, memory, and perception of dynamic environments to enable complex, multi-step execution \cite{Durante2024AgentAISurvey}. Significant milestones include the development of multi-agent systems (MAS) and orchestration frameworks that facilitate cooperation and communication between specialized agents to decompose and solve high-level tasks, yielding measurable improvements in efficiency across diverse domains. For example, the authors in \cite{Wang2024MobileAgentV2} focus on optimized task decomposition and resource allocation among agents for maximizing parallel efficiency and minimizing communication overhead. A key direction in this research involves defining explicit, communicative roles for LLM agents, as demonstrated by the work in \cite{Wang2024RethinkBoundofLLRReasoning}. In this work, the authors proposed the CAMEL framework, where agents are prompted into a cooperative role-playing dynamic to solve complex tasks through guided conversation. 
\blu{Recently, several works have explored practical applications of agentic AI in wireless networks. For example, the authors in \cite{tong2025wirelessagent} demonstrated that a single agent can interpret complex network states and devise adaptive network configurations. In \cite{li2026comagent}, the authors introduced a multi-LLM-based agentic AI framework that automates the transition from high-level user intents to mathematically consistent cross-layer optimization formulations through iterative planning and execution-based verification. To enhance real-time adaptability in dynamic environments, AutoMAS \cite{yuan2025automas} enables agents to autonomously select and adapt wireless optimization algorithms by bridging theoretically-grounded methods with agents' perception ability.}


AgentNet is an agentic AI-native networking system that is composed of diverse AI agents with autonomy, goal-oriented behavior, and adaptability that enable agents to make independent decisions and execute actions to achieve specific objectives with no or minimal human supervision\cite{acharya2025agentic}. In \cite{xiao2025AgentNet}, we introduced a Generative Foundation Model (GFM)-based architecture that facilitates interaction, collaborative learning, and efficient knowledge transfer among multiple GFM-as-agents. Key performance metrics for measuring the performance of AgentNet were introduced, including implementable environment complexity, model generalization, multi-domain knowledge-based reasoning capability, resource requirements, and complex goal achievability. Two 6G applications, digital-twin industrial automation and metaverse-based infotainment systems, were considered to describe how AgentNet can be designed to support efficient task-driven interactive networking systems with diverse AI agents. 

Agentic AI systems and AgentNet are still in the early stages of development. In this paper, we take steps to develop novel solution concepts and performance metrics of the general AgentNet system. Specifically, motivated by the fact that AgentNet is a decentralized system, we consider a Pareto-optimal solution, instead of the global optimal solution, to characterize the tradeoff among different performance metrics. We investigate the three-way tradeoff among optimization, generalization, and conflict-resolving performance of AgentNet from both theoretical and practical perspectives. To the best of our knowledge, this is the first work that investigates AgentNet for cross-layer optimization of a wireless networking system.

\begin{figure}
\centering
\includegraphics[width=1\linewidth]{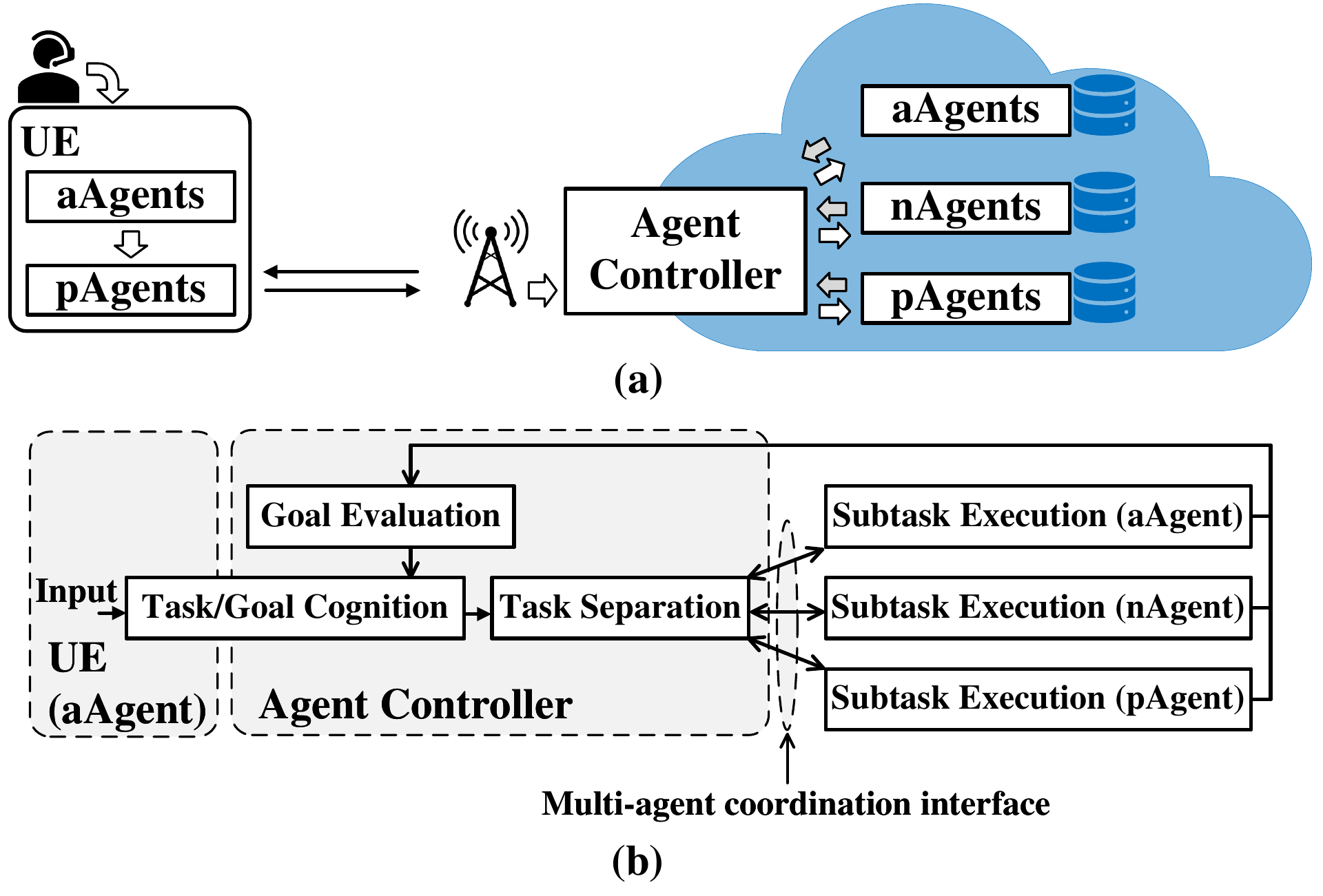}
\vspace{-0.7cm}
\caption{(a) System model, and (b) operational workflow for a general AgentNet-based cross-layer optimization framework for wireless networks.}
\label{Figure_SANNetModel}
\vspace{-0.6cm}
\end{figure}

\section{System Model and Problem Formulation}
\label{Section_SystemModel}
\subsection{System Model}
We consider a general AgentNet-based cross-layer optimization problem for a wireless networking system, as illustrated in Fig. \ref{Figure_SANNetModel}(a). 
%
Let $\cM$ be a finite set of tasks that can be performed by a set of agents. To simplify our description, in this paper, we focus on coordination and collaborative learning involving at most three different types of agents deployed in three different layers: application layer, network layer, and physical layer, labeled with superscripts $a$, $n$, $p$, respectively. 
Our proposed solution, however, can be directly extended to more complex systems that include agents in more layers, as will be illustrated later.  





\noindent{\em (1) Application layer agent (aAgent)}: This corresponds to the agent that can directly interact with users through some application interfaces and can adjust a range of application parameters according to the semantics of the users. For example, live streaming applications such as video conferencing and metaverse-based interactive gaming installed on smart devices can infer and predict the semantic request of the users based on the input of the application interface and adjust the application parameters, such as video resolutions and refresh rates, according to the predicted request. More formally, we define an aAgent by a tuple ${\cal G}^a = \langle \cA^a, \cS^a, \Gamma^a, \cD^a \rangle$ where $\cA^a$ is the set of acts that can be taken by aAgent, $\cS^a$ is the set of environmental states for an aAgent to sense and decide its act, $\Gamma^a$ is the set of task-specific loss functions for aAgent to optimize, and $\cD^a$ is the set of data samples for training the aAgent's model. In this paper, we consider a specific AgentNet system in which each agent consists of an AI model and an action function. In particular, an AI model with parameters $\bomega^a$ can be trained to proactively sense and predict the future semantic goal of users based on the currently observed state and the observation history, and an act function will decide the specific action to be taken based on the model output. Generally speaking, the aAgent cannot control the decision-making processes of other agents. Its performance, however, can be influenced by the actions of other agents, including those in the physical layer and network layer. We can write the loss function of aAgent when being selected to address task $m$ as $\cL^a_m(\bOmega_m, \cD^a) = \frac{1}{|\cD^a|} \sum_{d^a \in \cD^a}\ell^a_m(\bOmega_m, d^a)$ where $\cL^a_m \in {\Gamma^a}$, $\bOmega_m = \langle \bomega^a_m, \bomega^p_m, \bomega^n_m \rangle$, and $\bomega^p_m$ and $\bomega^n_m$ are model parameters associated with the agents in physical layer and network layer, respectively, as will be illustrated later.

\noindent{\em (2) Physical layer agent (pAgent)}: This corresponds to the agent that can interact and adapt to the physical layer environments. For example, a pAgent deployed at the gNB or UE in a mobile network can develop a model to estimate the spectrum availability, and in some cases, keep track of the channel state information (CSIs) of the uplink and downlink channels, and adopt an action function to make recommendations in transmitter and receiver parameter choices, such as bandwidth and channel selections. Similar to the aAgent, we can also define a pAgent by a tuple ${\cal G}^p = \langle \cA^p, \cS^p, \Gamma^p, \cD^p \rangle$ where $\cA^p$ is the action space, 
$\cS^a$ is the environmental state space, and $\Gamma^p$ is the set of task-related loss functions, and $\cD^p$ is the set of data samples for training the relevant models.
Similar to the aAgent, the decision-making process of the pAgent can also be performed by a model parameterized by $\bomega^p_m$ and an action function that can output an action decision based on the model output. We can write the loss function learned by pAgent for task $m$ as $\cL^p_m(\bOmega_m, \cD^p)$ for $\cL^p_m \in {\Gamma^p}$. 

\noindent{\em (3) Network layer agent (nAgent)}: This corresponds to the agent that interacts with the network layer environments. For example, an nAgent deployed in the transportation layer or the core network of a mobile networking system can keep track of the routing and bandwidth resources between any network entities connecting the UE and its intended destination and adjust the routing and the amount of bandwidth resources accordingly. We define an nAgent by a tuple ${\cal G}^n = \langle \cA^n, \cS^n, \Gamma^n, \cD^n \rangle$ where $\cA^n$ is the set of acts that can be taken by the nAgent, $\cS^n$ is the set of network layer environmental states for an nAgent to estimate and decide its act, $\cL^n$ is the set of task-related loss functions, and $\cD^n$ is the set of data samples for constructing the nAgent's model. Let $\bomega^n_m$ be the model parameters associated with the nAgent. 
We can write the loss function associated with task $m$ of pAgent as $\cL^n_m(\bOmega_m, \cD^n)$ for $\cL^n_m \in {\Gamma^n}$. 

We can observe that different types of agents generally have different local data modalities, optimization objectives, and models. The traditional way of developing AI models for multi-layer/multi-objective problems often involves training either distinct, task-specific models for individual layers or a single, monolithic foundation model for all three layers. Although the former offers high fidelity for each objective in each layer, it incurs significant computational and storage overheads due to redundant training and deployment. Conversely, the latter may struggle to tailor its output to the nuanced, objective-specific requirements of each layer of the objective. 

\subsection{Problem Formulation}

As mentioned earlier, each agent is an independent decision maker, and therefore, the agent controller cannot directly control the decision-making processes of the agents. It can only choose a specific subset of agents when a certain task demand is detected, and then coordinate the local decision-making processes of the selected agents for the task demand of the user. The multi-agent cross-layer optimization problem can be defined as follows: 
%
\begin{eqnarray}
\mbox{\bf P1:} \ \min\limits_{\bOmega_m} \cL_{m}(\bOmega_m) \!\!:=\! \langle \cL^a_m (\bOmega_m,\! \cD^a),\! \cL^p_m(\bOmega_m,\! \cD^p),\! \cL^n_m(\bOmega_m,\! \cD^n) \rangle. \nonumber
\end{eqnarray}

Note that, different from the previous works focusing on finding the global optimization solution to minimize a single objective function, e.g., the weighted sum of local losses of all agents, the optimization objective of problem {\bf P1} involves a vector of objectives from different agents that are associated with different layers of the mobile system. In this case, it is generally impossible to find a single global optimal solution that minimizes the loss functions of all the agents. 
%
To address this issue, in this paper, we consider the Pareto optimal solution, a metric commonly adopted for evaluating the tradeoff among a collection of possibly conflicting goals and objectives. 
More formally, we define the Pareto optimal solution for the multi-agent AgentNet system as follows:

\begin{definition}
A solution profile $\bOmega$ is called {\em Pareto stationary solution} if there exists a set of non-negative weights $\gamma^a$, $\gamma^p$, and $\gamma^n$ such that $\sum_{i\in\{a, p, n\}} \gamma^i = 1$ and $\sum_{i\in\{a, p, n\}} \gamma^i \nabla \cL^i (\bOmega, \cD^i)=0$. A solution profile $\bOmega^*$ is {\em Pareto optimal} if no other Pareto stationary solution $\bOmega$ for $\bOmega \neq \bOmega^*$ such that $\cL^i (\bOmega) \le \cL^i (\bOmega^*, \cD^i)$ for all $i \in \{a, p, n\}$ and $\cL^i (\bOmega) < \cL^i (\bOmega^*, \cD^i)$ for at least one $i \in \{a, p, n\}$.  
\end{definition}

From the above definition, we can observe that the Pareto stationary solution seeks a feasible solution set in which no other feasible solution can improve one agent's objective without causing a deterioration in at least one other agent's objective. In other words, in a Pareto stationary solution, there is no common descent direction for the selected agents in which all the gradients $\nabla \cL^i_m (\bOmega_m, \cD^i)$ have negative inner products with $\gamma^i_m$ for all $i \in \{a, p, n\}$. Suppose all the loss functions $\cL^i_m (\bOmega_m, \cD^i)$ for $i \in \{a, p, n\}$ are smooth and differentiable, a necessary condition for $\bOmega$ to be Pareto stationary is that the convex hull of all gradients $\langle \nabla \cL^a_m (\bOmega_m, \cD^a)$, $\nabla \cL^p_m (\bOmega_m, \cD^p)$, $\nabla \cL^n_m (\bOmega_m, \cD^n) \rangle$ contains the origin, which means there exist non-negative weights $\gamma^a_m$, $\gamma^p_m$, and $\gamma^n_m$ such that $\sum_{i\in\{a, p, n\}}\gamma^i_m \nabla \cL^i_m (\bOmega_m, \cD^i)=0$. In a more general setting, the minimum norm of the weighted sum of gradients, defined as 
\begin{eqnarray}
\min_{\gamma^i_m, \forall i \in \{a, p, n\}} \sum_{i\in\{a, p, n\}}\gamma^i_m \nabla \cL^i_m (\bOmega_m, \cD^i)
\label{eq_equivParetoStation}
\end{eqnarray}
has been commonly considered as the main measure of the Pareto stationary solution.

\section{Multi-agent Optimization Objectives}
\label{Section_OptimObjectives}

Since the main focus of AgentNet-based networking systems is no longer maximizing the volume of transported data packets throughout the network, traditional metrics such as data rate and bit/symbol-error-rate will be insufficient for evaluating its performance. In the remainder of this paper, we consider the following different but correlated performance metrics to investigate the collaborative performance of AgentNet systems. 



\subsection{Objective Optimization Error (O-error)} 
The main objective of each agent is to optimize its local objective, i.e., minimizing its local loss function. We assume that each agent employs a stochastic gradient descent (SGD)-based approach to train its local model. We can then evaluate the optimization performance of an individual agent's model based on the gradient of its local loss function, calculated based on its locally collected dataset. More specifically, we define the optimization error of individual agent $i$ when being selected for task $m$ as follows:



\begin{definition}
The optimization error (O-error) of an individual agent $i$ for $i\in \{a, n, p\}$ to perform task $m$ based on dataset $\cD^i$ is defined as 
\begin{eqnarray}
    \cE^i_{O} = \| \nabla \cL^{i}_m (\bomega_m^i, \cD^i) \|. 
    \label{eq_Definition_OptimizationError}
\end{eqnarray}
\end{definition} 
Suppose a set of agents, $\cal T$, has been selected to execute the same task $m$. We define the joint optimization error (O-error) as 
\begin{eqnarray}
    \cE_{O} = \| \sum_{i\in\{a, p, n\}}\gamma^i_m \nabla \cL^{i}_m (\bOmega_m, \cD^i) \|. 
\end{eqnarray}
where $\gamma^i_m$ is the weighting coefficient that reflects the influence of each individual objective's learning direction on the overall optimization process. Generally speaking, the weighting coefficients of different agents need to be carefully selected to achieve a robust and well-rounded balance among different agents' objectives towards the user's overall semantic goal. 




\subsection{Generalization Error (G-error)}
We also consider the agent's generalization capability when being deployed in an open dynamic environment that cannot be fully captured by its local training dataset. Specifically, we adopt a commonly adopted metric to quantify the generalization capability of each agent, defined as follows:

\begin{definition}
We define the generalization error (G-error) as the discrepancy between the agent deployment performance based on the real data distribution and the training performance obtained on the training dataset. More formally, let ${\tilde \cL}^{i}_m (\bomega_m^i)$ be the population loss of the agent $i$'s model for $i\in\{a, n, p\}$, evaluated over the real data distribution of task $m$. We can then define G-error of agent $i$'s model $\cE_G^i$ as the difference of the gradients between the population loss ${\tilde \cL}^{i}_m (\bomega_m^i)$ calculated based on the real distribution of agent deployment and the empirical loss $\cL^i_m(\bomega_m^i, \cD^i)$ obtained based on the training dataset $\cD^i$, i.e., the G-error of agent $i$ can be written as: 
\begin{eqnarray}
    \cE_G^i = \| \nabla \cL^i_m(\bomega_m^i, \cD^i) - \nabla {\tilde \cL}^{i}_m (\bomega_m^i) \|.
\end{eqnarray}

Similar to the O-error, when a set of agents, $\cal T$, has been selected to act collaboratively for the same task $m$, we define the generalization error of a collaborative set of agents as follows: 
\begin{eqnarray}\label{E_G}
    \cE_G = \| \sum_{i\in\{a, p, n\}}\gamma^i_m (\nabla \cL^i_m (\bOmega_m, \cD^i) - \nabla {\tilde \cL}^{i}_m (\bOmega_m) ) \|,
\end{eqnarray}
\end{definition}

\subsection{Multi-objective Conflicting Error (C-error)}


As mentioned earlier, one of the key differences between AgentNet and the traditional centralized networking system is that, in AgentNet, each individual agent focuses on optimizing its own objective, and different objectives of different agents may conflict with each other.  
We can therefore introduce the following metric to quantify the performance degradation caused by conflicting objectives among different agents. 

\begin{definition}
We define the (multi-agent objective) conflicting error (C-error) as the error caused by the conflicting gradients among different agents, defined as follows: 
\begin{eqnarray}
    \cE_C := \| \sum_{i\in\{a, p, n\}}(\gamma^i_m-\gamma^{i*}_m) \nabla \cL^i_m (\bOmega_m, \cD^i) \|,
\end{eqnarray}
where $\gamma^{i*}_m$ denotes the optimal weight of the agent $i$ for task $m$.
\end{definition}

\section{SANet Architecture and Optimization Algorithms} 
\label{Section_Architecture_Algorithms}

\subsection{Architecture Overview}

We propose SANet, a semantic-aware AgentNet framework that supports multi-agent/multi-objective cross-layer optimization. 

The workflow of the SANet is described as follows: 
\blu{
{\bf (1) (Semantic-aware) task/goal cognition:} A set of prompts that reflects the key semantic intent of the users is predefined and can be recognized by an interface deployed at the UE to capture the user's input. The user's input for these key prompts will be tokenized and converted into low-dimensional embeddings for transmission to the agent controller. The agent controller leverages an LLM-based interface to identify the user's semantic goal, which is then linked to a specific task $m$ in a finite task space ${\cM}$. {\bf (2) Agent selection and model partition:} The agent controller will reason about the appropriate model architecture and logically break down the identified task goal into a set of discrete subtasks, each of which can be solved by an individual agent. A set of agents that can solve the identified subtasks will then be selected to solve the task goal. In this case, each agent will be assigned a specific loss function corresponding to a particular sub-task. {\bf (3) Multi-agent model partition and sharing: } The agent controller will learn the parameters of the shared-part model according to the identified task and the capabilities of the selected agents. The agent-specific part of the model parameters will be learned by each agent during the model training. {\bf (4) Collaborative inference: } After the model training process, each agent will send its model output to the agent controller based on its local input. The agent controller will then output the final results of the selected agents, which will then be sent back to the agents to decide their actions. The agent controller will evaluate the performance and completeness of the users' goals based on the output results. 
}

\subsection{\blu{Semantic-aware Task/goal Cognition}}
\blu{ We propose a hierarchical pipeline-based semantic cognition and task-to-agent allocation framework, in which the high-level semantic intent recognized from the user's input can be mapped to executable multi-agent workflows with near-zero latency. 
The detailed procedures are described as follows. 
%
First, an ontological mapping scheme needs to be defined, in which specific keyword clusters and prompts can be mapped to discrete, executable tasks. A specialized aAgent is deployed at the user-end devices to monitor user input in real time. Upon the detection of targeted semantics, the aAgent transforms the context data associated with the detected semantics into a low-dimensional embedding to be sent to the agent controller with small communication overhead. The agent controller hosts an LLM that can perform latent-space reasoning to resolve semantic ambiguities.  
To eliminate the latency typically associated with real-time reasoning, our proposed AgentNet architecture integrates Transformer-based prediction models across all layers of agents. These models forecast future semantic trajectories, allowing the system to preemptively prepare resources before a formal demand is finalized. Once the LLM confirms the high-fidelity intent, the system requires a robust engine to translate this intent into actionable logic to be executed by a selected subset of agents.  
While the term ``semantic" may encompass a broad spectrum of linguistic and cognitive theories regarding meaning, in this paper, we adopt a functionalist perspective tailored to agentic orchestration in AgentNet systems. We define semantic intention not as an abstract concept, but as a goal-oriented directive that is fundamentally mappable to a finite set of discrete, executable tasks. Specifically, we focus on intentions that can be decomposed and dispatched across a distributed networking system, where the fulfillment of the user's intent is realized through the collaborative synergy of a selected set of agents.
In this paper, we employ OpenManus integrated with an LLM, e.g., Qwen-14B, Qwen-7B, and Qwen-1.7B, as the core semantic cognition and agentic orchestration engine at the agent controller to facilitate the transition from abstract user requests to a structured plan, utilizing its specialized agentic logic to allocate these sub-tasks across a distributed set of agents.} 

\subsection{Multi-agent Model Partition and Sharing}

\begin{figure*}[t]
\centering
\includegraphics[width=1\linewidth]{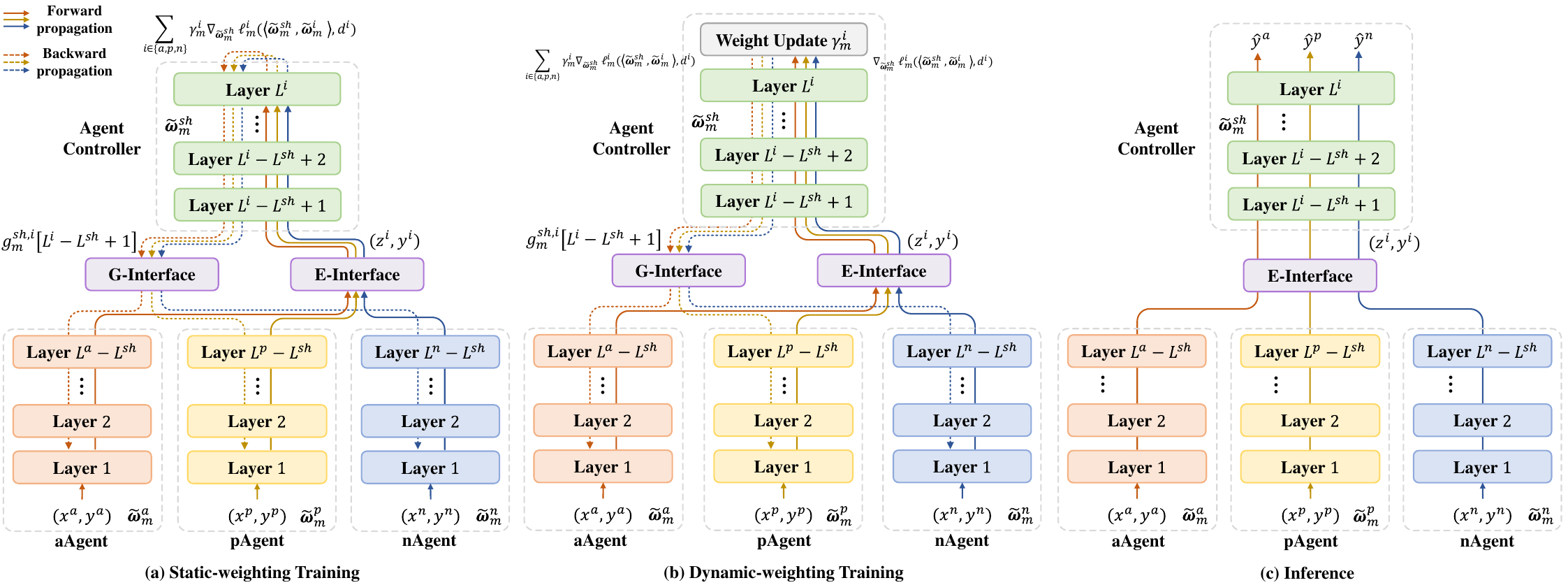}
\vspace{-0.6cm}
\caption{Model training procedures of (a) static-weighting and (b) dynamic-weighting algorithms, and (c) inference procedures of both algorithms.}
\label{Fig_StaticDynamicAlgorithms}
\vspace{-0.6cm}
\end{figure*}

We propose a novel multi-agent model partition and sharing (MoPS)-based framework for collaborative model construction in which multiple large models, e.g., deep learning models, deployed at different agents for different subtasks can be partitioned into {\em shared-part} and {\em agent-specific part} according to the computational resource constraints of different agents. 
More specifically, in MoPS, the agent controller hosts the shared-part model's parameters, forming a common foundation for all subtasks, and each individual agent, constrained by its local computational capacity, receives a carefully selected, localized part of the model parameters, trained exclusively on the agent's local dataset for the targeted subtasks. MoPS leverages available resources at the agents efficiently, avoiding the need for each agent to process the entire model. Our proposed framework is scalable and applicable to large AgentNet systems, composed of diverse agents implemented in complex environments. 

More formally, 
suppose each agent $i$ needs to construct an $L_i$-layer model for executing its assigned subtask, in which $L^{sh}$-layer shared-part model parameterized by $\btomega^{sh}_{m}$ is deployed at the agent controller and the rest $(L^i - L^{sh})$-layered agent-specific model  parameterized by $\btomega^{i}_{m}$ is deployed locally, for $L^{sh} \le L^i$, $\forall i \in \{a, p, n\}$. Note that the number of layers of agent-specific part models can be different at different agents, i.e., $L^i \ne L^j$ for $i \neq j$ and $i,j \in \{a, p, n\}$.
In this case, the local loss function of each agent $i$ can be written as $\cL^i_m (\langle \btomega^{sh}_m, \btomega^i_m \rangle, \cD^i)$ where $\bomega^{i}_m = \langle \btomega^{sh}_m, \btomega^i_m \rangle$ 
and $\cD^i$ is its local dataset. 

The optimization problem {\bf P1} can then be rewritten in the following form: 
\begin{eqnarray}
    \mbox{\bf P2:} \min\limits_{\substack{\langle \btomega^{sh}_m, \btomega^a_m, \\ \btomega^p_m,\btomega^n_m\rangle}} \| \sum_{i \in \{a, p, n\}} \gamma_m^i \nabla \cL_m^i (\langle \btomega^{sh}_m, \btomega^i_m \rangle, \cD^i) \|^2.
\end{eqnarray}

Before we describe the detailed model training and inference procedures to solve the above problem, let us introduce the interface for privacy-preserving coordination among different types of agents. Since agents cannot expose their local datasets, but can coordinate their local learning and inference process based on their intermediate model output embeddings and the gradients, we introduce the following two types of interfaces for coordinating the intermediate embeddings and gradients for multi-agent collaborative model training and inference:

    
\noindent{\bf (1) E-interface (Embedding coordination interface)}: 
    The E-interface establishes a standardized, fixed-size vector space, ensuring that the rich, high-dimensional representations output by partial models at different agents are compatible with each other for the joint model training and inference. In particular, E-interface specifies a specific protocol for data serialization and numerical normalization to manage disparate output formats and scales of different agents. 
    More specifically, let $f_{\btomega^i_m}(\cdot)$ be the mapping function that maps any input data sample into the final layer output embedding of the model with parameter $\btomega^i_m$. 
    In our proposed MoPS framework, each agent will not upload any local data samples, but can periodically coordinate with others by uploading a batch of embedding-label pairs $\langle z^i_k, y^i_k \rangle$ to the agent controller, where $z^i_k = f_{\btomega^i_m}(x^i_k)$ is the final layer output embedding of the model with parameter $\btomega^i_m$ by passing the input data $x^i_k$ through it, and $y^i_k$ is the label. 
    

\noindent{\bf (2) G-interface (Gradient coordination interface)}: G-interface, which complements the E-interface, is necessary for coordinating the learned gradients of subtask specific losses among agents. 
    Unlike the E-interface, which is primarily designed for inference, the G-interface regularizes the coordination during multi-agent model training. The interface defines a clear and consistent structure for gradient tensors, regardless of the local model's complexity or the specific objective function being used. 
    In our proposed MoPS framework, after receiving the embedding-label pair $\langle z^i_k, y^i_k \rangle$ from each agent $i$, the agent controller will update the shared-part model $\btomega^{sh}_{m}$ by calculating the corresponding gradient $\nabla_{\btomega^{sh}_{m}} \ell^i_m(\langle \btomega^{sh}_m, \btomega^i_m \rangle, d^i)$ using the backpropagation operation. More formally, we use the subscript $t$ to denote the model parameters in the $t$th round of model coordination. For the $t$th round of coordination, let $g^{sh,i}_{m,t} [L]$ be the $L$th-layer gradient tensor calculated by the agent controller during the $t$th round of coordination based on the loss function of agent $i$, i.e., we have $g^{sh, i}_{m,t} [L] = \nabla_{\btomega^{sh}_{m, t}[L]} \ell^i_m( \langle \btomega^{sh}_m, \btomega^i_m \rangle, d^i)$ for $L^i-L^{sh}+1 \le L \le L^i$. We also define $g^{i}_{m,t} [L']$ as the $L'$th-layer gradient tensor of the agent-specific model $g^{i}_{m,t} [L'] = \nabla_{\btomega^{i}_{m, t}[L']} \ell^i_m( \langle \btomega^{sh}_m, \btomega^i_m \rangle, d^i)$ calculated during the $t$th round of coordination for $1 \le L' \le L^i-L^{sh}$. At the end of the $t$th coordination, the controller will distribute the $g^{sh,i}_{m,t} [L^i-L^{sh}+1]$ 
    to each agent $i$ via the G-interface.      

Let us now describe the multi-agent gradient descent algorithm that can solve problem {\bf P2} as follows: 
We consider $T$ rounds of collaborative model training among a set of selected agents $\{a, p, n\}$ to meet the demand of the user's task $m$. 
During each round of training, each agent will first update the agent-specific part model based on its local dataset and then coordinate with the agent controller to update the shared-part model. More formally, at the beginning of the $t$th round of coordination, we assume the agent controller has already deployed the shared part of the model $\btomega^{sh}_{m,t}$ and assigned the subtask-specific part model $\btomega^{i}_{m,t}$ to each agent $i$. Then, during the $t$th round of coordination, the following steps are sequentially executed by the agents and the agent controller:  

\noindent{\bf (1) Local model updating (at each agent):} Each agent $i$ first downloads the gradient tensor $g^{sh,i}_{m,t-1}[L^i-L^{sh}+1]$ from the agent controller via G-interface 
It then computes the gradient of its local model parameters based on the chain rule as follows: 
\begin{flalign}
    &\nabla_{\btomega^{i}_{m, t-1}} \ell^i_m(\langle \btomega^{sh}_{m, t-1}, \btomega^i_{m, t-1} \rangle, d^i_{t-1}) \\
    &= \left[ \begin{array}{c}
    g^{i}_{m,t-1}[L^i - L^{sh}] \\
	  g^{i}_{m,t-1}[L^i - L^{sh} - 1] \\
    ... \\
	g^{i}_{m,t-1}[1] \\
    \end{array} \right] \nonumber
\end{flalign}

Agent $i$ then updates its local model parameter $\btomega^{i}_{m,t}$ via the standard gradient descent as follows:
\begin{equation}\label{local_model_update}
    \btomega^{i}_{m, t} = \btomega^{i}_{m,t-1} - \beta_t \nabla_{\btomega^{i}_{m, t-1}} \ell^i_m(\langle \btomega^{sh}_{m, t-1}, \btomega^i_{m, t-1} \rangle, d^i_{t-1}),
\end{equation}
where $\beta_t$ is the step size of the model update.


\noindent{\bf (2) Embedding uploading (via E-interface): } Each agent $i$ calculates the model output embedding $z^i_{t} = f_{\btomega^i_{m, t}}(x^i_{t})$ and then uploads the embedding-label pair $\langle z^i_t, y^i_t \rangle$ 
to the agent controller via E-interface. 

\noindent{\bf (3) Shared model updating (at agent controller):} Upon receiving the embedding-label pairs from all called agents, the agent controller updates the shared-part model $\btomega^{sh}_{m, t}$ by performing the aggregated gradient descent as follows: 
\begin{flalign}
    &\btomega^{sh}_{m, t}\nonumber \\
    =& \ \btomega^{sh}_{m,t-1} - \beta_t \sum_{i \in \{a, p, n\}} \gamma^i_m \nabla_{\btomega^{sh}_{m, t-1}} \ell^i_m(\langle \btomega^{sh}_{m, t-1}, \btomega^i_{m, t} \rangle, d^i_t) \nonumber\\
    =& \ \btomega^{sh}_{m,t-1} \!-\! \beta_t \!\!\! \sum_{i \in \{a, p, n\}} \gamma^i_m 
    \left[ \begin{array}{c}
    g^{sh,i}_{m,t-1}[L^{i}] \\
	  g^{sh,i}_{m,t-1}[L^i-1] \\
    ... \\
	g^{sh, i}_{m,t-1}[L^i-L^{sh}+1] \\
    \end{array} \right], \label{eq_shared_model_update}
\end{flalign}
where $\gamma^i_m$ is the weighting coefficient of agent $i$. 

\noindent{\bf (4) Updated gradient distribution (via G-interface):} The agent controller will calculate the output gradient tensor $g^{sh,i}_{m,t}[L^i-L^{sh}+1]$ to be sent to each agent $i$ via the G-interface. 

The above steps will be repeated in each round of coordination until the collaboratively trained model meets the requirements of the user, which can be specified by a combination of performance measures, including O-error, G-error, C-error, and R-cost, which will be discussed in detail in Section \ref{Section_theory}. 

During the inference phase, each agent will calculate the output embedding of the agent-part model based on the observed data samples and send the embeddings to the agent controller via the E-interface. The agent controller will output the final decisions based on the output of the shared-part model. 

Note that, since in the above algorithm, the weighting coefficients of $\gamma^i_m$ of each agent need to be pre-defined and assumed to be static during the entire training process, we refer to the above algorithm as the static-weighting algorithm. The model training and inference procedures are illustrated in Fig. \ref{Fig_StaticDynamicAlgorithms}(a) and (c), respectively.

\subsection{\blu{Bandwidth-Adaptive Compression Module Design}}

\blu{The MoPS framework proposed in the previous subsection allows for dynamic reallocation of computational overhead, thus distributing the burdens of multi-objective large model training and inference in accordance with the heterogeneous computational resource constraints of individual agents. In practice, however, MoPS is frequently bottlenecked by stringent bandwidth limitations between different agents and the agent controller. Such constraints necessitate a robust mechanism for data compression that does not compromise the integrity of the learned representations. To address this challenge, we propose an integrated, bandwidth-adaptive compression and decompression architecture designed as an "add-on" module for MoPS. This module enables each agent to compress intermediate embeddings in situ based on instantaneous local bandwidth availability, which are subsequently reconstructed at the agent controller before being fed into the shared-part model. }

\blu{Formally, we define the compression module at agent $i$ and decompression module at the agent controller as $C_{\psi_i}$ and $D_{\varphi_i}$, respectively, where $\psi_i$ and $\varphi_i$ are the parameters of the compression and decompression modules.  
We describe both modules in detail as follows: }


\begin{figure}
\centering
\includegraphics[width=1\linewidth]{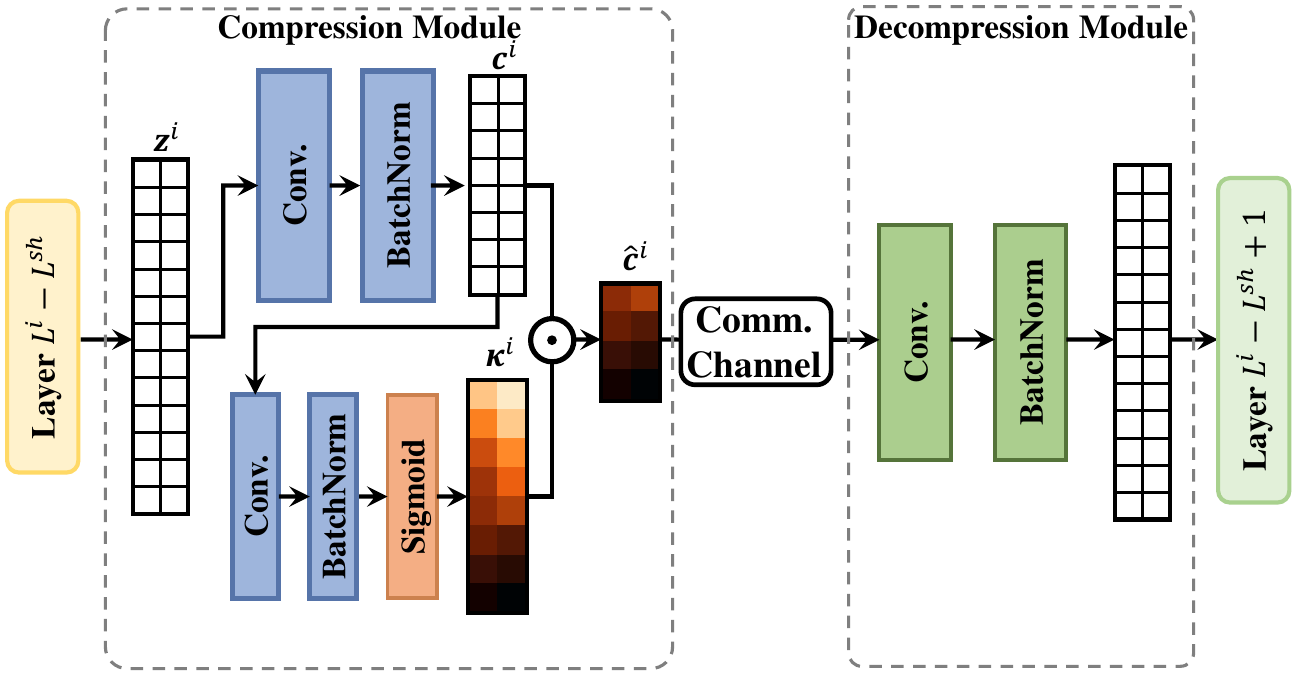}
\caption{\small \blu{Architectural framework of the bandwidth-adaptive compression and decompression module.}}
\label{Figure_compression_module}
\vspace{-0.6cm}
\end{figure}

\noindent{\blu{\bf Compression module $C_{\psi_i}$:}} \blu{This module can be considered as an importance filter that filters out the relatively less important information from the high-dimensional embedding output of the agent-specific part models and generates a low-dimensional representation consisting of the most important information to be sent over the bandwidth-limited channels. Suppose the final layer output embedding ${\boldsymbol z}^i$ of the $i$th agent is a feature vector of dimension $e^i$. The output of the compression module is given by ${\boldsymbol c}^i$ where ${\boldsymbol c}^i \in \mathbb{R}^{{\hat e}^i}$ and ${\hat e}^i \ll e^i$. In this paper, we consider a CNN-based compression module consisting of a convolutional layer for feature extraction and a batch normalization layer for stabilizing the model training and enhancing the feature expressiveness, as illustrated in Fig. \ref{Figure_compression_module}. The compression level of the module is controlled by an importance filter, which corresponds to vector ${\boldsymbol \kappa}^i \in \mathbb{R}^{\hat{e}_i}$ of size $\hat{e}_i$. Each element ${\kappa}^i_{l}$ in ${\boldsymbol \kappa}^i$ specifies the importance score of the $l$th dimension of the output embedding ${\boldsymbol c}^i$ for $l\in \{ 1, 2, \ldots, \hat{e}_i \}$. More specifically, by passing $\bc^i$ into a CNN network and a Sigmoid activation function, we can obtain the importance score ${\boldsymbol \kappa}^i =  C_{\psi_i} ({\boldsymbol z}^i)$ where $\psi_i$ is the parameters of the CNNs that maps $\bz^i$ to $\bc^i$ and that maps $\bc^i$ to ${\boldsymbol \kappa}^i$. We can then multiply the importance score with ${\boldsymbol c}^i$ to obtain a soft masked version of the feature vector $\hat{\boldsymbol c}^i = {{\boldsymbol \kappa}}^i \odot {\boldsymbol c}^i$ 
where $\odot$ denotes the element-wise product. In this way, the higher the value of ${\kappa}^i_{l}$, the higher the importance of the $l$th dimension of the output feature vector. The compression level can then be controlled by keeping only ${\hat e}^i$ dimensions of the output feature vector with the highest importance scores and ignoring the dimensions that have low importance scores, e.g., if the communication bandwidth between agent $i$ and the agent controller allows only $B_i$ dimensions of the feature vector to be transmitted throughout the channel, only the $B_i$ dimensions with the highest importance scores will be kept and the rest low-score dimensions will be discarded. We will discuss the training procedures of the importance filter when introducing the decompression module.} 



\noindent{\bf \blu{Decompression module $D_{\varphi_i}$:}} \blu{This module recovers the complete feature vector by first reconstructing the full $e^i$ dimensional feature vectors from the received compression vector and setting the rest of the dimensions to zeros, and then applying the reverse the procedures of the compression module, that is passing the reconstructed vectors to a convolutional layer and a batch normalization layer. In this paper, we consider an "add-on" solution of the compression and decompression modules, in which a loss function ${\cal L}^C_m \left(\psi_i, \varphi_i \right)$ for training the compression and decompression modules for each agent $i$ is included in the task-specific optimization loss function, so both the agent models and compression models are trained simultaneously. Accordingly, the 
Problem {\bf P1} can be re-defined as:}
\begin{eqnarray}
    \blu{ {\bf P3:}\;\; \min_{\bOmega_m} \hat{\cal L}_m (\bOmega_m) = \langle \hat{\cal L}^i_m (\bOmega_m, {\cal D}^i) \rangle_{i \in \{a, p, n\} },} 
\end{eqnarray}
\blu{where $\hat{\cal L}^i_m (\bOmega_m, {\cal D}^i)$ 
is the loss function for joint training of compression and decompression modules. In this paper, we consider the following loss function for the joint training: } 
\begin{eqnarray}
\blu{ \hat{\cal L}^i_m (\bOmega_m, {\cal D}^i) = {\cal L}^i_m (\bOmega_m) + \epsilon_i  
{\cal L}^{M}_m (
\psi_i, \varphi_i),  }
\end{eqnarray}
\blu{where the first term is the original optimization objective of agent $i$ and ${\cal L}^{M}_m (\psi_i, \varphi_i)$ is 
the importance filter, a soft-constraint function that penalizes the deviation between the total amount of extracted features and the target bandwidth $B_i$, both of which are given by}
\blu{ 
\begin{eqnarray}
{\cal L}^{M}_m (\psi_i, \varphi_i) &=& \exp {\left(\sum_{l = 1}^{\hat{e}_i} {\kappa}^i_l -  B_i\right). }
\end{eqnarray}
}


\blu{The above design of compression and decompression modules has the following advantages. First, by integrating the optimization objectives of the agents with those of the compression and decompression modules into a unified training objective, the above framework obviates the requirement for a separate development phase for compression and decompression modules, which significantly mitigates both the communication and computational overhead traditionally associated with the isolated training of these modules. Furthermore, as both modules are implemented based on a lightweight, CNN-based feature extractor with input and output layers being reconfigurable to align with original embedding dimensions and instantaneous bandwidth constraints, respectively, the associated costs for training and implementation remain negligible. Finally, the proposed compression and decompression modules are bandwidth adaptive and can utilize switchable layers that adjust the number of active neurons based on the current network environment. This allows a single trained model to operate across varying bandwidth limitations and network environments without retraining. As discussed later in the experimental results, since the compression module extracts the most important features relevant to the task-specific optimization objectives, the performance degradation caused by compression is limited. Note that the proposed compression and decompression modules are developed as ``plug-and-play" add-ons to the MoPS framework, and can be skipped when the bandwidth limitations or latency requirements are not stringent. }

\section{Dynamic Weighting Algorithm for Conflict-resolving}
\label{Section_DynamicWeight}



As mentioned earlier, the diverse objectives and data modalities of different subtasks performed by the agents often lead to inherent conflicts. These conflicts arise because the optimization trajectories for different subtasks are not always aligned and can, at times, be opposed, as illustrated in Fig. \ref{Fig_IntroExample}. This conflict is mainly reflected in the shared-part model updated in (\ref{eq_shared_model_update}) in which the gradients output of different agent-specific part models are aggregation with each other to update the shared-part model. 

Motivated by the fact that the weight of each objective's gradient directly influences the model's overall optimization direction, in this section, we propose a dynamic weighting algorithm in which the weighting coefficients of different agents can be dynamically optimized throughout the training process to resolve conflicts between different objectives. In particular, during each round $t$ of model training, each agent $i$ needs to calculate and upload two additional embedding-label pairs, calculated based on two independent data samples denoted by $d^i_{t(1)}$ and $d^i_{t(2)}$, 
to the agent controller. The agent controller can then update the weighting coefficient $\gamma^i_{m,t}$ of each agent $i$ based on the 
gradients calculated based on these two data samples as follows: 
%
\begin{flalign}
    \gamma_{m,t}^i = \gamma_{m,t-1}^i - &\eta_t   \nabla_{\btomega^{sh}_{m, t-1}} \ell^i_m (\langle \btomega^{sh}_{m, t-1}, \btomega^i_{m, t} \rangle, d^i_{t(1)} )^{\top}  \nonumber\\
    & \cdot \nabla_{\btomega^{sh}_{m, t-1}} \ell^i_m(\langle \btomega^{sh}_{m, t-1}, \btomega^i_{m, t} \rangle, d^i_{t(2)}),
\label{eq_gamma_update} 
\end{flalign}
where $\eta_t$ is the step size for weight update. The above weighting coefficient will be substituted into (\ref{eq_shared_model_update}) to update the shared-part model. The rest of the algorithm follows the same line as the static-weighting-based solution introduced in the previous section. The inference procedures are the same as the static-weighting algorithm. 

We refer to the above algorithm as the dynamic-weighting algorithm. 
The model training and inference procedures are illustrated in Fig. \ref{Fig_StaticDynamicAlgorithms}(b) and (c), respectively.

\section{Theoretical Analysis}\label{Section_theory}


In this section, we present the theoretical results of O-, G-, and C-errors for both static- and dynamic-weighting algorithms proposed in the previous section. We summarize all the theoretical bounds derived in this section in Table \ref{comparison}. 


      

\begin{table}[t]
    \centering
    \begin{threeparttable}
      \caption{Theoretical bounds of O-, G-, and C-errors for static- and dynamic-weighting algorithms, where 
      $\cO(\cdot)$ denotes the asymptotic upper bound, and $\Theta(\cdot)$ the asymptotic tight bound.}\label{comparison}
      \begin{tabular}{c|cc}
        \toprule
        Algorithm & Static-weighting & Dynamic-weighting \\
        \midrule
        O-Error & $\cO((\beta T)^{-\frac{1}{2}} + \beta^{\frac{1}{2}})$ & $\cO((\beta T)^{-\frac{1}{2}} + \beta^{\frac{1}{2}} + \eta^{\frac{1}{2}})$ \\
        G-Error & $\cO(T^{\frac{1}{2}}D^{-\frac{1}{2}})$ & $\cO(T^{\frac{1}{2}}D^{-\frac{1}{2}})$ \\
        C-Error & $\Theta (1)$ & $\cO((\eta T)^{-1} + \beta^{\frac{1}{2}}\eta^{-\frac{1}{2}} + \eta)$ \\
        \bottomrule
      \end{tabular}  
    \end{threeparttable}
    \vspace{-0.6cm}
\end{table}

\subsection{O-Error}
We can prove the following theoretical bound on the O-error for the static-weighting algorithm. 

\begin{theorem}[O-error bound of static-weighting algorithm]\label{Theorem_UpperBoundOError_without CA}
    Suppose the following assumptions hold: 
    \begin{assumption}\label{smoothness}
    For any $i \in \{ a,p,n \}$, $\nabla \ell^i_m (\bOmega_m, d^i)$ is $\mu_g$-Lipschitz continuous for any data sample, i.e., $\| \nabla \ell^i_m (\bOmega_m, d^i) - \nabla \ell^i_m (\bOmega_m', d^i) \| \le \mu_g \| \bOmega_m - \bOmega_m' \| $.
    \end{assumption}
    \begin{assumption}\label{continuous}
    For any $i \in \{ a,p,n \}$, $ \ell^i_m (\bOmega_m, d^i)$ is $\mu_l$-Lipschitz continuous for any data sample, i.e., $\| \ell^i_m (\bOmega_m, d^i) - \ell^i_m (\bOmega_m', d^i) \| \le \mu_{l} \| \bOmega_m - \bOmega_m' \| $.
    \end{assumption}
    \begin{assumption}\label{initial model}
    For any $\gamma^i_m, i \in \{a, p, n\}$, the initialized model $\bOmega_{m,0}$ satisfies that 
    \begin{equation}
    \mathbb{E}[\!\!\!\! \sum\limits_{i\in \{a, p, n\}}\!\!\!\!\! \gamma^i_m \ell^i_m (\bOmega_{m,0}, d^i)] \!-\! \min\limits_{\bOmega_m} \mathbb{E}[ \!\!\!\! \sum\limits_{i\in \{a, p, n\}} \!\!\!\!\! \gamma^i_m \ell^i_m (\bOmega_m, d^i)] \!\le\! c_{I}. \nonumber
    \end{equation}
    \end{assumption}
    Suppose the model $\{ \bOmega_{m,t} \}_{t=1}^T$ is trained by the static-weighting algorithm with $\beta_t = \beta \le \frac{1}{2 \mu_g}$. We can prove the following result of O-error: 
    \begin{eqnarray}
        \frac{1}{T} \sum_{t=0}^{T-1}\mathbb{E} [\cE_{O,t}] \le \sqrt{\frac{c_I}{\beta T}} + \sqrt{\frac{\beta \mu_g \mu_l^2}{2}}.
        \label{eq_OErrorTheoryBound}
    \end{eqnarray}
\end{theorem}

\begin{IEEEproof}
    See Appendix \ref{proof of O-error without CA}.
\end{IEEEproof}

Assumptions 1)-3) are commonly introduced for most theoretical analysis of SGD-based AI model training processes and are quite reasonable in many practical scenarios \cite{FernandoSLCMC23, chen2023three}. In particular, these assumptions essentially state that the loss function and its gradient are well-conditioned and the training process starts within a reasonable range. More specifically, Assumptions 1) and 2) imply that the rate of change of the function is bounded, preventing sudden, massive jumps in the loss value for small changes in the model's parameters, i.e., any small adjustments to the model's weights lead to predictable and manageable changes in the loss. These assumptions hold for many popular loss functions such as the mean squared error (MSE) and cross-entropy. Assumption 3) is also a very mild assumption, as the goal of model training is always to achieve a significant, but bounded, improvement from a randomly selected initial state. 

The upper bound in (\ref{eq_OErrorTheoryBound}) quantifies the joint impact of three key variables on the O-error, including the initially selected model parameters $\bOmega_{m,0}$, the number of coordination rounds $T$ among agents, and the learning rate $\beta$. 
For example, if we choose $\beta = \Theta ( T^{-\frac{1}{2}} )$, the O-error converges at the optimal rate of $\mathcal{O} ( T^{-\frac{1}{4}} )$.

We can similarly prove the following theoretical bound on the O-error of the dynamic-weighting algorithm.

\begin{corollary}[O-error bound of dynamic-weighting algorithm]\label{Theorem_UpperBoundOError}
    Suppose assumptions \ref{smoothness}-\ref{initial model} hold. Suppose model $\{ \bOmega_{m,t} \}_{t=1}^T$ is trained by by the dynamic-weighting algorithm 
    with $\beta_t = \beta \le \frac{1}{2 \mu_g}, \eta_t = \eta$. We can prove the following bound: 
    \begin{eqnarray}
        \frac{1}{T} \sum_{t=0}^{T-1}\mathbb{E} [\cE_{O,t}] \le \sqrt{\frac{c_I}{\beta T}} + 3\sqrt{\frac{\eta \mu_l^4}{2}} + \sqrt{\frac{\beta \mu_g \mu_l^2}{2}}.
    \end{eqnarray}
\end{corollary}

\begin{IEEEproof}
    See Appendix \ref{proof of O-error}.
\end{IEEEproof}

From Corollary \ref{Theorem_UpperBoundOError}, we can observe that the dynamic-weighting algorithm results in $3\sqrt{\frac{\eta \mu_l^4}{2}}$ increase on the upper bound of O-error, compared to the static-weighting algorithm. This is because, in each round of coordination, the dynamic-weighting algorithm requires extra steps to adjust the optimization directions of each agent's loss by updating its weighting coefficient, i.e., in equation (\ref{eq_gamma_update}), for resolving the multi-objective conflicts among agents. In other words, the increase in the O-error of the dynamic-weighting algorithm can be considered as the cost introduced to resolving conflicts.  
We can also observe that, by carefully choosing $\beta = \Theta ( T^{-\frac{1}{2}} )$ and $\eta = \Theta ( T^{-\frac{1}{2}} )$, the dynamic-weighting algorithm can achieve the same order of convergence $\cO ( T^{-\frac{1}{4}} )$ as the static-weighting algorithm.

\subsection{G-Error}

Before introducing the theoretical bounds of G-error, let us first define the concept of uniform stability as follows. 
\begin{definition}[Uniform stability \cite{chen2023three}]\label{def_of_stability}
    A randomized algorithm $\cA$ is $\epsilon$-uniformly stable if for any given datasets ${\cD}$ and ${\cD}'$ such that ${\cD}$ and ${\cD}'$ differ in at most one data sample, the following inequality holds:
    \begin{eqnarray}
        \sup_{d^i \in {\cal D}^i} \mathbb{E}_{\cA} [\| \nabla \ell^i_m (\cA(\cD), d^i) \!-\! \nabla \ell^i_m (\cA({\cD'}), d^i) \|^2 ] \!\le \epsilon^2. \nonumber
    \end{eqnarray}
\end{definition}

Uniform stability is an important property for any stable and trainable algorithms, as it suggests that any small change, such as the removal or alteration of a single data sample, results in only a negligible change to the model's output. Most SGD-based algorithms, especially those with regularization, inherently satisfy this property as they always require the result of the learning algorithms, i.e., the learned models, to be robust and less sensitive to individual data samples. 

We can prove the following bound of G-error for the static-weighting algorithm. 


\begin{theorem}[G-error bound of static-weighting algorithm]\label{Theorem_UpperBoundGError_without CA}
    Suppose the following assumption holds:
    \begin{assumption}\label{Bounded_Gradient} 
    The summation of gradients of all agents is upper bounded, i.e, $\mathbb{E}[\| \sum_{i\in \{a, p, n\}} \gamma^i_m \nabla \cL^i_m (\bOmega_m, \cD^i) \|^2]$ $\le G^2$ where $G$ is any given constant.
    \end{assumption}
    Then, for any uniformly stable static-weighting algorithms $\cA_m$, the G-error is upper bounded by
    \begin{eqnarray}
        \mathbb{E}[\cE_G] \le 8G \sqrt{\frac{T}{D}} + \sqrt{\frac{V}{D}},
        \label{eq_Gerror_staticalg_bound}
    \end{eqnarray}
    where $D$ is the total number of data samples in the datasets of all the agents, i.e., $D = \sum_{i\in \{a, p, n\}} |{\cD^i}|$, and $V$ is the variance of gradients, given by 
    \begin{eqnarray}
        V = \mathbb{E}[ \| \!\!\!\!\!\! \sum_{i\in\{a, p, n\}} \!\!\!\!\!\! \gamma^i_m (\nabla \ell^i_m (\cA_m(\cD), d^i) \!-\! \mathbb{E}[\nabla \ell^i_m (\cA_m(\cD), d^i)]) \|^2 ]. \nonumber
    \end{eqnarray}
    where $\cA_m (\cD)$ is the model learned by algorithm $\cA_m$ based on dataset $\cD$.  
\end{theorem}

\begin{IEEEproof}
    See Appendix \ref{proof of G-error without CA}.
\end{IEEEproof}

Inequality (\ref{eq_Gerror_staticalg_bound}) quantifies the impact of the number of coordination rounds $T$ and the size of the dataset $D$ on the G-error. We can observe that the generalization performance of algorithms generally increases with the size of the dataset and decreases with the total number of coordination rounds. This is because increasing $T$ beyond a certain point can lead to overfitting, causing the model to memorize the specific non-generalizable features of the training data, thereby degrading its performance on unseen data that exhibit different features than the training dataset. On the other hand, for a fixed number of coordination rounds, increasing the size of the training dataset improves generalization performance by exposing the model to a wider variety of data, which reduces the influence of any single data point and encourages the learning of more robust, universal features.


We can similarly derive the following theoretical bounds of the dynamic-weighting algorithm. 

\begin{corollary}[G-error bound of dynamic-weighting algorithm]
\label{Theorem_UpperBoundGError}
Suppose the Assumption \ref{Bounded_Gradient} holds. Then, the G-error of the dynamic-weighting algorithm is upper-bounded by
\begin{equation}
    \mathbb{E}[\cE_G] \le 8G \sqrt{\frac{T}{D}} + \sqrt{\frac{V}{D}}.
    \label{eq_Gerror_dynamicweighting}
\end{equation}
\end{corollary}

\begin{IEEEproof}
See Appendix \ref{proof of G-error}.
\end{IEEEproof}

We can observe that the theoretical bound of the dynamic-weighting algorithm is exactly the same as that of the static-weighting algorithm. This means that the extra steps for optimizing the weighting coefficients to resolve the conflicts among different agents cannot affect the generalization performance of the algorithm. 


\subsection{C-Error}

We can prove the following result for the C-error of the static-weighting algorithm. 

\begin{theorem}[C-error bound of static-weighting algorithm]\label{Theorem_UpperBoundCError_without CA}
    Suppose Assumption \ref{smoothness} holds. Then, there exists a set of static weighting $\langle \gamma^a_m, \gamma^p_m, \gamma^n_m \rangle$ such that 
    \begin{eqnarray}
        \mathbb{E} [\cE_C] = \Theta (1), 
        \label{eq_Cerrorbound_staticweighting}
    \end{eqnarray}
    where $\Theta (\cdot)$ is the asymptotic tight bound and $\Theta (1)$ means that the C-error remains a constant as other variables, such as the number of coordination rounds and the size of the dataset, change. 
\end{theorem}

\begin{IEEEproof}
    See Appendix \ref{proof of C-error without CA}.
\end{IEEEproof}


The above result suggests that the performance degradation caused by conflicting objectives of different agents is fixed with any given weighting coefficients and cannot be alleviated during the model training. 

We can prove the following upper bound on the C-error of the dynamic-weighting algorithm.

\begin{theorem}[C-error bound of dynamic-weighting]
\label{Theorem_UpperBoundCError}
Suppose the Assumptions \ref{smoothness} and \ref{continuous} hold. If the learning rate satisfies $\beta_t \le \beta$ and $\eta_t \le \eta$, the following upper bound on the C-error holds for the dynamic-weighting algorithm:
    \begin{eqnarray}
        \frac{1}{T} \sum_{t=1}^{T-1} \mathbb{E}[\cE_{C,t}] \le \frac{4}{\eta T} + 6 \sqrt{3 \mu_g \mu_l^2\frac{\beta}{\eta}} + 3\eta \mu_l^4.
        \label{eq_Cerror_dynamicweighting}
    \end{eqnarray}
\end{theorem}

\begin{IEEEproof}
See Appendix \ref{proof of C-error}.
\end{IEEEproof}

Inequality (\ref{eq_Cerror_dynamicweighting}) quantifies the impact of the number of coordination rounds and learning rate on the C-error of the dynamic-weighting algorithm. We can observe that the C-error reduces with the number of coordination rounds. Also, if we set $\beta = \Theta ( T^{-\frac{3}{4}} )$ and $\gamma = \Theta ( T^{-\frac{1}{4}} )$, the C-error can converge at the fastest rate of 
$\cO (T^{-\frac{1}{4}})$. 


\subsection{Three-way Tradeoff among O-error, G-error, and C-error}

From the results in Theorems \ref{Theorem_UpperBoundOError} to \ref{Theorem_UpperBoundGError}, we can observe that the O-, G-, and C-errors are closely related to each other. In particular, from Theorems \ref{Theorem_UpperBoundOError} and \ref{Theorem_UpperBoundGError}, we can observe a fundamental tradeoff between O-error and G-error with any given training dataset, i.e., as the number of coordination rounds $T$ increases, the O-error (or G-error) decreases (or increases). This observation is consistent with previous results about the generalization and optimization tradeoff, or bias-variance tradeoff, which suggests that any model that fits a given set of training data too perfectly will result in poor performance on new unseen data samples \cite{yang2020rethinking}.    

From Theorem \ref{Theorem_UpperBoundCError}, we can observe that the C-error in the dynamic-weighting algorithm is also closely linked to the O-error. As mentioned earlier, O-error reduces with the highest rate when the learning rate can be set to $\beta = \Theta ( T^{-\frac{1}{2}} )$ and $\eta = \Theta ( T^{-\frac{1}{2}} )$, while the highest convergence rate of C-error requires the learning rates to satisfy $\beta = \Theta ( T^{-\frac{3}{4}} )$ and $\gamma = \Theta ( T^{-\frac{1}{4}} )$. This means that the learning rates need to be carefully decided to balance the C-error and O-error with fixed numbers of coordination rounds and dataset sizes.

\section{Prototype and Experimental Setup}
\label{Section_Prototype}

\begin{figure}[t]
\centering
\includegraphics[width=1\linewidth]{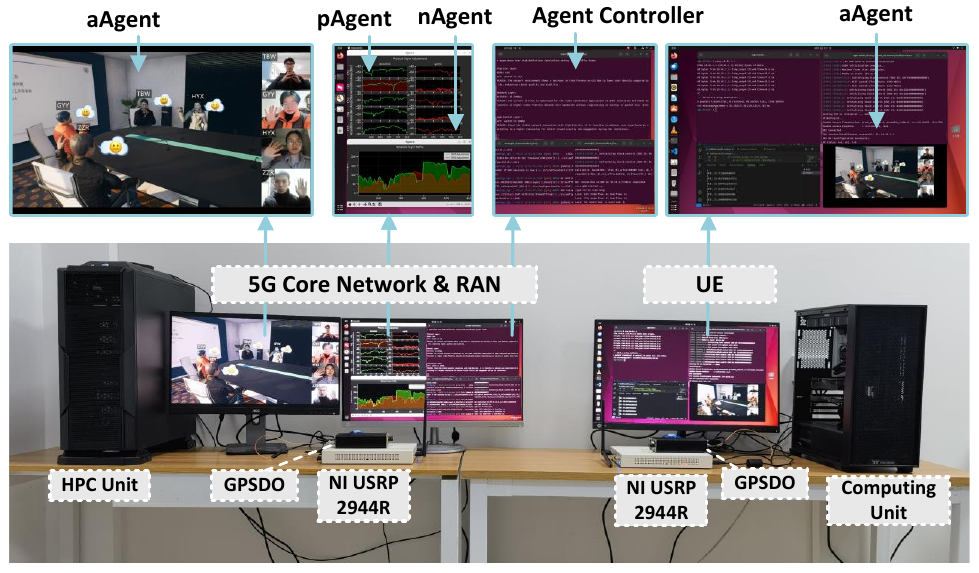}
\vspace{-0.7cm}
\caption{SANet prototype.}
\label{Figure_Prototype}
\vspace{-0.6cm}
\end{figure}

We developed a SANet prototype based on an open-source RAN and softwareized 5G core network, as shown in Fig. \ref{Figure_Prototype}. In the rest of this section, we first introduce the details of our prototype platform and then present the experimental setup and datasets used in our experiments. Detailed experimental results are presented in the next section. 


\noindent{\bf Prototype:} The SANet prototype hardware consists of three major components: the gNodeB (gNB), the user equipment (UE), and the 5G core (5GC) network, as illustrated in Fig. \ref{Figure_Prototype}. Specifically, the gNB--which handles all Radio Access Network (RAN) functions--is instantiated using the srsRAN open-source software suite. 
This software-defined gNB connects to the UE via a Universal Software Radio Peripheral (USRP), NI USRP 2944R, which provides the essential hardware interface for over-the-air communication. 
The 5GC is implemented entirely using the Open5GS project installed on a workstation equipped with an Intel(R) Core(TM) i9-13900K CPU@5.8GHz, 128GB of DDR5 RAM, and an NVIDIA GeForce RTX 4090 GPU.  

\noindent{\bf Agent Controller:} We installed Qwen-7B, 
an open-source LLM, for reasoning and OpenManus 
for task separation at the agent controller deployed at the 5GC. We consider two types of events that the agent controller can cognize to trigger the task separation and multi-agent collaborative learning: (1) key prompt detection and (2) traffic and environmental dynamics. In the former case, a set of key prompts related to the specific task demands of the user is predefined and installed at the UE and agent controller. The UE will keep track of the input of the user and, once a key prompt is detected, it will convert it into embeddings to be sent to the agent controller.   
In the second case, if the current traffic and environmental dynamics result in performance degradation exceeding a maximum tolerable level, i.e., O-error, G-error, and/or C-error exceeds a threshold, the agent controller will initiate the model updating process.

\noindent{\bf Agents:} We developed an immersive 3D video conference application based on the SF3D 2D-to-3D video construction model and the Unity 3D rendering engine. This system consists of client applications installed at the UEs and user devices connected to the Internet to stream 2D videos generated by the users and an application server installed at the 5GC to convert the users' 2D videos into corresponding 3D avatars in a virtual conference. The 3D video generated by the application server will be streamed back to the users to offer an immersive 3D conference experience. The agent controller coordinates aAgent, nAgent, and pAgent according to different users' semantic demands. For example, the prompts relevant to users' satisfaction with the streamed video quality, e.g., current video resolution, can be predefined, and if it detects that a user is unsatisfied with the current resolution of the streamed 3D video, the agent controller will separate the task demand into different requirements at the application layer, physical layer, and network layer and then reason about different steps and model separation and sharing schemes to meet the user's demand. In particular, if the detected semantic demand of the user is to increase the resolution of the streamed 3D video, the agent controller will select an aAgent associated with the demand requesting user to evaluate the required data rates generated by the application server for supporting a higher resolution video streaming, as well as a pAgent to estimate the required physical layer spectrum and the CSI and an nAgent to analyze the required network bandwidth to support the required data rates between the application server and client. We develop three different agents: aAgent, pAgent, and nAgent, all of which consist of Transformer-based time-series prediction models and action functions to select the parameters associated with application layer, physical layer, and network layer, respectively, e.g., aAgent, pAgent, and nAgent consist of models to predict the future application-generated data rates, CSIs of different subcarriers between the UE and gNB, and network bandwidths to support the 5G compatible data transmissions, as well as action functions to select the appropriate video streaming resolution, wireless channels, and network bandwidth according to model outputs, respectively. We consider model partition and sharing among the agent controller, aAgent, nAgent, and pAgent, in which the shared-part model parameters associated with the attention layers shared among different Transformer models of aAgent, nAgent, and pAgent are deployed at the agent controller, and the agent-specific model parameters are modality-specific time-series embedding, position encoding module, and attention layers trained by each individual agent based on its local set and modality of data. 



\noindent{\bf \blu{Dataset:}} \blu{Utilizing the developed SANet prototype platform, we constructed a comprehensive multi-channel, multi-modal dataset designed for aAgents, nAgents, and pAgents deployed across the physical, network, and application layers, respectively. Specifically, for pAgents, the Channel State Information (CSI) sequences were recorded across five distinct frequency bands: n1 ($2110\text{--}2170\text{ MHz}$), n2 ($1930\text{--}1990\text{ MHz}$), n3 ($1805\text{--}1880\text{ MHz}$), n5 ($869\text{--}894\text{ MHz}$), and n7 ($2620\text{--}2690\text{ MHz}$), totaling 7,200 seconds of data per band. To characterize nAgents in the network layer, we measured the maximum achievable bandwidth between the UE and the 5G core via the iPerf3 utility. Finally, the dataset associated with aAgents in the application-layer was obtained by recording the real-time data stream generated by our developed immersive 3D conference application. }
\begin{figure*}[t]
    \begin{minipage}[t]{0.19\linewidth}
    \includegraphics[width=1\textwidth]{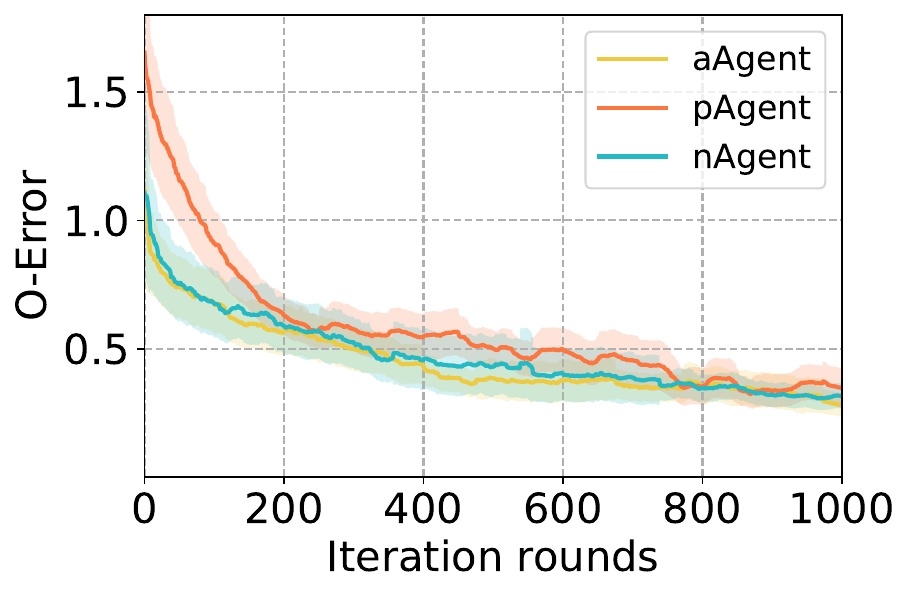}
    \captionsetup{labelformat=empty}
    \vspace{-0.7cm}
    \caption*{(a)}
    \end{minipage}
    \hfill
    \begin{minipage}[t]{0.19\linewidth}
    \includegraphics[width=1\textwidth]{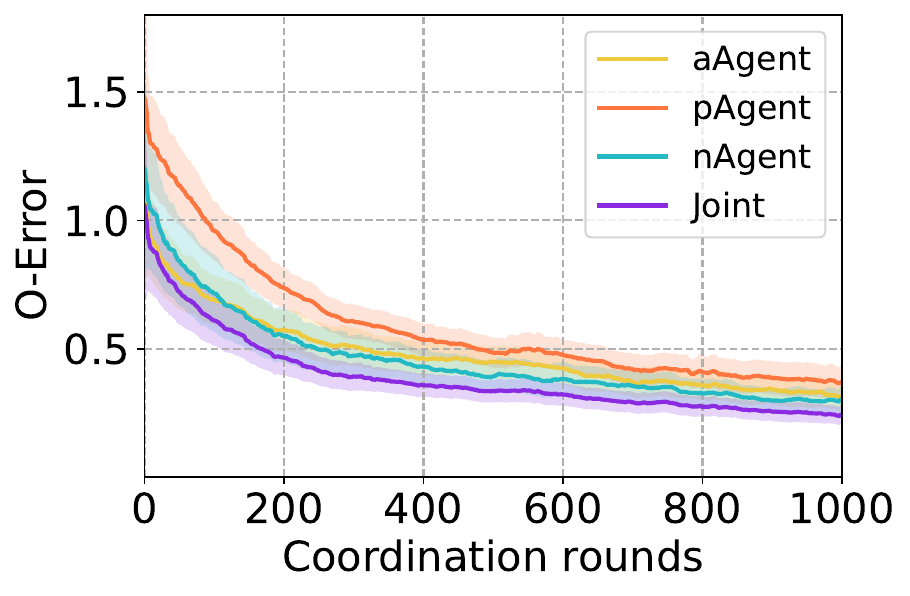}
    \captionsetup{labelformat=empty}
    \vspace{-0.7cm}
    \caption*{(b)}
    \end{minipage}
    \hfill
    \begin{minipage}[t]{0.19\linewidth}
    \includegraphics[width=1\textwidth]{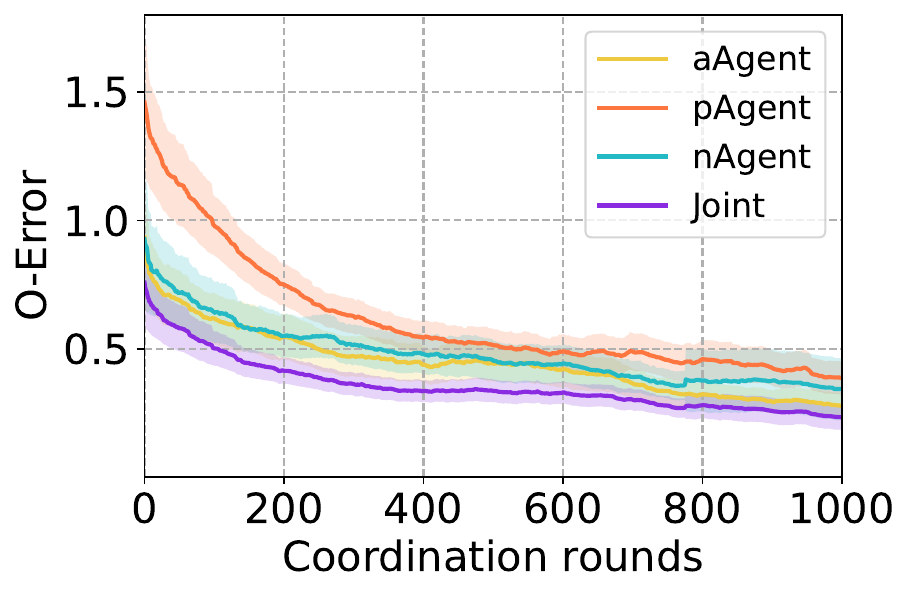}
    \captionsetup{labelformat=empty}
    \vspace{-0.7cm}
    \caption*{(c)}
    \end{minipage}
    \hfill
    \begin{minipage}[t]{0.19\linewidth}
    \includegraphics[width=1\textwidth]{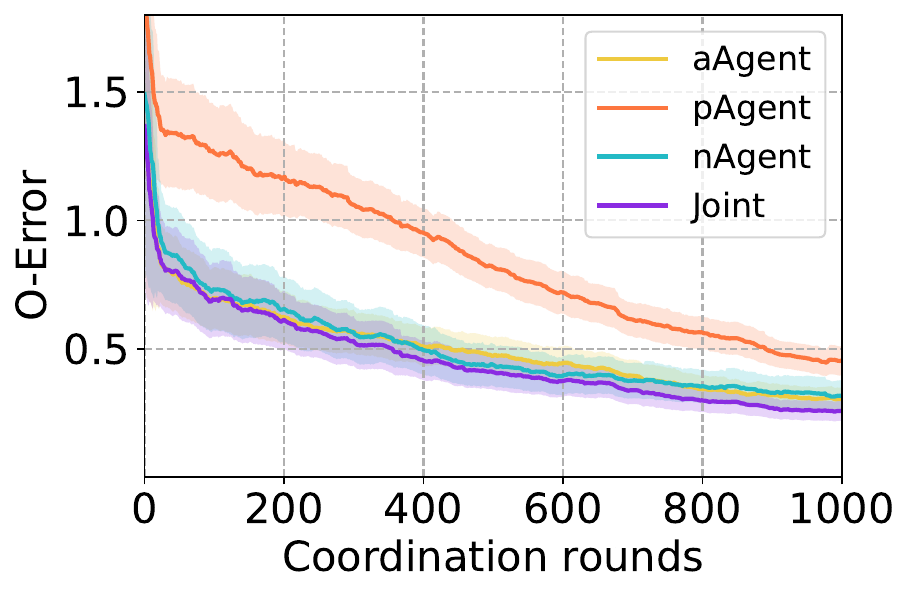}
    \captionsetup{labelformat=empty}
    \vspace{-0.7cm}
    \caption*{(d)}
    \end{minipage}
    \hfill
    \begin{minipage}[t]{0.19\linewidth}
    \includegraphics[width=1\textwidth]{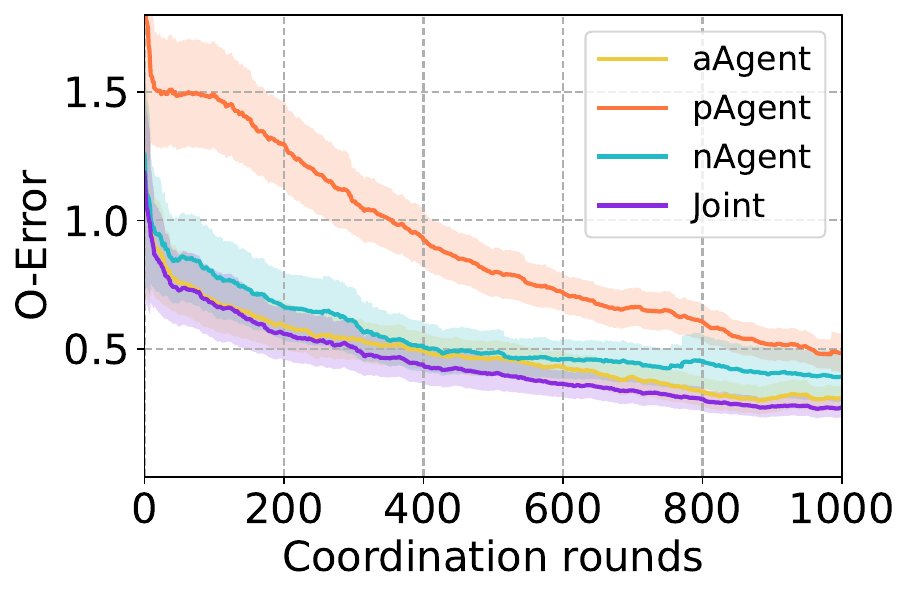}
    \captionsetup{labelformat=empty}
    \vspace{-0.7cm}
    \caption*{(e)}
    \end{minipage}
    \vspace{-0.3cm}
    \caption{\small O-error of different agents using MoPS with (a) Scheme 1, (b) Scheme 2 and (c) Scheme 3 trained by the static-weighting algorithm and (d) Scheme 2 and (e) Scheme 3 trained by the dynamic-weighting algorithm.}
    \label{Figure_OerrorMoPSSchemes}
    \vspace{-0.4cm}
\end{figure*}

\begin{figure*}[t]
    \begin{minipage}[t]{0.19\linewidth}
    \includegraphics[width=1\textwidth]{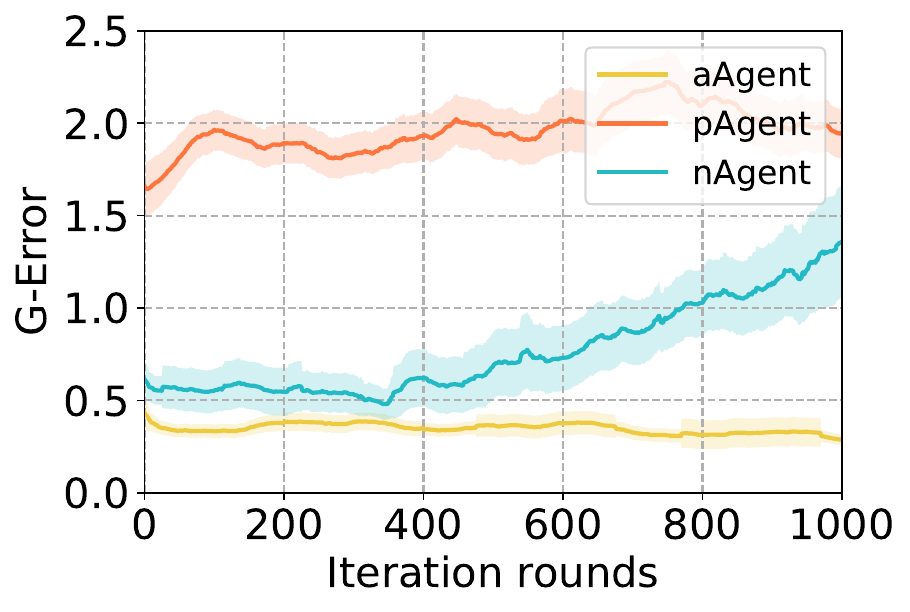}
    \captionsetup{labelformat=empty}
    \vspace{-0.7cm}
    \caption*{(a)}
    \end{minipage}
    \hfill
    \begin{minipage}[t]{0.19\linewidth}
    \includegraphics[width=1\textwidth]{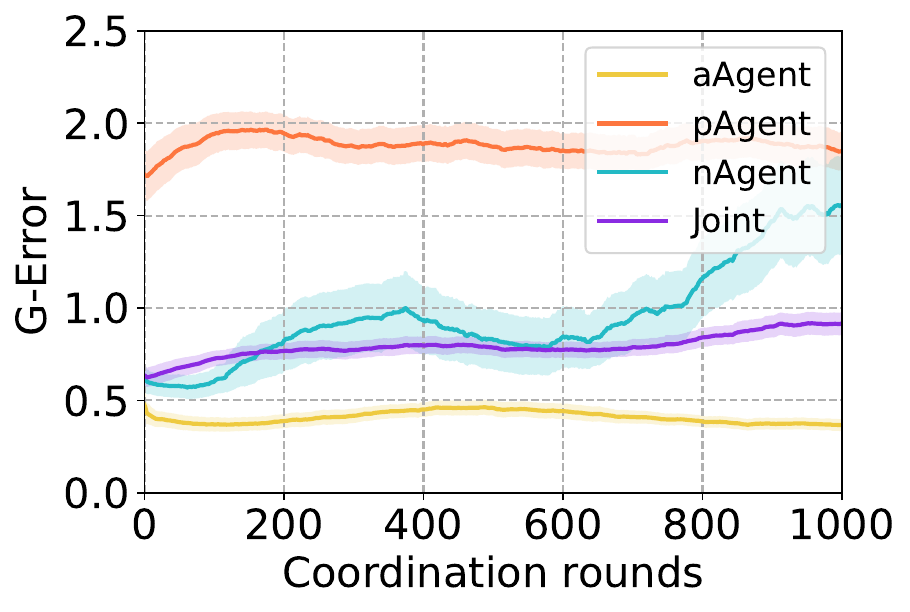}
    \captionsetup{labelformat=empty}
    \vspace{-0.7cm}
    \caption*{(b)}
    \end{minipage}
    \hfill
    \begin{minipage}[t]{0.19\linewidth}
    \includegraphics[width=1\textwidth]{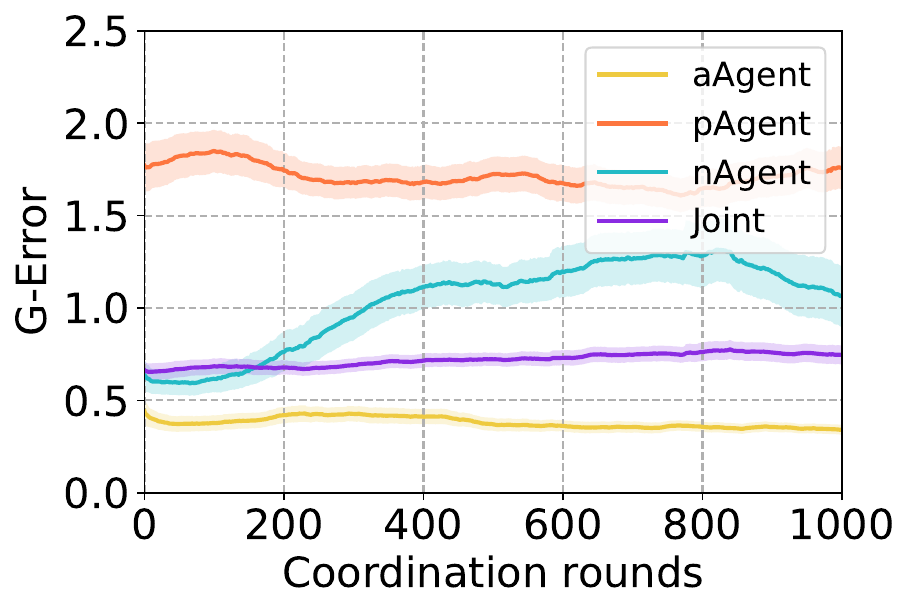}
    \captionsetup{labelformat=empty}
    \vspace{-0.7cm}
    \caption*{(c)}
    \end{minipage}
    \hfill
    \begin{minipage}[t]{0.19\linewidth}
    \includegraphics[width=1\textwidth]{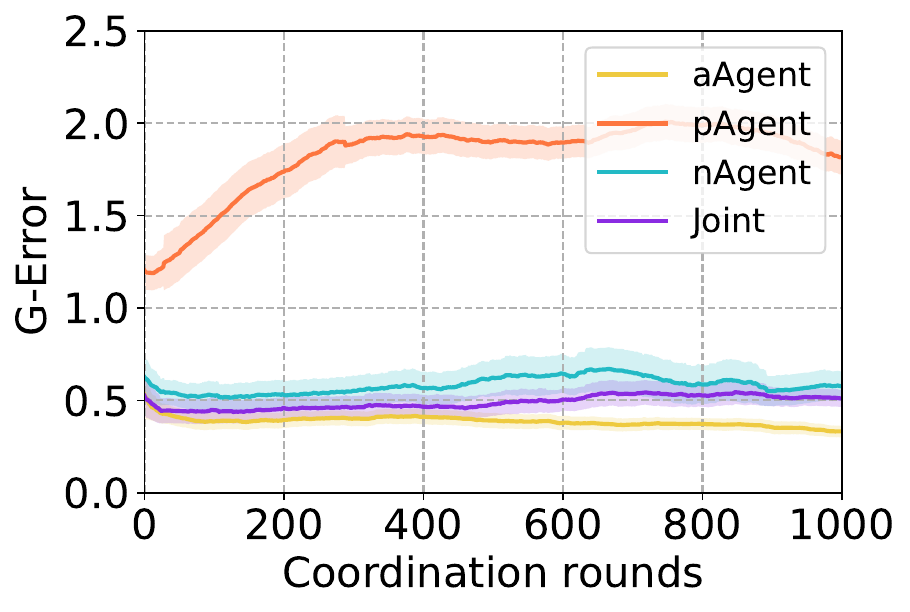}
    \captionsetup{labelformat=empty}
    \vspace{-0.7cm}
    \caption*{(d)}
    \end{minipage}
    \hfill
    \begin{minipage}[t]{0.19\linewidth}
    \includegraphics[width=1\textwidth]{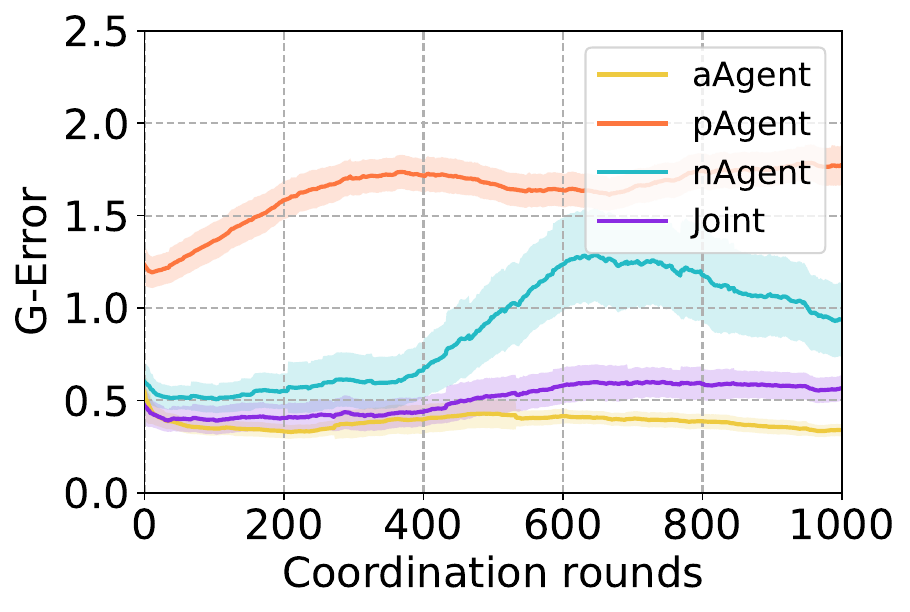}
    \captionsetup{labelformat=empty}
    \vspace{-0.7cm}
    \caption*{(e)}
    \end{minipage}
    \vspace{-0.3cm}
    \caption{\small G-error of different agents using MoPS with (a) Scheme 1, (b) Scheme 2 and (c) Scheme 3 trained by the static-weighting algorithm and (d) Scheme 2 and (e) Scheme 3 trained by the dynamic-weighting algorithm.}
    \label{Figure_GerrorMoPSSchemes}
    \vspace{-0.6cm}
\end{figure*}

\section{Experimental Results}\label{Section_experimental_results}

\subsection{Multi-agent Model Partition and Sharing}
To evaluate the impact of different model partition and sharing schemes on the optimization and generalization performance of AgentNet, we compare the following different partition and sharing schemes and evaluate their performance based on the metrics introduced in Section \ref{Section_OptimObjectives}: 

    \noindent{\bf Scheme 1 (No sharing):} Each agent trains an independent model based on its local data, and the agent controller does not host any model parameters.
    
    \noindent{\bf Scheme 2 (Partial sharing I):} Each agent trains a modality-specific time-series embedding and position encoding module, and the agent controller hosts all the Transformer layers (including 6 encoder/decoder layers each with 8 self-attention heads) shared by all agents. 
    
    \noindent{\bf Scheme 3 (Partial sharing II):} Each agent trains a modality-specific time-series embedding and position encoding module and the first 6 encoder layers. The agent controller hosts the remaining 6 decoder layers that are shared among all agents. 

We present the O-error and G-error of all the above schemes trained based on static-weighting and dynamic-weighting algorithms under different numbers of coordination rounds in Fig. \ref{Figure_OerrorMoPSSchemes} and \ref{Figure_GerrorMoPSSchemes}, respectively. We can observe that, as the number of coordination rounds increases, different agents' O-errors reduce at different rates, especially when trained by the static-weighting algorithm. Disparities in the convergence rates of collaboratively trained agents lead to the divergence, i.e., conflicts, in their respective optimization trajectories, which impede the overall performance of the AgentNet.
The G-errors of pAgent and nAgent increase with the number of coordination rounds in both static-weighting and dynamic-weighting algorithms. This aligns with our derived theoretical results in Section \ref{Section_theory}, which suggests that increasing the coordination round numbers for given datasets will result in improved model fitting and degraded generalization performance for new environments. 
Note that in Fig. \ref{Figure_OerrorMoPSSchemes} and \ref{Figure_GerrorMoPSSchemes}, the pAgent's G-error is generally the largest among all the agents, while its O-error also reduces at the slowest rate and remains in a relatively fixed value when the number of coordination rounds continues to increase. This is because the model training and testing datasets are collected from the real 5G physical channel, and generally speaking, the wireless channel exhibits much higher randomness and dynamics compared to the network and application layers. Therefore, it is generally more challenging to construct a time-series prediction model that achieves a relatively good optimization and generalization performance for wireless CSI signals.

An interesting observation in Fig. \ref{Figure_OerrorMoPSSchemes} and \ref{Figure_GerrorMoPSSchemes} is that allowing partial model sharing among different agents not only enhance the generalization performance, but also improve the optimization performance for each individual agent, e.g., both partial sharing schemes (Schemes 2 and 3) have lower O-error for each agent, compared to the no sharing scheme (Scheme 1) in which each agent trains its local model independently, i.e., Schemes 2 and 3 based on our proposed MoPS provide around 17.61\% and 13.87\% reduction per agent on O-error and 23.57\% and 37.62\% reduction on G-error, compared to Scheme 1. In particular, different agents train their models based on different data modalities, and also data samples collected from the physical layer, network layer, and application layer are often assumed to be independent. Our results suggest that even these seemly independent environments may still share some common features. This is aligned with the surprising observation in statistical decision theory, the Stein's paradox \cite{stein1956inadmissibility}, suggesting that it is better to simultaneously estimate three or more independent parameters, instead of estimating these parameters separately. 
The above experimental results further justify the unique advantage of our proposed MoPS for supporting flexible and scalable deployment of large models in resource-limited agents. 

\begin{figure*}[t]
    \begin{minipage}[t]{0.16\linewidth}
    \includegraphics[width=1\textwidth]{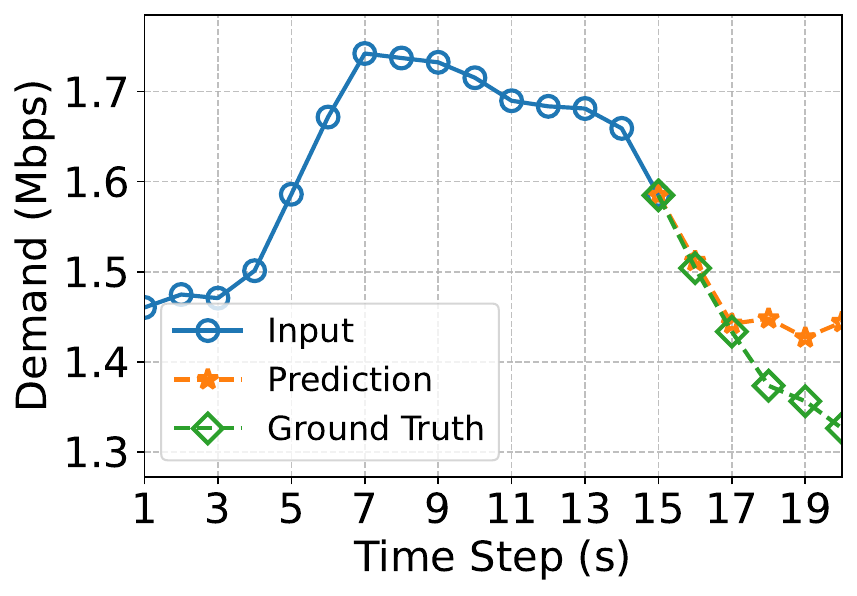}
    \captionsetup{labelformat=empty}
    \vspace{-0.7cm}
    \caption*{(a)}
    \end{minipage}
    \hfill
    \begin{minipage}[t]{0.16\linewidth}
    \includegraphics[width=1\textwidth]{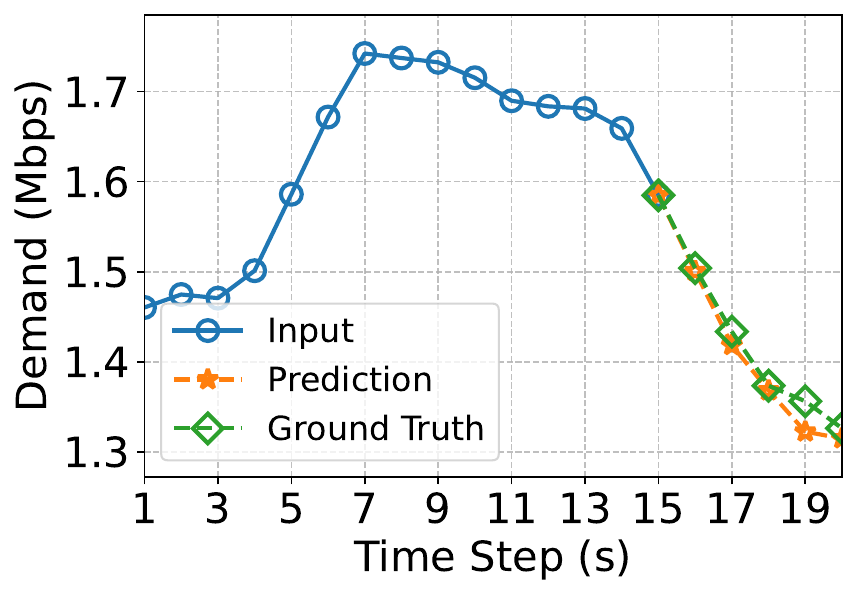}
    \captionsetup{labelformat=empty}
    \vspace{-0.7cm}
    \caption*{(b)}
    \end{minipage}
    \hfill
    \begin{minipage}[t]{0.16\linewidth}
    \includegraphics[width=1\textwidth]{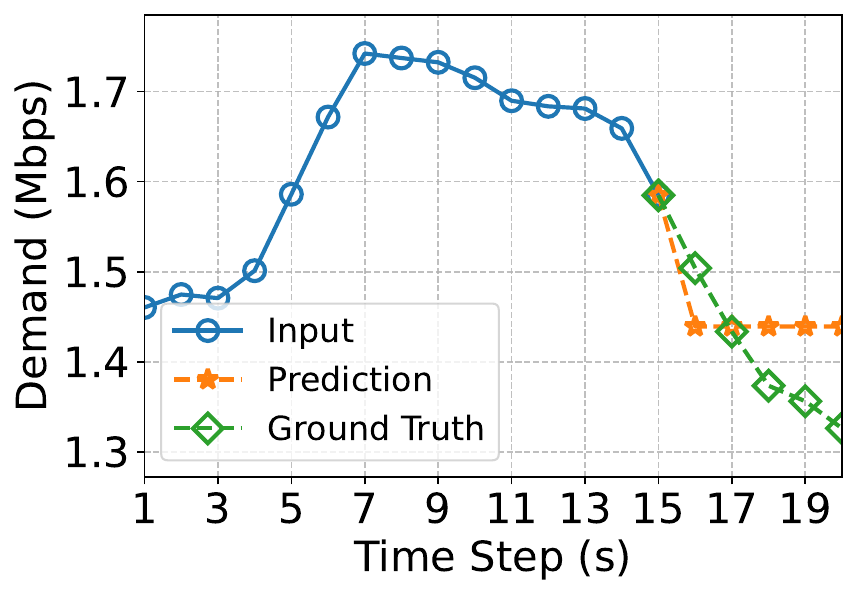}
    \captionsetup{labelformat=empty}
    \vspace{-0.7cm}
    \caption*{(c)}
    \end{minipage}
    \hfill
    \begin{minipage}[t]{0.16\linewidth}
    \includegraphics[width=1\textwidth]{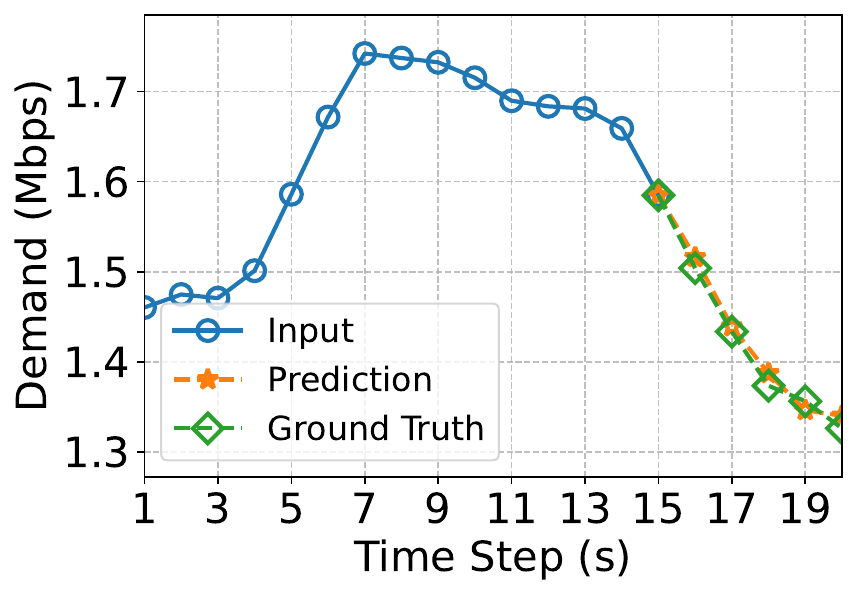}
    \captionsetup{labelformat=empty}
    \vspace{-0.7cm}
    \caption*{(d)}
    \end{minipage}
    \hfill
    \begin{minipage}[t]{0.16\linewidth}
    \includegraphics[width=1\textwidth]{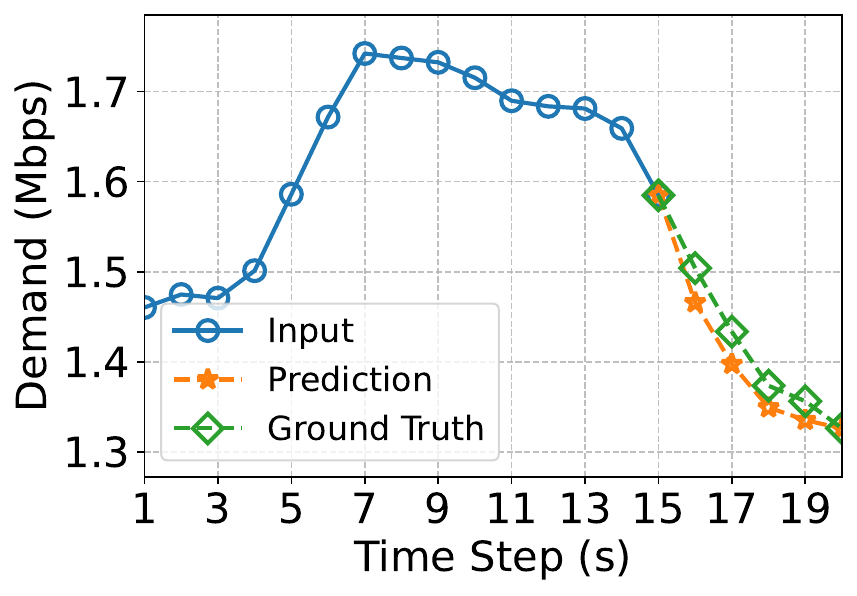}
    \captionsetup{labelformat=empty}
    \vspace{-0.7cm}
    \caption*{(e)}
    \end{minipage}
    \hfill
    \begin{minipage}[t]{0.16\linewidth}
    \includegraphics[width=1\textwidth]{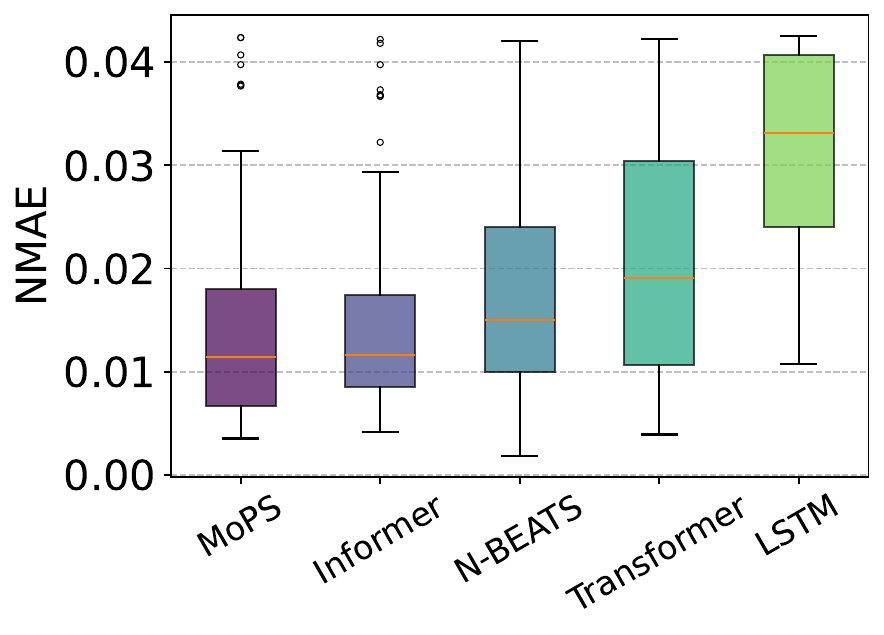}
    \captionsetup{labelformat=empty}
    \vspace{-0.7cm}
    \caption*{(f)}
    \end{minipage}
    \vspace{-0.3cm}
    \caption{\small Prediction of user demand in aAgent based on (a) LSTM, (b) N-BEATS, (c) Transformer, (d) Informer, (e) MoPS, and (f) comparison of different models' prediction accuracies.}
    \label{Figure_MethodComaprision_prediction_aAgent}
    \vspace{-0.4cm}
\end{figure*}

\begin{figure*}[t]
    \begin{minipage}[t]{0.16\linewidth}
    \includegraphics[width=1\textwidth]{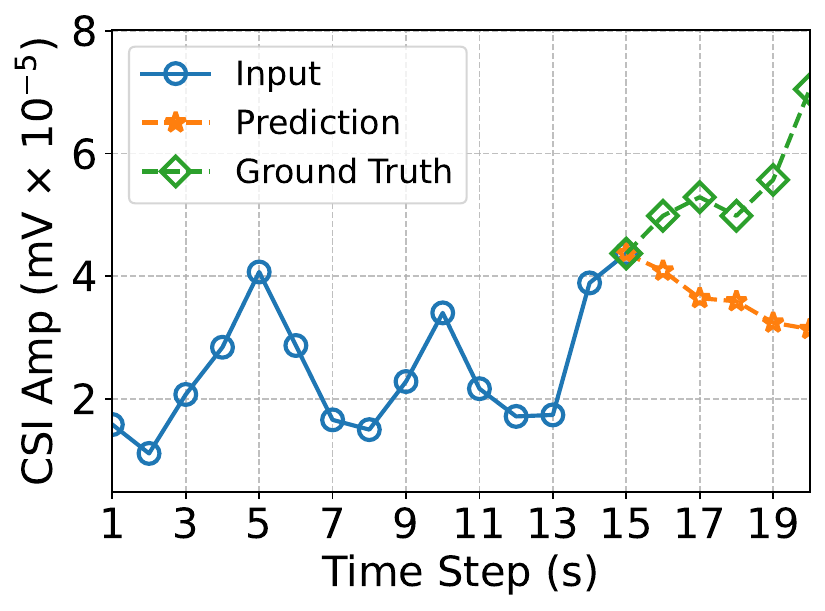}
    \captionsetup{labelformat=empty}
    \vspace{-0.7cm}
    \caption*{(a)}
    \end{minipage}
    \hfill
    \begin{minipage}[t]{0.16\linewidth}
    \includegraphics[width=1\textwidth]{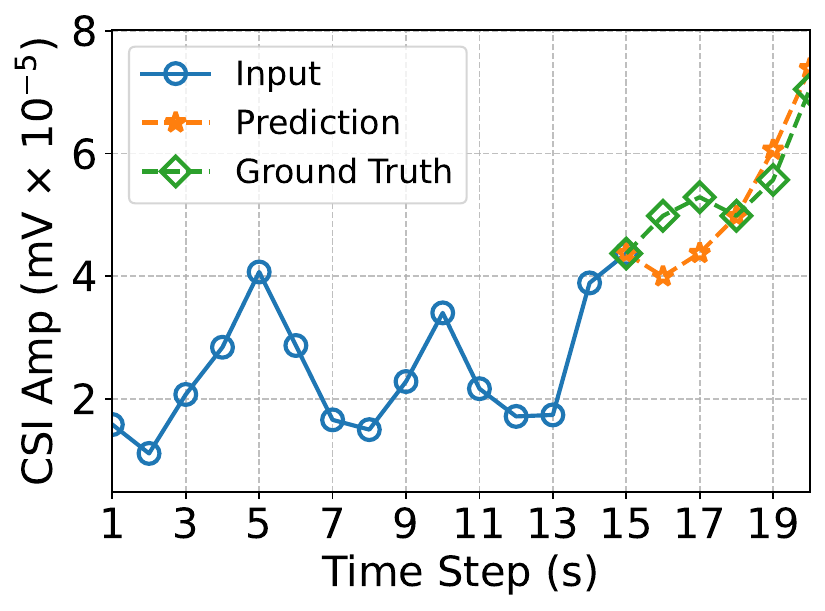}
    \captionsetup{labelformat=empty}
    \vspace{-0.7cm}
    \caption*{(b)}
    \end{minipage}
    \hfill
    \begin{minipage}[t]{0.16\linewidth}
    \includegraphics[width=1\textwidth]{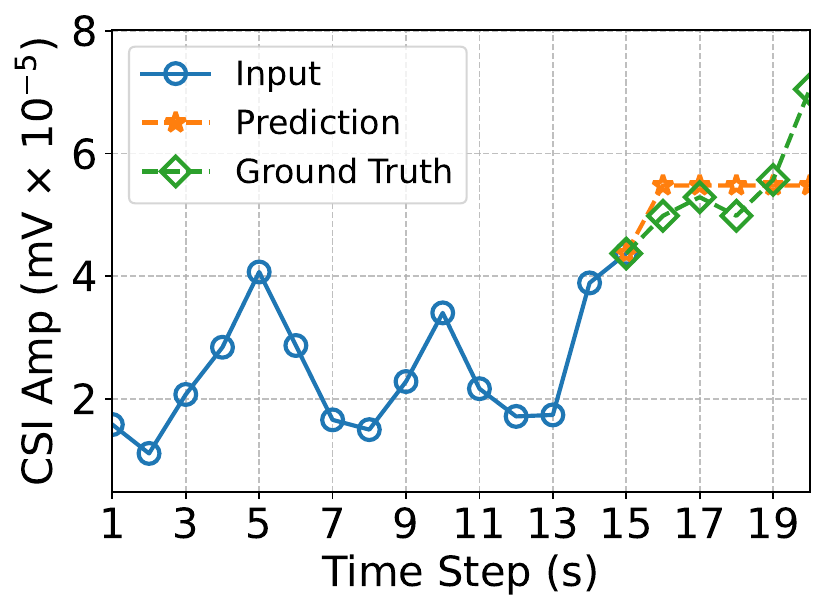}
    \captionsetup{labelformat=empty}
    \vspace{-0.7cm}
    \caption*{(c)}
    \end{minipage}
    \hfill
    \begin{minipage}[t]{0.16\linewidth}
    \includegraphics[width=1\textwidth]{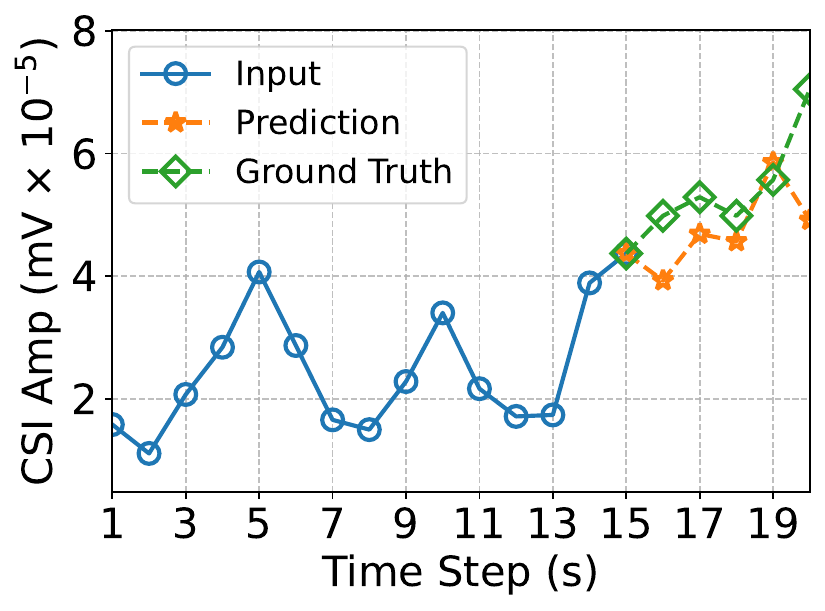}
    \captionsetup{labelformat=empty}
    \vspace{-0.7cm}
    \caption*{(d)}
    \end{minipage}
    \hfill
    \begin{minipage}[t]{0.16\linewidth}
    \includegraphics[width=1\textwidth]{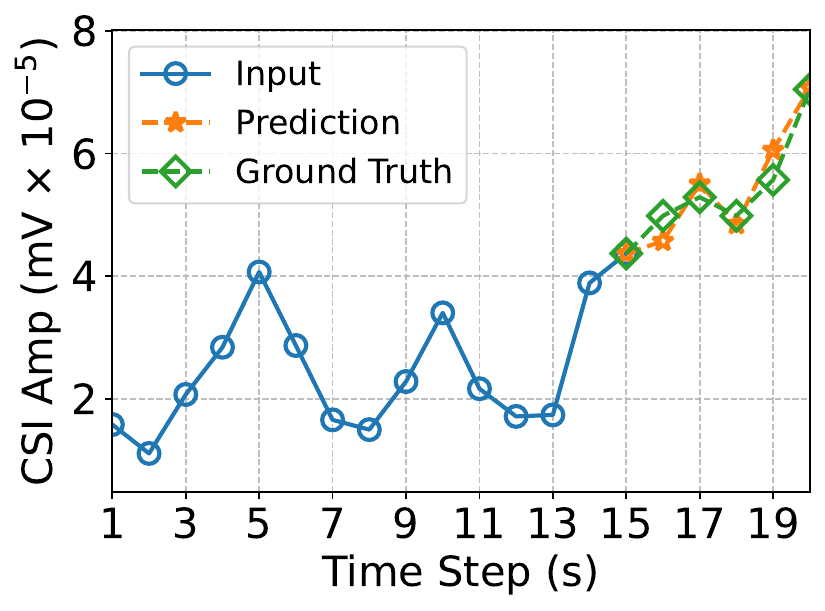}
    \captionsetup{labelformat=empty}
    \vspace{-0.7cm}
    \caption*{(e)}
    \end{minipage}
    \hfill
    \begin{minipage}[t]{0.16\linewidth}
    \includegraphics[width=1\textwidth]{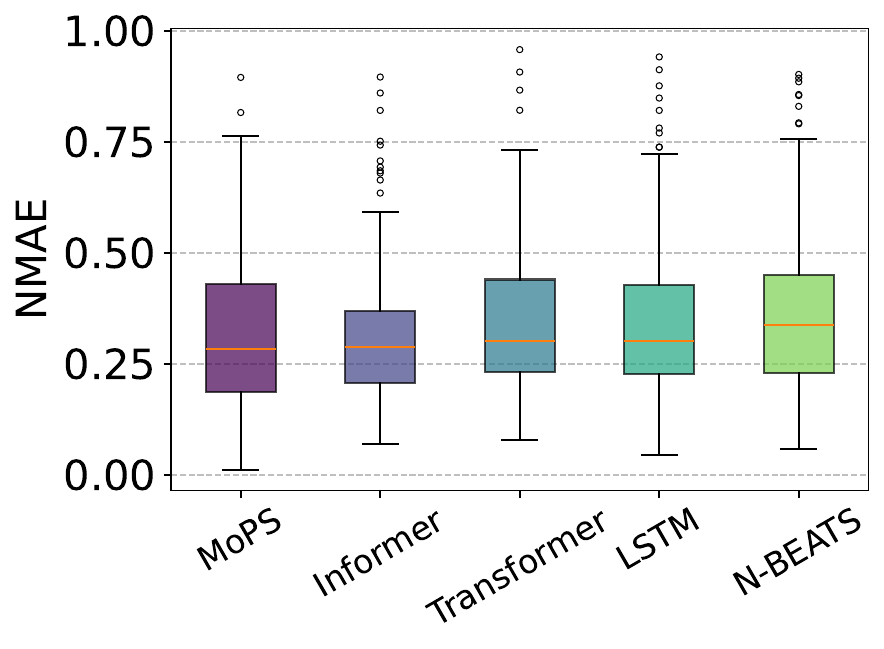}
    \captionsetup{labelformat=empty}
    \vspace{-0.7cm}
    \caption*{(f)}
    \end{minipage}
    \vspace{-0.3cm}
    \caption{\small Prediction of CSIs (n1 band, 2110-2170 MHz) in pAgent based on (a) LSTM, (b) N-BEATS, (c) Transformer, (d) Informer, (e) MoPS, and (f) comparison of different models' prediction accuracies.}
    \label{Figure_MethodComaprision_prediction_pAgent}
    \vspace{-0.4cm}
\end{figure*}

\begin{figure*}[t]
    \begin{minipage}[t]{0.16\linewidth}
    \includegraphics[width=1\textwidth]{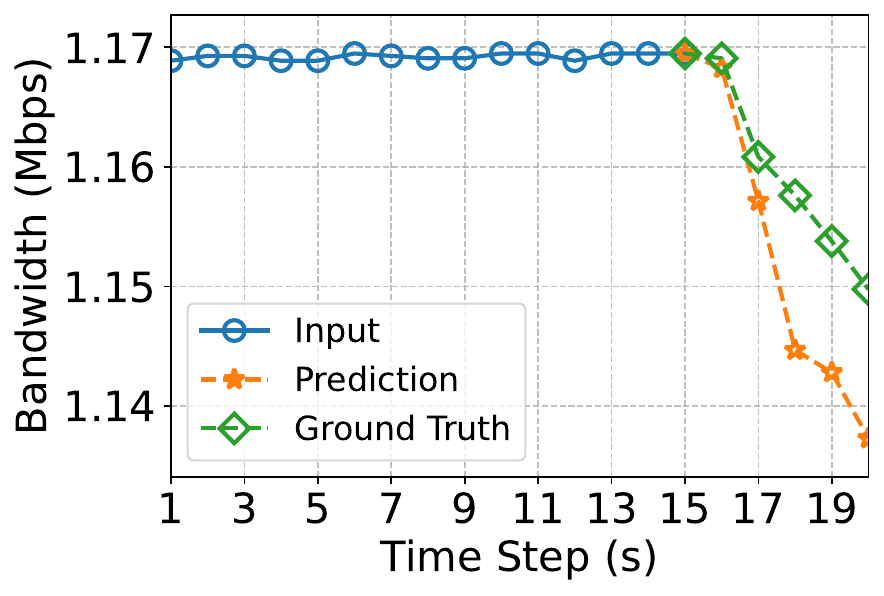}
    \captionsetup{labelformat=empty}
    \vspace{-0.7cm}
    \caption*{(a)}
    \end{minipage}
    \hfill
    \begin{minipage}[t]{0.16\linewidth}
    \includegraphics[width=1\textwidth]{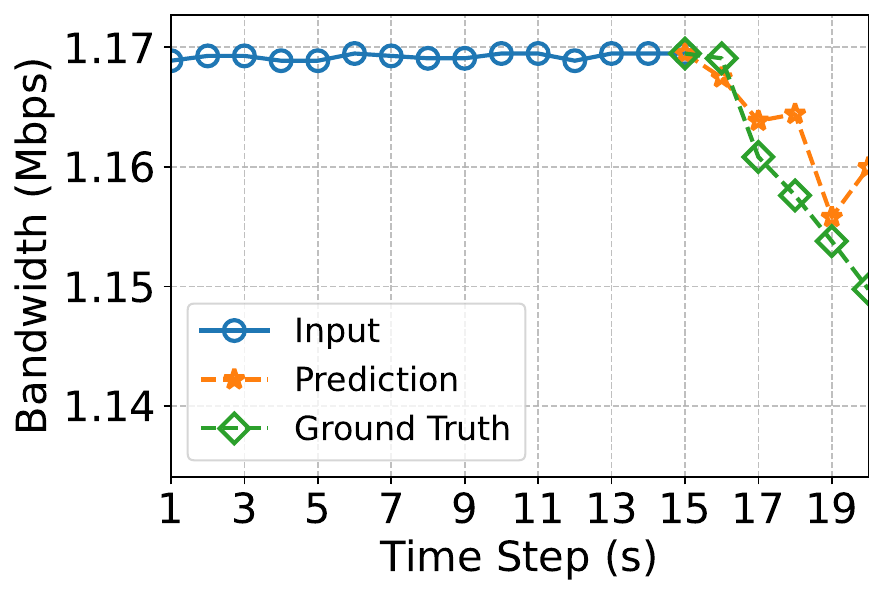}
    \captionsetup{labelformat=empty}
    \vspace{-0.7cm}
    \caption*{(b)}
    \end{minipage}
    \hfill
    \begin{minipage}[t]{0.16\linewidth}
    \includegraphics[width=1\textwidth]{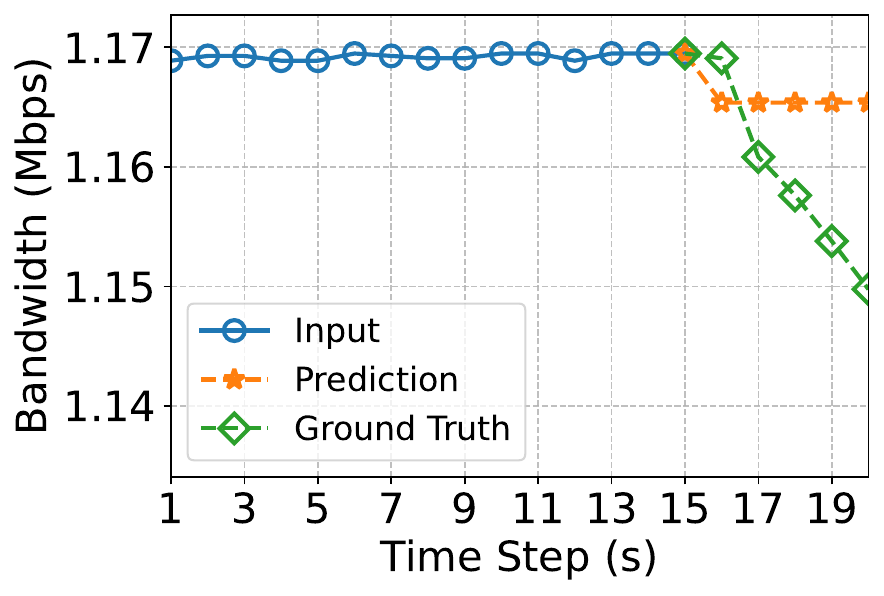}
    \captionsetup{labelformat=empty}
    \vspace{-0.7cm}
    \caption*{(c)}
    \end{minipage}
    \hfill
    \begin{minipage}[t]{0.16\linewidth}
    \includegraphics[width=1\textwidth]{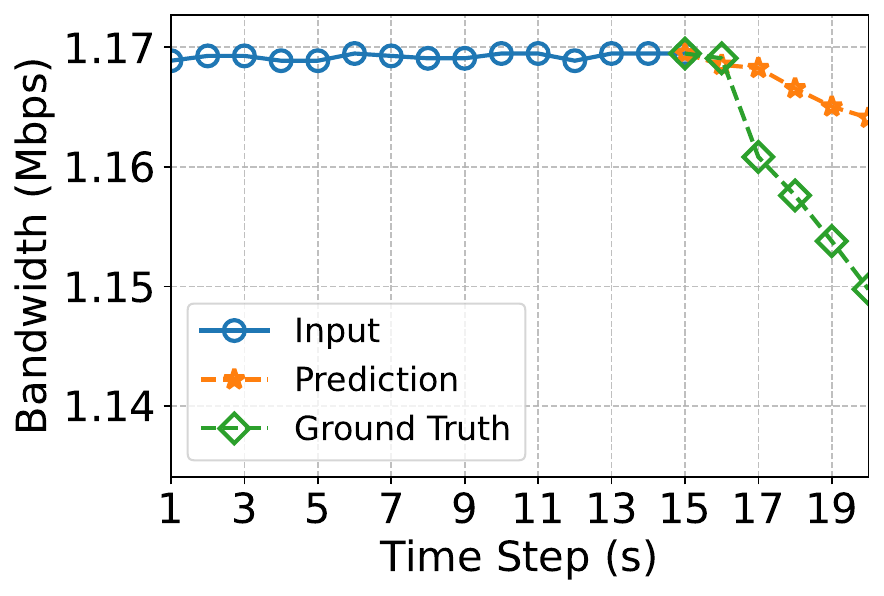}
    \captionsetup{labelformat=empty}
    \vspace{-0.7cm}
    \caption*{(d)}
    \end{minipage}
    \hfill
    \begin{minipage}[t]{0.16\linewidth}
    \includegraphics[width=1\textwidth]{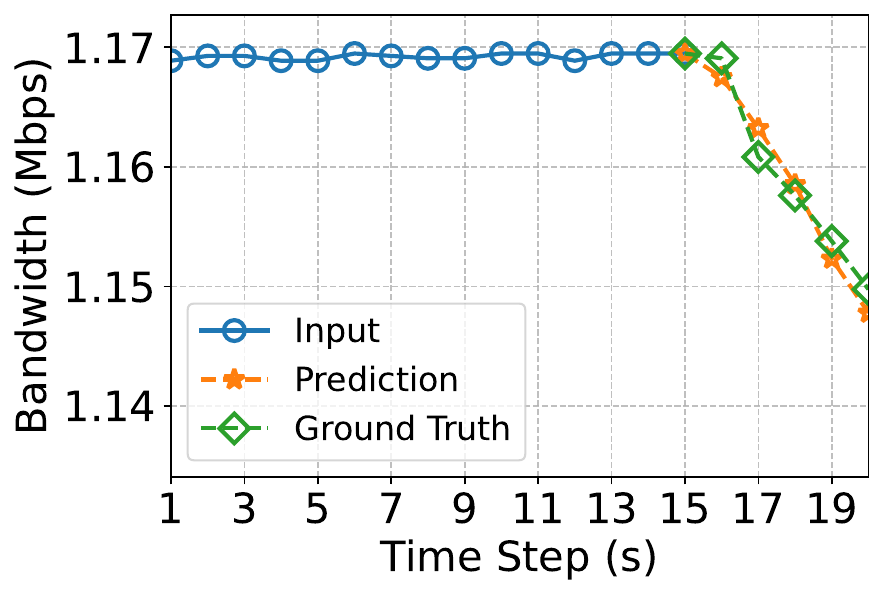}
    \captionsetup{labelformat=empty}
    \vspace{-0.7cm}
    \caption*{(e)}
    \end{minipage}
    \hfill
    \begin{minipage}[t]{0.16\linewidth}
    \includegraphics[width=1\textwidth]{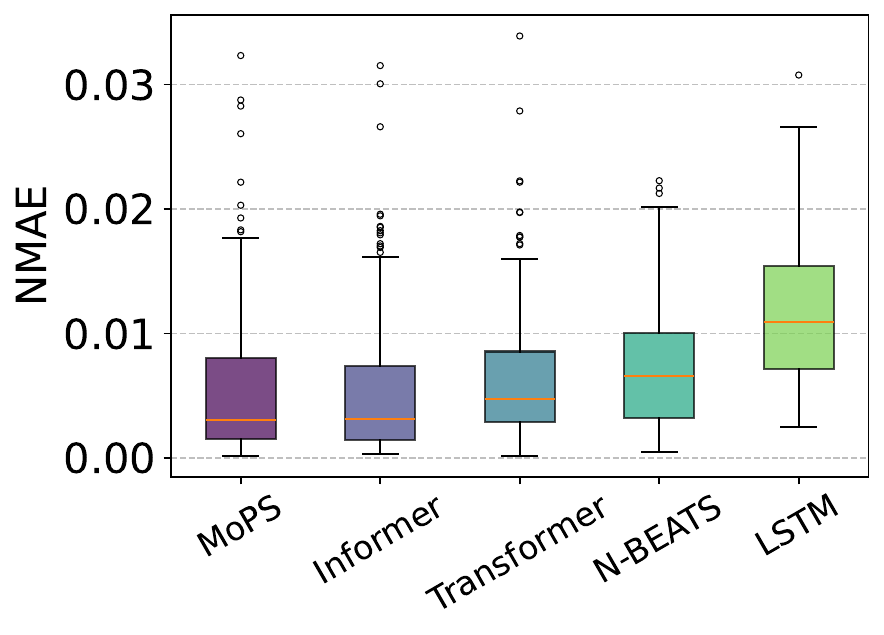}
    \captionsetup{labelformat=empty}
    \vspace{-0.7cm}
    \caption*{(f)}
    \end{minipage}
    \vspace{-0.3cm}
    \caption{\small Predicted network traffics of nAgent based on (a) LSTM, (b) N-BEATS, (c) Transformer, (d) Informer, (e) MoPS, and (f) comparisons of accuracies.}
    \label{Figure_MethodComaprision_prediction_nAgent}
    \vspace{-0.6cm}
\end{figure*}

To further verify the improvements on each agent's prediction results by allowing partial model sharing among multiple agents that interact with independent environments, in Fig. \ref{Figure_MethodComaprision_prediction_aAgent}--\ref{Figure_MethodComaprision_prediction_nAgent}, we compare the prediction accuracy of each agent using MoPS to that of no model sharing, in which each agent trains an independent model using the state-of-the-art time-series prediction algorithms, including LSTM \cite{albeladi2023time}, N-BEATS \cite{oreshkin2019n}, Transformer \cite{zerveas2021transformer}, and Informer \cite{zhou2021informer}. We also present the normalized mean absolute error (NMAE) of the prediction results from different models using the box-and-whisker plot to evaluate their performance. We can observe that MoPS achieves the highest prediction accuracy, i.e., with the lowest median NMAE (the orange line), for all three agents, e.g., the median NMAE values of MoPS are approximately 0.0109, 0.284, and 0.00267 at the physical, network, and application layers, respectively, corresponding to performance gains of 4.59\%, 1.41\%, and 14.61\%, respectively, compared to the best-performing baseline methods. Also, the MoPS and Informer have relatively smaller variability compared to other models. This means that these two models offer higher consistencies in their prediction results than others.

\begin{figure}
    \begin{minipage}[t]{0.45\linewidth}
    \includegraphics[width=1\textwidth]{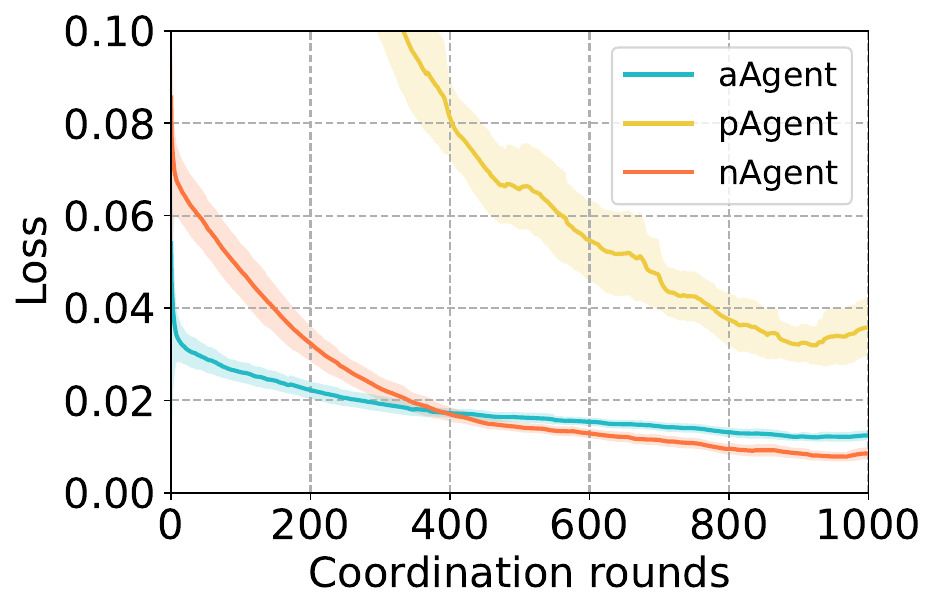}
    \captionsetup{labelformat=empty}
    \vspace{-0.7cm}
    \caption*{(a)}
    \end{minipage}
    \hfill
    \begin{minipage}[t]{0.45\linewidth}
    \includegraphics[width=1\textwidth]{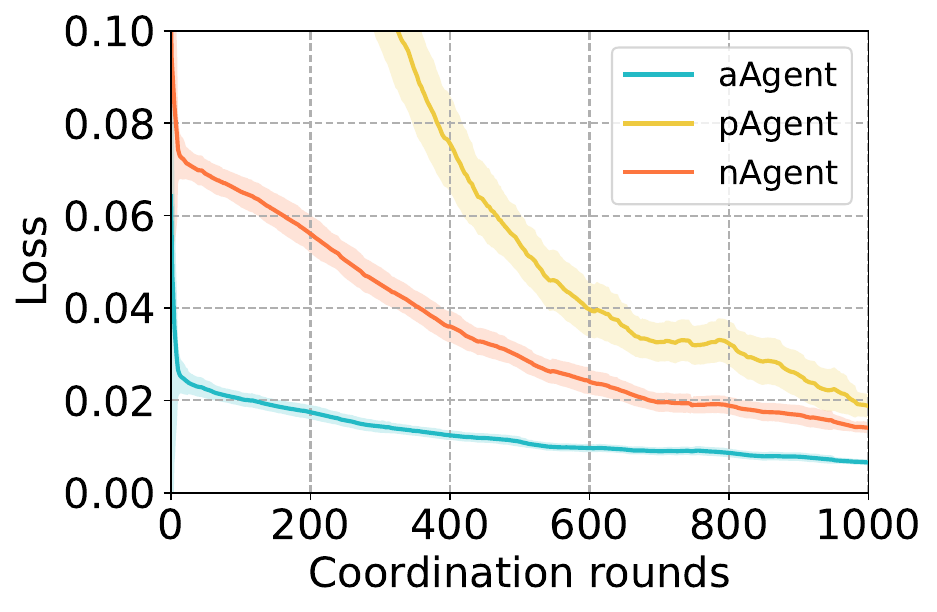}
    \captionsetup{labelformat=empty}
    \vspace{-0.7cm}
    \caption*{(b)}
    \end{minipage}
    \vspace{-0.3cm}
    \caption{\small Convergence of different agents based on (a) static- and (b) dynamic-weighting algorithms using Scheme 2.}
    \label{Figure_training_dynamics}
    \vspace{-0.4cm}
\end{figure}

\begin{figure}[t]
    \begin{minipage}[t]{0.45\linewidth}
    \includegraphics[width=1\textwidth]{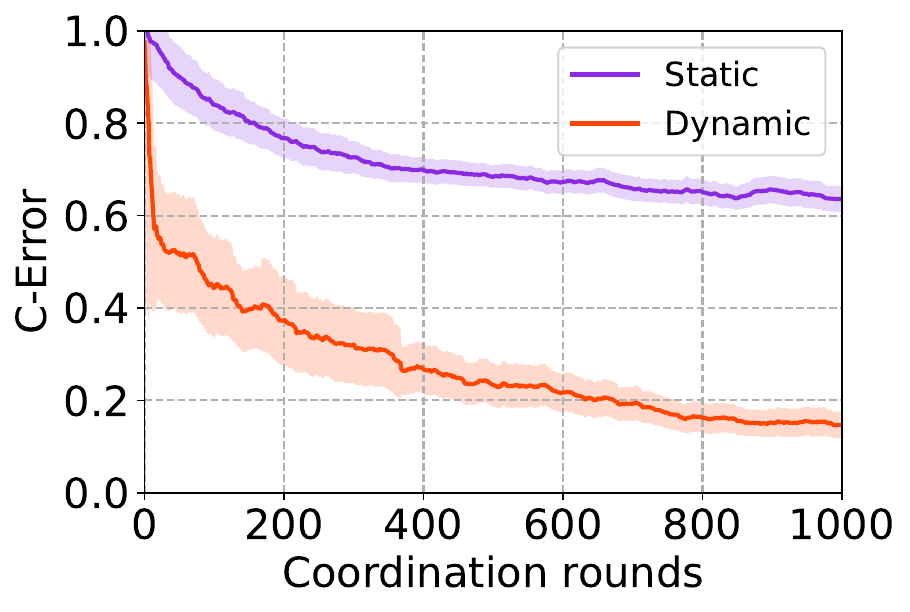}
    \captionsetup{labelformat=empty}
    \vspace{-0.7cm}
    \caption*{(a)}
    \end{minipage}
    \hfill
    \begin{minipage}[t]{0.45\linewidth}
    \includegraphics[width=1\textwidth]{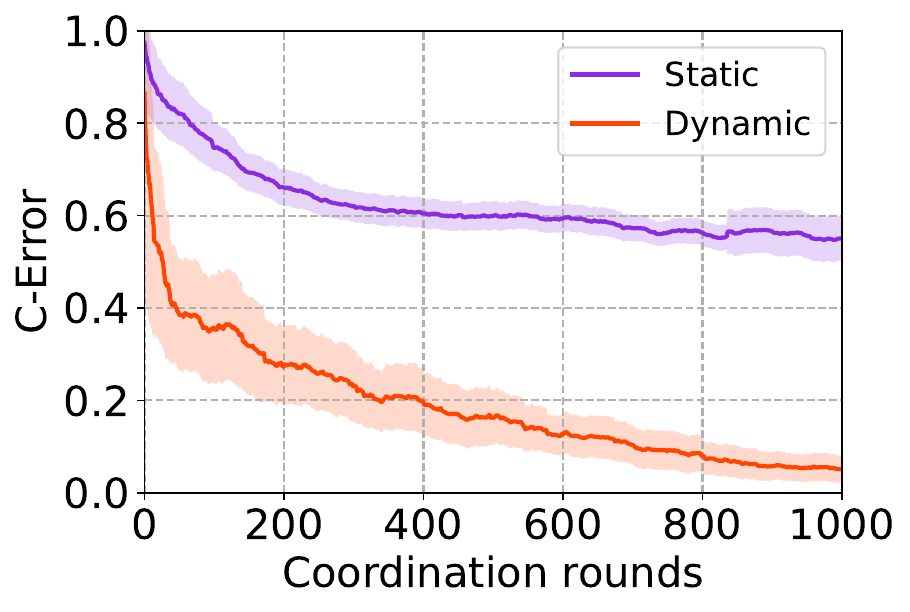}
    \captionsetup{labelformat=empty}
    \vspace{-0.7cm}
    \caption*{(b)}
    \end{minipage}
    \vspace{-0.3cm}
    \caption{\small \blu{C-error of MoPS under: (a) Scheme 2, and (b) Scheme 3 based on the static and dynamic-weighting algorithms.}} 
    \label{Figure_CerrorMoPSSchemes}
    \vspace{-0.6cm}
\end{figure}


    
\subsection{Dynamic- vs. Static-weighting Algorithms}

To evaluate the performance of the proposed dynamic-weighting algorithm in reducing the conflicts among different agents with different objective functions, we compare the convergences of loss functions of different agents when being collaboratively trained in Fig. \ref{Figure_training_dynamics} and also present the C-errors of the two partial sharing schemes (Schemes 2 and 3) with static-weighting and dynamic-weighting algorithms in Fig. \ref{Figure_CerrorMoPSSchemes}. We can observe that the loss function values of different agents exhibit different dynamics when being trained by the static-weighting algorithm, e.g., the loss values of the pAgent stop decreasing after around 860 rounds of coordination. In the dynamic-weighting algorithm, the loss functions of all the agents converge to a smaller value. The effectiveness of the dynamic-weight algorithm in resolving conflicts among agents is further verified in Fig. \ref{Figure_CerrorMoPSSchemes}, where we can directly observe that the C-errors of both partial sharing schemes (Schemes 2 and 3) trained by the dynamic-weighting algorithm are much lower, i.e., around $76.60\%$ and $83.81\%$ reductions in Schemes 2 and 3, than those trained by the static-weighting algorithm. We can also observe that the C-error of Scheme 3 is always lower than that of Scheme 2. This is because the shared-part of model parameters hosted by the agent controller mainly captures the common features of different agents' objectives, and therefore, the higher the portion of the shared parameters, the more C-error for the AgentNet. 



\footnotesize
\begin{table}
    \centering
    \belowrulesep=0pt
    \aboverulesep=0pt
    \caption{Comput. and commun. resource cost}    \label{resourse_consumption}
    \begin{tabular}{c|c|cccccc}
        \Xhline{1pt}
        \multirow{2}{*}{{\bf Algorithm}} & \multirow{2}{*}{{\bf Scheme}} & \multicolumn{3}{c}{{\bf Comput. (MFLOPS)}} & \multirow{2}{*}{{\makecell[c]{\bf Commun. \\ \bf Rounds}}} \\
        \cmidrule(lr){3-5}
        & & {\bf Agent} & {\bf Agent Cont.} & \blu{\bf Total} &\\
        \Xcline{1-2}{0.4pt}
        \Xhline{0.5pt}
        \multirow{3}{*}{\makecell[c]{Static \\ (Training)}} 
        & Scheme1 & 116.5 & / & \blu{349.5} & / \\
        & Scheme2 & 47.9 & 277.0 & \blu{420.7} & 690 \\
        & Scheme3 & 87.3 & 121.4 & \blu{383.3} & 651 \\
        \Xhline{0.5pt}
        \multirow{2}{*}{\makecell[c]{Dynamic \\(Training)}} 
        & Scheme2 & 158.0 & 589.5 & \blu{1063.5} & 514 \\
        & Scheme3 & 276.8 & 255.1 & \blu{1085.5} & 486 \\
        \Xhline{0.5pt}
        \multirow{3}{*}{Inference} 
        & Scheme1 & 31.1 & / & \blu{93.3} & / \\
        & Scheme2 & 13.8 & 52.1 & \blu{93.5} & / \\
        & Scheme3 & 26.7 & 22.3 & \blu{102.4} & / \\
        \Xhline{1pt}
    \end{tabular} 
    \vspace{-0.7cm}
\end{table}
\normalsize

To evaluate the communication and computational resource consumption during training and inference with dynamic-weighting and static-weighting algorithms, we record the number of coordination rounds and the million FLOPs (MFLOPs) of the agent controller and agents required to achieve a satisfactory O-error, e.g., less than or equal to a fixed threshold 0.4, with different MoPS schemes trained by the static-weighting and dynamic-weighting algorithms in Table \ref{resourse_consumption}. Note that in our experiments, we adopt the same-sized Transformer model for different agents and, therefore, their required MFLOPs are equal. We can observe that increasing the number of shared parameters at the agent controller can significantly reduce the computational load required by each agent. This, however, results in higher communication cost for both static-weighting and dynamic-weighting training algorithms to reach a certain inference accuracy. This is because, as mentioned earlier, allowing more shared-part model parameters generally results in higher conflicts among agents, leading to slower convergence in collaborative learning. More importantly, this observation reveals a fundamental tradeoff between the computational load distributed to each agent and the communication costs required to achieve a specific optimization objective.
\blu{We can also observe that, while the dynamic-weighting algorithm effectively mitigates conflicts among multi-objective agents, it entails a significantly higher computational overhead during the training phase. Specifically, the computational complexity, quantified in MFLOPs, of the dynamic-weighting approach is approximately three times that of the static-weighting baseline. This increase is primarily due to the additional weight-update procedures required during each coordination cycle to achieve conflict resolution.}
We also present the resource consumption of the inference phase in Table \ref{resourse_consumption}. We can observe that during the inference phase, computational costs are significantly reduced compared to the training phase, e.g., in Scheme 2 (or Scheme 3), the FLOPs of each agent and the agent controller reduce to only 8.73\% and 8.84\% (or 9.65\% and 8.74\%) of those, required for the training phase, respectively. 

\begin{figure}[t]
    \centering
    \begin{minipage}[t]{0.32\linewidth}
    \includegraphics[width=1\textwidth]{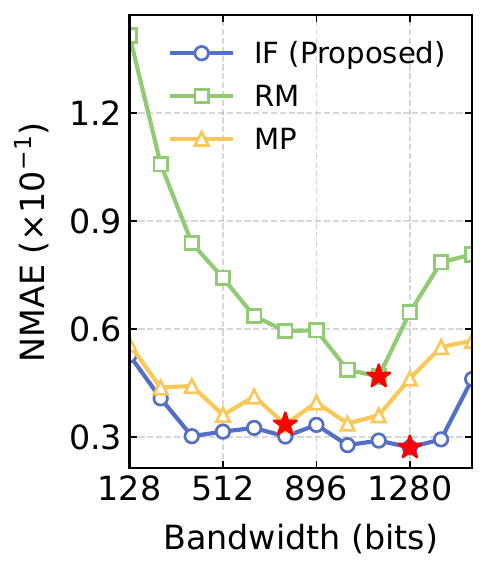}
    \captionsetup{labelformat=empty}
    \vspace{-0.7cm}
    \caption*{(a)}
    \label{}
    \end{minipage}
    \hfill
    \begin{minipage}[t]{0.32\linewidth}
    \includegraphics[width=1\textwidth]{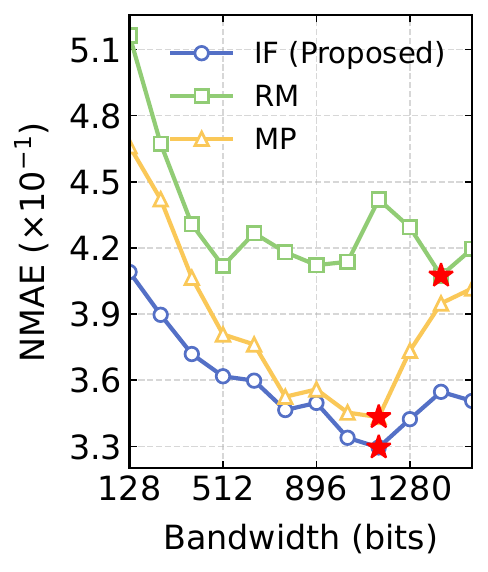}
    \captionsetup{labelformat=empty}
    \vspace{-0.7cm}
    \caption*{(b)}
    \label{}
    \end{minipage}
    \hfill
    \begin{minipage}[t]{0.32\linewidth}
    \includegraphics[width=1\textwidth]{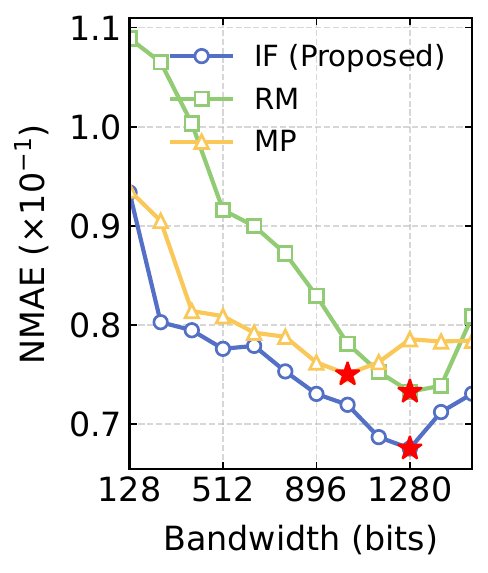}
    \captionsetup{labelformat=empty}
    \vspace{-0.7cm}
    \caption*{(c)}
    \label{}
    \end{minipage}
    \vspace{-0.3cm}
    \caption{\small \blu{Prediction accuracy of different compression solutions under different bandwidth constraints: (a) user demand in aAgent, (b) CSI in pAgent, and (c) network traffic in nAgent.}}
    \label{Figure_compression}
    \vspace{-0.6cm}
\end{figure}

\subsection{\blu{Compression and Decompression Modules}}

\blu{We evaluate the performance of different agents when compression and decompression modules are deployed. In Fig. \ref{Figure_compression}, we present the inference performance of aAgent, nAgent, and pAgent, quantified by the Normalized Mean Absolute Error (NMAE), under various bandwidth constraints. Lower NMAE values indicate higher model prediction performance. We compare our proposed importance filter (IF)-based compression module with two benchmark solutions: (1) random masking (RM), also called random dropout or stochastic feature selection in the literature, in which, due to the bandwidth constraint, only a limited number of output features will be randomly selected for each agent to be sent to the agent controller, and (2) max-magnitude pruning
(MP), also called top-$k$ selection scheme, in which only the output features with the highest magnitudes (largest weight values) will be selected and sent to the agent controller. Experimental results in Fig. \ref{Figure_compression} indicate an interesting observation:  there exists a non-monotonic relationship between compression and accuracy: all three agents' performance initially improves as the compression ratio rises, yet begins to degrade once specific thresholds are exceeded. This trend aligns with our characterization of the compression module as an importance filter. By selectively removing redundant information from the observed data, the compression module effectively eliminates irrelevant or noisy information that might otherwise impede the agents' learning processes. Consequently, the initial stages of compression facilitate a "denoising" effect that enhances the agents' predictive accuracy. However, as the compression ratio continues to increase, the loss of task-relevant information eventually outweighs the benefits of noise reduction, leading to a decline in performance. Notably, our proposed compression module exhibits much higher inference accuracy even when the bandwidth constraints become stringent, i.e., our proposed IF-based compression module reduces the NMAE values by up to $62.70\%$, $20.76\%$, and $24.63\%$, compared to the RM scheme, and $42.70\%$, $12.24\%$, and $11.30\%$, compared to the MP scheme for the aAgent, nAgent, and pAgent, respectively. These findings demonstrate that the proposed module maintains a balance between communication efficiency and prediction accuracy, even under significant bandwidth constraints.}

\subsection{\blu{Different Network Scales}}

\begin{figure}[t]
    \begin{minipage}[t]{0.32\linewidth}
    \includegraphics[width=1\textwidth]{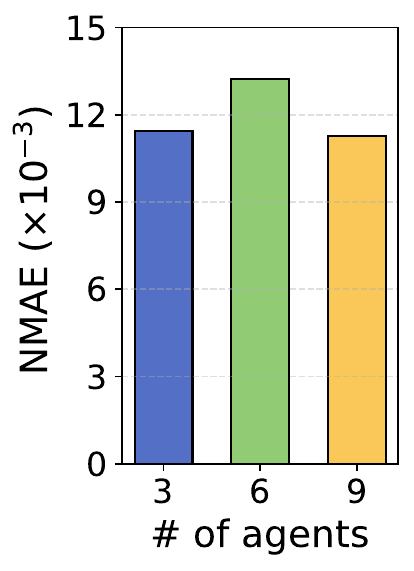}
    \captionsetup{labelformat=empty}
    \vspace{-0.7cm}
    \caption*{(a)}
    \end{minipage}
    \hfill
    \begin{minipage}[t]{0.32\linewidth}
    \includegraphics[width=1\textwidth]{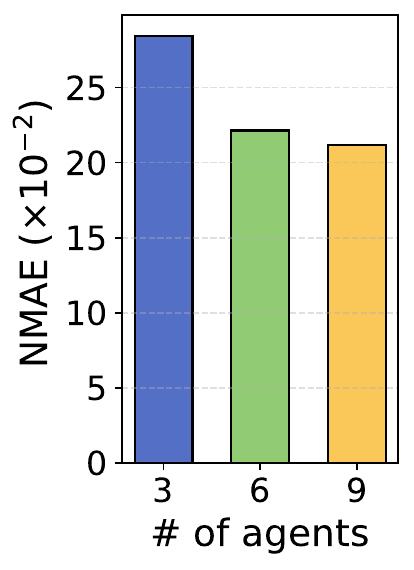}
    \captionsetup{labelformat=empty}
    \vspace{-0.7cm}
    \caption*{(b)}
    \end{minipage}
    \begin{minipage}[t]{0.32\linewidth}
    \includegraphics[width=1\textwidth]{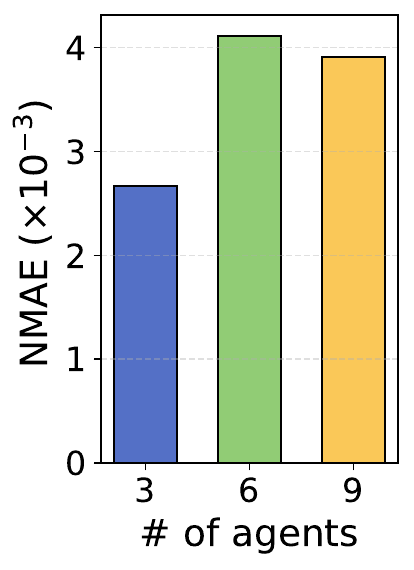}
    \captionsetup{labelformat=empty}
    \vspace{-0.7cm}
    \caption*{(c)}
    \end{minipage}
    \vspace{-0.3cm}
    \caption{\small \blu{Prediction accuracy of (a) user demand in aAgent, (b) CSI in pAgent, and (c) network traffic in nAgent with different numbers of agents.} }
    \label{Figure_number_of_agents}
    \vspace{-0.6cm}
\end{figure}

\blu{Next, we evaluate the performance of SANet under various network scales, i.e., different numbers of agents. In Fig. \ref{Figure_number_of_agents}, we present the average performance of agents deployed in three different layers of the network system when each layer has different numbers of agents. 
We can observe that the model's predictive performance, quantified by the NMAE, remains remarkably stable as the agent count increases. This consistency is attributed to the decentralized architecture of SANet, which facilitates parallel task execution among all active agents. Furthermore, the integrated dynamic-weighting algorithm effectively mitigates potential multi-agent conflicts, a critical requirement for maintaining stability in a large-scale AgentNet where a large number of agents with divergent objectives co-exist. These empirical results validate that the proposed framework is both highly scalable and robust. } 

\subsection{Three-way Tradeoff}

\begin{figure}
    \begin{minipage}[t]{0.49\linewidth}
    \includegraphics[width=1\textwidth]{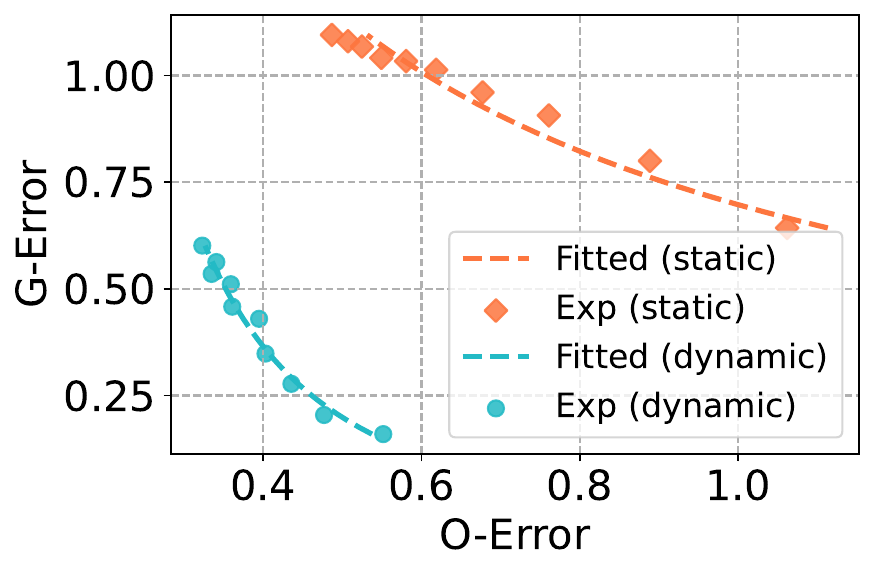}
    \captionsetup{labelformat=empty}
    \vspace{-0.7cm}
    \caption*{(a)}
    \end{minipage}
    \hfill
    \begin{minipage}[t]{0.49\linewidth}
    \includegraphics[width=1\textwidth]{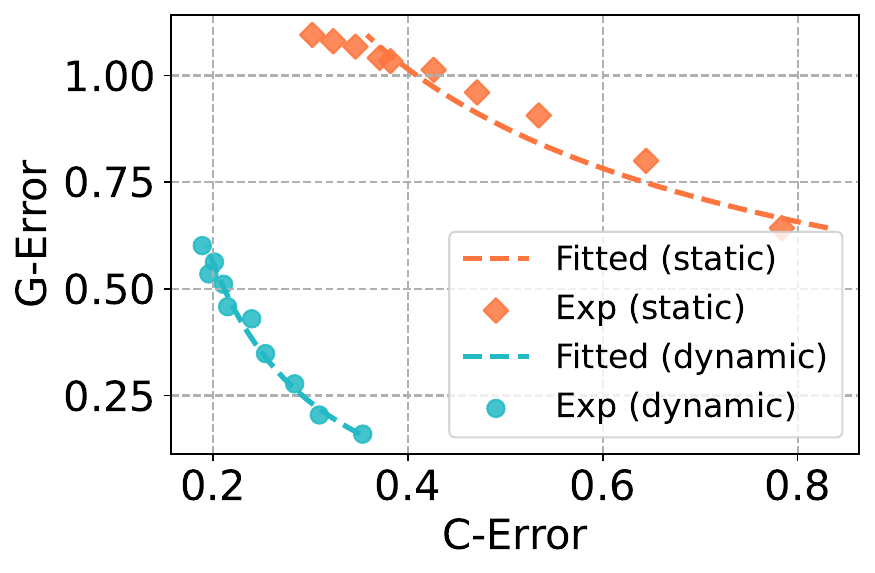}
    \captionsetup{labelformat=empty}
    \vspace{-0.7cm}
    \caption*{(b)}
    \end{minipage}
    \vspace{-0.3cm}
    \caption{\small (a) Tradeoff between G-error and O-error and (b) that between G-error and C-error verified by both theoretical results and experimental results. } 
    \label{Figure_Three-wayTradeoff}
    \vspace{-0.6cm}
\end{figure}

To verify the three-way tradeoff observed in our theoretical results, we investigate the correlation between G-error, O-error, and C-error for the static-weighting and dynamic-weighting algorithms based on a series of experiments conducted based on our SANet prototype in Fig. \ref{Figure_Three-wayTradeoff}, where we plot the data points recorded in these experiments into the two-dimensional plane of G-error and O-error as well as that of G-error and C-error. To verify the trends discovered in our theoretical bounds, we first estimate the empirical values of hyperparameters $\mu_g$, $\mu_l$, $c_I$, and $G$ in Theorems \ref{Theorem_UpperBoundOError_without CA}-\ref{Theorem_UpperBoundCError} by fitting the derived mathematical expressions to the experimental data points. We then substitute these estimated empirical values back into the derived mathematical expressions to obtain the theoretical curves, which are plotted in the same two-dimensional plane as the experimental data points, so we can have a direct, visual comparison between the experimental results and the derived theoretical results. 
We can observe that the experimental results match the trend of the theoretical curves, that is, increasing G-error will cause decreases in both O-error and C-error for both the static-weighting and dynamic-weighting algorithms. We can also observe that the dynamic-weighting algorithm can significantly reduce both O-error and C-error for any given G-error value. As mentioned earlier, the dynamic-weighting algorithm requires slightly more computational resources at each agent than the static-weighting algorithm, and is therefore more suitable for the agents with relatively higher computational capacities and more stringent requirements on the generalization and optimization performance.   


\section{Conclusion}
\label{Section_Conclusion}

This paper has proposed SANet, a novel semantic-aware AgentNet architecture that autonomously identifies the user's semantic goal and autonomously divides the identified goal into different subtasks for different agents. We have formulated the decentralized optimization of SANet as a multi-agent multi-objective problem and focus on finding the Pareto optimal solution among agents with distinct and potentially conflicting objectives. Three novel metrics for evaluating the performance of SANet have been proposed. 
A model partition and sharing framework, MoPS, has been proposed in which deep learning models of different agents can be partitioned into shared and agent-specific parts that are jointly constructed and deployed according to agents' local computational resources. Two decentralized optimization algorithms, static-weighting and dynamic-weighting algorithms, have been introduced to optimize the proposed metrics. \blu{We have developed a bandwidth-adaptive compression framework for different agents to perform in situ compression of their intermediate embeddings, according to their localized resource constraints and task requirements.} Theoretical 
results have proved that there exists a three-way tradeoff among optimization, generalization, and conflicting errors. 
Experiment results based on a SANet prototype have shown that the proposed framework achieves up to $14.61\%$ performance gains with only $44.37\%$ of the MFLOPs required for inference at each agent, compared to existing benchmarks. 

\bibliographystyle{IEEEtran}
\bibliography{DeepLearningRef}

\begin{thebibliography}{10}
\providecommand{\url}[1]{#1}
\csname url@samestyle\endcsname
\providecommand{\newblock}{\relax}
\providecommand{\bibinfo}[2]{#2}
\providecommand{\BIBentrySTDinterwordspacing}{\spaceskip=0pt\relax}
\providecommand{\BIBentryALTinterwordstretchfactor}{4}
\providecommand{\BIBentryALTinterwordspacing}{\spaceskip=\fontdimen2\font plus
\BIBentryALTinterwordstretchfactor\fontdimen3\font minus
  \fontdimen4\font\relax}
\providecommand{\BIBforeignlanguage}[2]{{%
\expandafter\ifx\csname l@#1\endcsname\relax
\typeout{** WARNING: IEEEtran.bst: No hyphenation pattern has been}%
\typeout{** loaded for the language `#1'. Using the pattern for}%
\typeout{** the default language instead.}%
\else
\language=\csname l@#1\endcsname
\fi
#2}}
\providecommand{\BIBdecl}{\relax}
\BIBdecl

\bibitem{XY2025SANNet}
Y.~Xiao, H.~Zhou, X.~Li, Y.~Gao, G.~Shi, and P.~Zhang, ``{SANNet}: A
  semantic-aware agentic {AI} networking framework for multi-agent cross-layer
  coordination,'' in \emph{IEEE GLOBECOM}, Taipei, Taiwan, Dec. 2025.

\bibitem{Nguyen2024MetaverseBrainComputer}
N.~Q. Hieu \emph{et~al.}, ``Enhancing immersion and presence in the metaverse
  with over-the-air brain-computer interface,'' \emph{IEEE Trans. Wireless
  Commun.}, vol.~23, no.~12, pp. 18\,532--18\,548, Dec. 2024.

\bibitem{XY2018TactileInternet}
Y.~Xiao and M.~Krunz, ``Distributed optimization for energy-efficient fog
  computing in the {T}actile {I}nternet,'' \emph{IEEE J. Sel. Area Commun.},
  vol.~36, no.~11, pp. 2390--2400, Nov. 2018.

\bibitem{shi2020semantic}
G.~Shi, Y.~Xiao, Y.~Li, and X.~Xie, ``From semantic communication to
  semantic-aware networking: Model, architecture, and open problems,''
  \emph{IEEE Communications Magazine}, vol.~59, no.~8, pp. 44--50, Aug. 2021.

\bibitem{Deniz2024JSCCProcIEEE}
D.~Gunduz, M.~A. Wigger, T.-Y. Tung, P.~Zhang, and Y.~Xiao, ``Joint
  source–channel coding: Fundamentals and recent progress in practical
  designs,'' \emph{Proceedings of the IEEE}, vol. 113, no.~9, pp. 888--919,
  Nov. 2024.

\bibitem{Albert2020AmbientIntelligence}
A.~Haque, A.~Milstein, and L.~Fei-Fei, ``Illuminating the dark spaces of
  healthcare with ambient intelligence,'' \emph{Nature}, vol. 585, no. 7824,
  pp. 193--202, Sept. 2020.

\bibitem{xiao2025AgentNet}
Y.~Xiao, G.~Shi, and P.~Zhang, ``Towards agentic {AI} networking in 6{G}: A
  generative foundation model-as-agent approach,'' \emph{IEEE Communications
  Magazine}, vol.~63, no.~9, pp. 68--74, Sept. 2025.

\bibitem{XY2020Selflearning}
Y.~Xiao, G.~Shi, Y.~Li, W.~Saad, and H.~V. Poor, ``{Toward self-learning edge
  intelligence in 6{G}},'' \emph{IEEE Communications Magazine}, vol.~58,
  no.~12, pp. 34--40, Dec. 2020.

\bibitem{Yang2022NetMagazine}
Y.~Yang \emph{et~al.}, ``6{G} network {AI} architecture for everyone-centric
  customized services,'' \emph{IEEE Network}, vol.~37, no.~5, pp. 71--80, 2023.

\bibitem{Morris2024PositionAGI}
M.~R. Morris \emph{et~al.}, ``Position: Levels of {AGI} for operationalizing
  progress on the path to {AGI},'' in \emph{Proceedings of ICML}, Vienna,
  Austria, Jul. 2024.

\bibitem{Shavit2023PracticesAgenticAI}
Y.~Shavit, S.~Agarwal, M.~Brundage, S.~Adler, C.~O’Keefe, R.~Campbell,
  T.~Lee, P.~Mishkin, T.~Eloundou, A.~Hickey \emph{et~al.}, ``Practices for
  governing agentic {AI} systems,'' \emph{Research Paper, OpenAI}, 2023.

\bibitem{Chang2024CrossLayerOpt}
C.~Wu \emph{et~al.}, ``Cross-layer optimization for statistical qos provision
  in {C-RAN} with finite-length coding,'' \emph{IEEE Trans. Commun.}, vol.~72,
  no.~6, pp. 3393--3407, Jun. 2024.

\bibitem{Shiwen2023CrossLayerOpt}
S.~He, Z.~An, J.~Zhu, M.~Zhang, Y.~Huang, and Y.~Zhang, ``Cross-layer
  optimization: Joint user scheduling and beamforming design with {QoS} support
  in joint transmission networks,'' \emph{IEEE Transactions on Communications},
  vol.~71, no.~2, pp. 792--807, Feb. 2023.

\bibitem{LiWei2025CrossLayerOpt}
L.-W. Chien and S.-M. Tseng, ``Proximal policy optimization for cross-layer
  joint design of {MISO} beamforming and {RIS} phase configuration,''
  \emph{IEEE Open Journal of the Communications Society}, vol.~6, pp. 8128 --
  8141, Jul. 2025.

\bibitem{alzailaa2025review}
A.~AlZailaa, J.~Corona, R.~Teixeira, H.~R. Chi, M.~Antunes, A.~Radwan, and
  R.~L. Aguiar, ``A review of the current usage of {AI/ML} for radio access
  network ({RAN}),'' \emph{IEEE Access}, vol.~13, pp. 119\,457 -- 119\,499,
  Jul. 2025.

\bibitem{taleb2023ai}
T.~Taleb, C.~Benzaid, R.~A. Addad, and K.~Samdanis, ``A{I}/{ML} for beyond 5{G}
  systems: Concepts, technology enablers \& solutions,'' \emph{Computer
  Networks}, vol. 237, p. 110044, Dec. 2023.

\bibitem{Zhu2024SANe}
H.~Zhu \emph{et~al.}, ``{SANS}ee: A physical-layer semantic-aware networking
  framework for distributed wireless sensing,'' \emph{IEEE Transactions on
  Mobile Computing}, vol.~24, no.~3, pp. 1636--1653, Mar. 2025.

\bibitem{lin2025bridge}
X.~Lin, ``The bridge toward 6{G}: 5{G}-advanced evolution in 3{GPP} release
  19,'' \emph{IEEE Communications Standards Magazine}, vol.~9, no.~1, pp.
  28--35, Mar. 2025.

\bibitem{ITU2023SAN}
G.~Shi and Y.~Xiao, ``Requirements of semantic-aware networking for future
  networks,'' ITU-T TR.Reqts-SAN, Nov. 2023.

\bibitem{XY2024ACMGetMobileSAN}
------, ``An introduction to semantic communication and semantic-aware
  networking standardization for 6{G},'' \emph{GetMobile: Mobile Comp. and
  Comm.}, vol.~28, no.~3, p. 14–19, Oct. 2024.

\bibitem{Sapkota2025AIagentsandAgenticAI}
R.~Sapkota, K.~I. Roumeliotis, and M.~Karkee, ``A{I} agents vs. agentic {AI}: A
  conceptual taxonomy, applications and challenges,'' \emph{Information
  Fusion}, vol. 126, p. 103599, Feb. 2026.

\bibitem{Durante2024AgentAISurvey}
Z.~Durante \emph{et~al.}, ``Agent {AI}: Surveying the horizons of multimodal
  interaction,'' \emph{arXiv preprint arXiv:2401.03568}, 2024.

\bibitem{Wang2024MobileAgentV2}
J.~Wang, H.~Xu, H.~Jia, X.~Zhang, M.~Yan, W.~Shen, J.~Zhang, F.~Huang, and
  J.~Sang, ``Mobile-{A}gent-v2: Mobile device operation assistant with
  effective navigation via multi-agent collaboration,'' in \emph{Proceedings of
  NeurIPS Conference}, vol.~37, Vancouver, Canada, Dec. 2024, pp. 2686--2710.

\bibitem{Wang2024RethinkBoundofLLRReasoning}
Q.~Wang, Z.~Wang, Y.~Su, H.~Tong, and Y.~Song, ``Rethinking the bounds of {LLM}
  reasoning: Are multi-agent discussions the key?'' in \emph{ACL}, Bangkok,
  Thailand, Aug. 2024, pp. 6106--6131.

\bibitem{tong2025wirelessagent}
J.~Tong, W.~Guo, J.~Shao, Q.~Wu, Z.~Li, Z.~Lin, and J.~Zhang,
  ``{WirelessAgent}: Large language model agents for intelligent wireless
  networks,'' \emph{arXiv preprint arXiv:2505.01074}, 2025.

\bibitem{li2026comagent}
H.~Li, M.~Xiao, K.~Wang, R.~Schober, D.~I. Kim, and Y.~L. Guan, ``{ComAgent}:
  Multi-{LLM} based agentic {AI} empowered intelligent wireless networks,''
  \emph{arXiv preprint arXiv:2601.19607}, 2026.

\bibitem{yuan2025automas}
D.~Yuan, J.~Peng, J.~Fan, B.~Ren, L.~Yang, and P.~Liu, ``{AutoMAS}: A generic
  multi-agent system for algorithm self-adaptation in wireless networks,''
  \emph{arXiv preprint arXiv:2511.18414}, 2025.

\bibitem{acharya2025agentic}
D.~B. Acharya, K.~Kuppan, and B.~Divya, ``Agentic {AI}: Autonomous intelligence
  for complex goals---a comprehensive survey,'' \emph{IEEE Access}, vol.~13,
  pp. 18\,912--18\,936, Jan. 2025.

\bibitem{FernandoSLCMC23}
H.~D. Fernando \emph{et~al.}, ``Mitigating gradient bias in multi-objective
  learning: {A} provably convergent approach,'' in \emph{ICLR}, Kigali, Rwanda,
  May 2023.

\bibitem{chen2023three}
L.~Chen \emph{et~al.}, ``Three-way trade-off in multi-objective learning:
  Optimization, generalization and conflict-avoidance,'' \emph{NeurIPS},
  vol.~36, pp. 70\,045--70\,093, New Orleans, LA, Dec. 2023.

\bibitem{yang2020rethinking}
Z.~Yang, Y.~Yu, C.~You, J.~Steinhardt, and Y.~Ma, ``Rethinking bias-variance
  trade-off for generalization of neural networks,'' in \emph{Proceedings of
  ICML Conference}, Virtual, Jul. 2020, pp. 10\,767--10\,777.

\bibitem{stein1956inadmissibility}
C.~Stein, \emph{Inadmissibility of the usual estimator for the mean of a
  multivariate normal distribution}.\hskip 1em plus 0.5em minus 0.4em\relax
  University of California Press, 1956, vol.~3.

\bibitem{albeladi2023time}
K.~Albeladi, B.~Zafar, and A.~Mueen, ``Time series forecasting using {LSTM} and
  {ARIMA},'' \emph{International Journal of Advanced Computer Science and
  Applications}, vol.~14, no.~1, pp. 313--320, 2023.

\bibitem{oreshkin2019n}
B.~N. Oreshkin, D.~Carpov, N.~Chapados, and Y.~Bengio, ``N-beats: Neural basis
  expansion analysis for interpretable time series forecasting,'' in
  \emph{ICLR}, New Orleans, LA, May 2019.

\bibitem{zerveas2021transformer}
G.~Zerveas, S.~Jayaraman, D.~Patel, A.~Bhamidipaty, and C.~Eickhoff, ``A
  transformer-based framework for multivariate time series representation
  learning,'' in \emph{Proceedings of SIGKDD Conference}, Virtual, Aug. 2021,
  pp. 2114--2124.

\bibitem{zhou2021informer}
H.~Zhou, S.~Zhang, J.~Peng, S.~Zhang, J.~Li, H.~Xiong, and W.~Zhang,
  ``Informer: Beyond efficient transformer for long sequence time-series
  forecasting,'' in \emph{Proceedings of AAAI Conference}, vol.~35, no.~12,
  Virtual, Feb. 2021, pp. 11\,106--11\,115.

\bibitem{lei2023stability}
Y.~Lei, ``Stability and generalization of stochastic optimization with
  nonconvex and nonsmooth problems,'' in \emph{Annual Conference on Learning
  Theory}, Bangalore, India, Jul. 2023, pp. 191--227.

\end{thebibliography}
\begin{IEEEbiography}[{\includegraphics[width=1.1in,height=1.3in,clip,keepaspectratio]{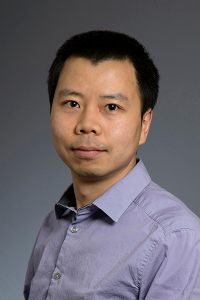}}]{Yong Xiao} (Senior Member, IEEE) 
is a professor in the School of Electronic Information and Communications at the Huazhong University of Science and Technology (HUST), Wuhan, China. He is also with Peng Cheng Laboratory, Shenzhen, China, and Pazhou Laboratory (Huangpu), Guangzhou, China. He is the associate group leader of the network intelligence group of IMT-2030 (6G promoting group). 
His research interests include AI/ML, game theory, distributed optimization, and their applications in agentic AI networking and semantic communications.
\end{IEEEbiography}
\vspace{-1cm}
\begin{IEEEbiography}[{\includegraphics[width=1.1in,height=1.3in,clip,keepaspectratio]{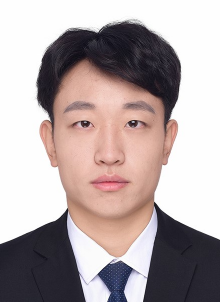}}]{Xubo Li} (Student Member, IEEE) received his B.S. degree in biomedical engineering from Huazhong University of Science and Technology, Wuhan, China in 2023. He is currently pursuing his Ph.D. in the School of Electronic Information and Communications at Huazhong University of Science and Technology, Wuhan, China. His research interest includes machine learning, network intelligence, and next-generation wireless communication technology.
\end{IEEEbiography}
\vspace{-1cm}
\begin{IEEEbiography}[{\includegraphics[width=1.1in,height=1.3in,clip,keepaspectratio]{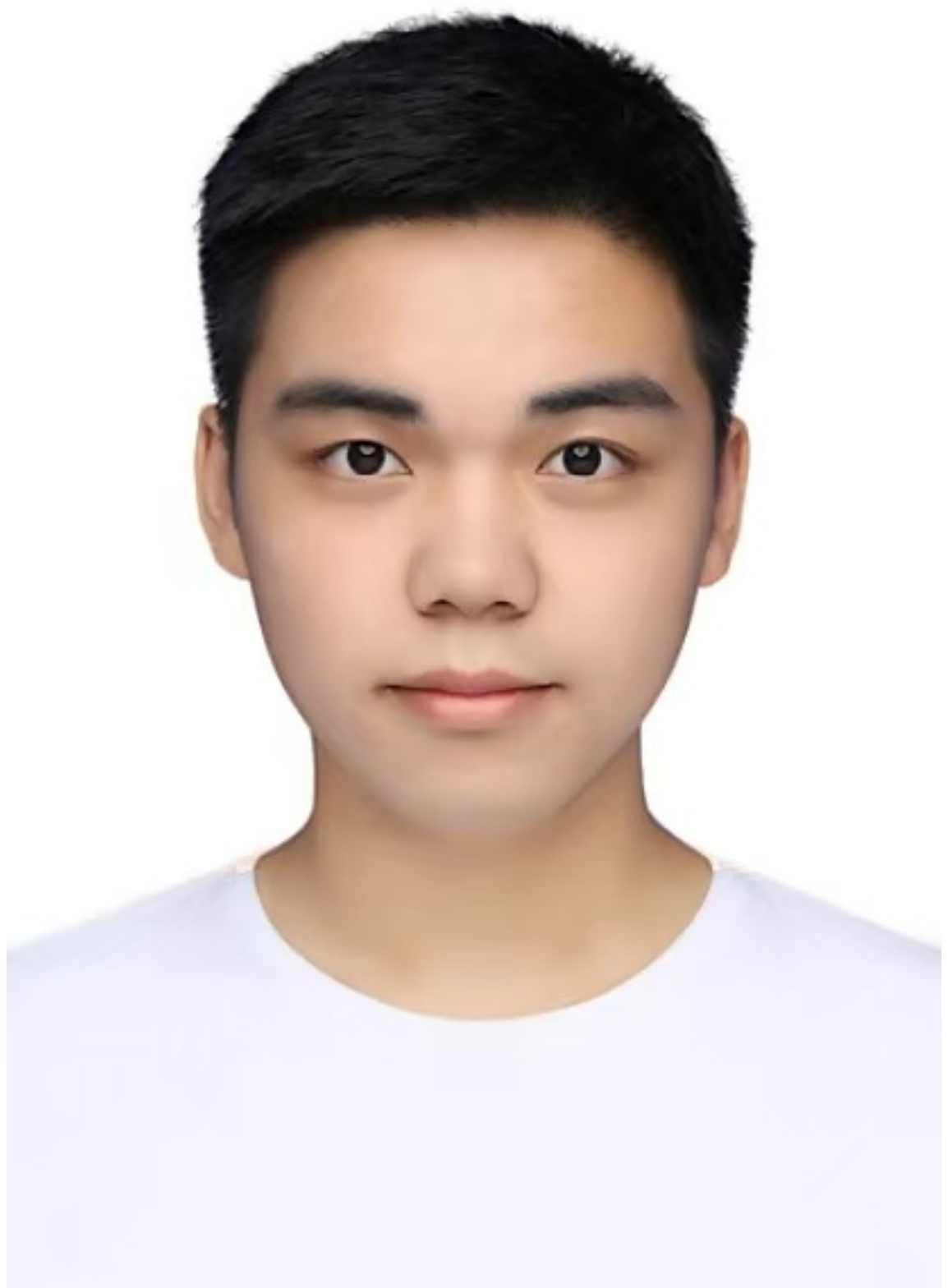}}]{Haoran Zhou} received his B.S. degree in electrical engineering from China University of Geosciences, Wuhan, China, in 2024. He is currently pursuing his Ph.D. in the School of Electronic Information and Communications at Huazhong University of Science and Technology, Wuhan, China. His research interest includes machine learning, network intelligence, and multi-agent communication.
\end{IEEEbiography}
\begin{IEEEbiography}[{\includegraphics[width=1.1in,height=1.3in,clip,keepaspectratio]{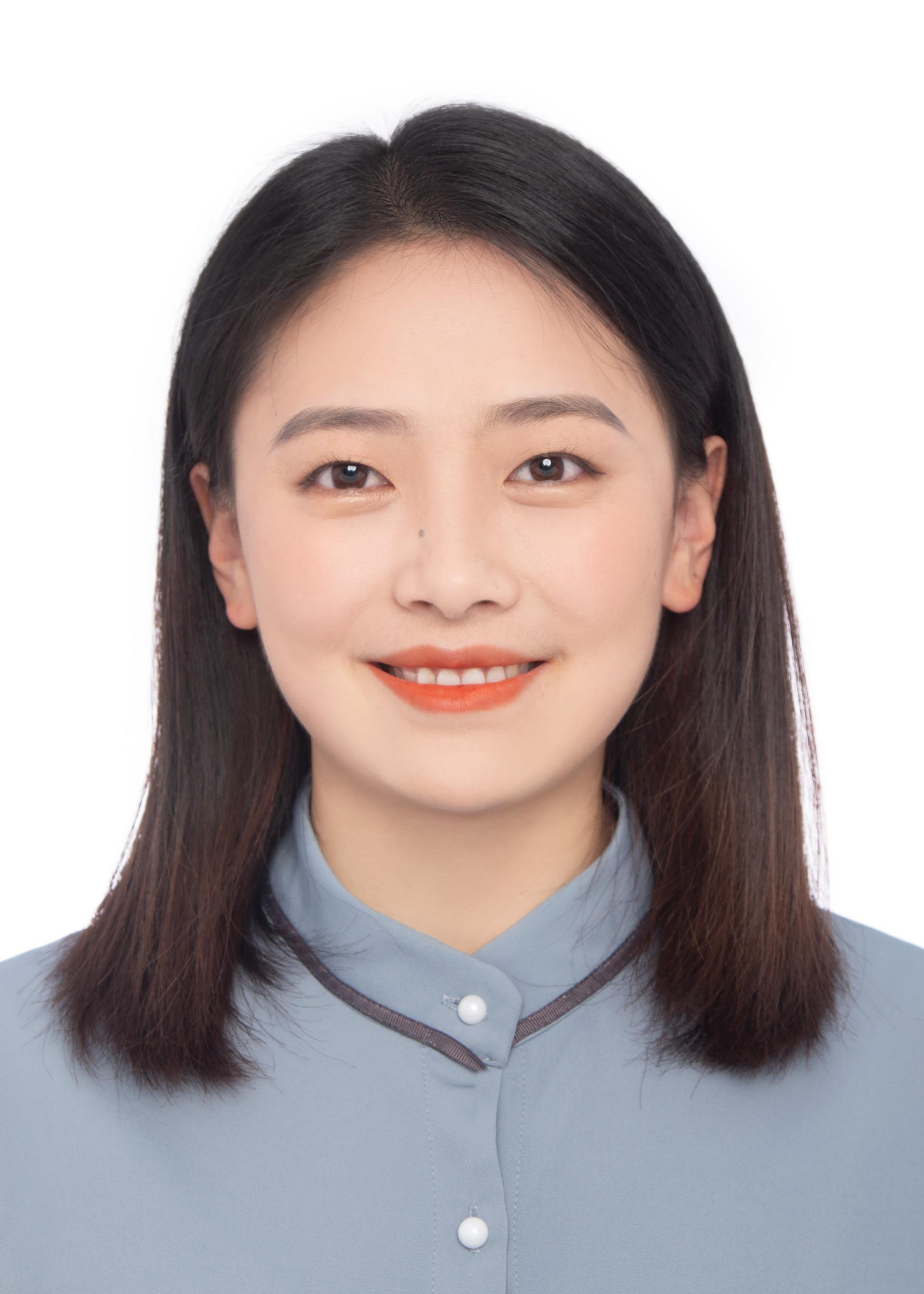}}]{Yingyu Li} (Member, IEEE) 
is an Associate Professor at the School of Mechanical Engineering and Electronic Information, China University of Geosciences (Wuhan). Her research interests include machine learning and artificial intelligence for next-generation wireless networks, federated edge intelligence, green/low-carbon communication networks, distributed optimization, semantic communications, and intelligent Internet of Things.
\end{IEEEbiography}
\begin{IEEEbiography}[{\includegraphics[width=1.1in,height=1.3in,clip,keepaspectratio]{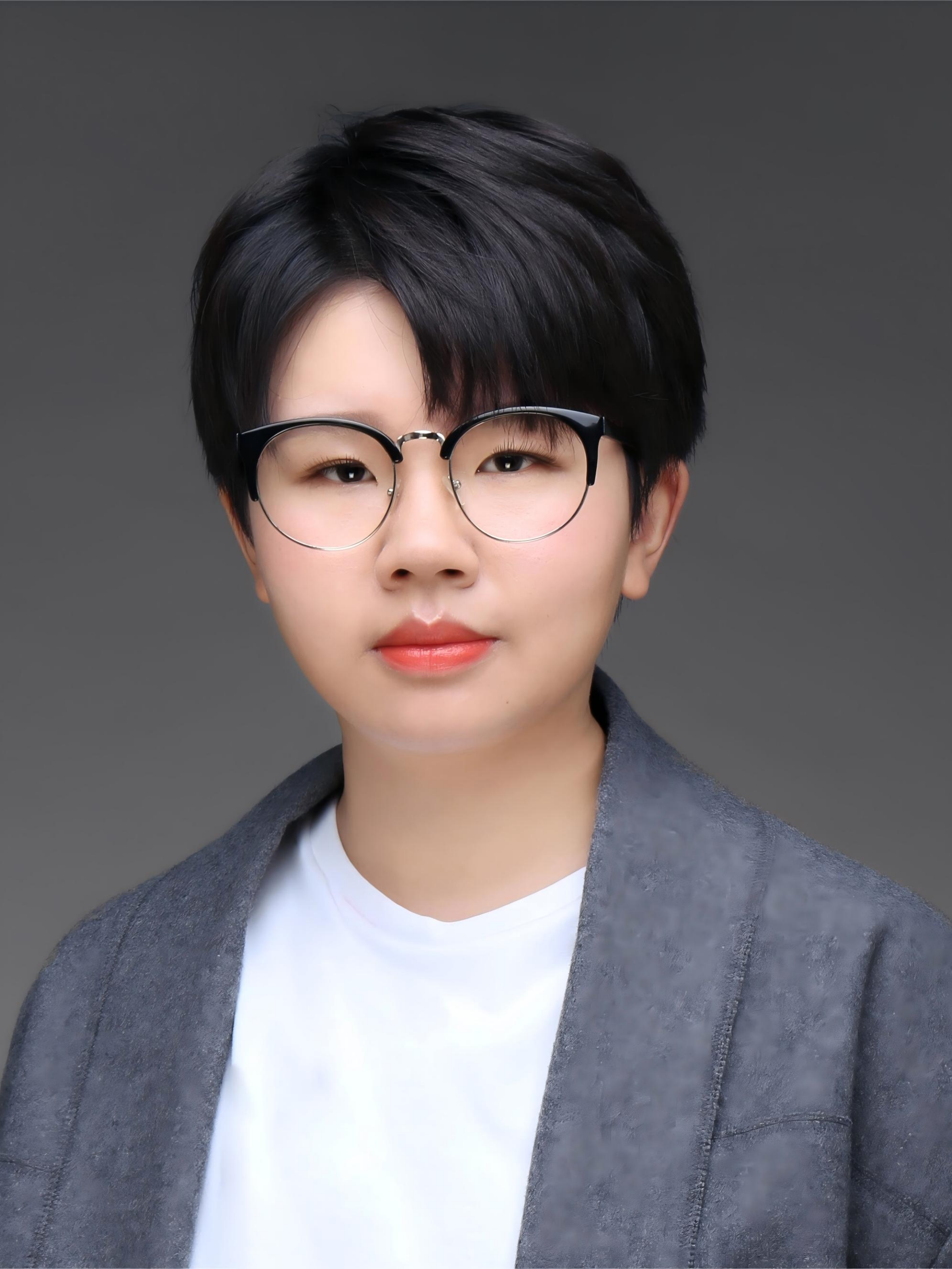}}]{Yayu Gao} (Member, IEEE) is an Associate Professor in the School of Electronic Information and Communications at the Huazhong University of Science and Technology (HUST), Wuhan, China. 
Her research interests include modeling and optimization of next-generation mobile communication networks, agentic AI networking, edge intelligence, and artificial intelligence for networking.
\end{IEEEbiography}
\begin{IEEEbiography}[{\includegraphics[width=1.1in,height=1.3in, clip,keepaspectratio]{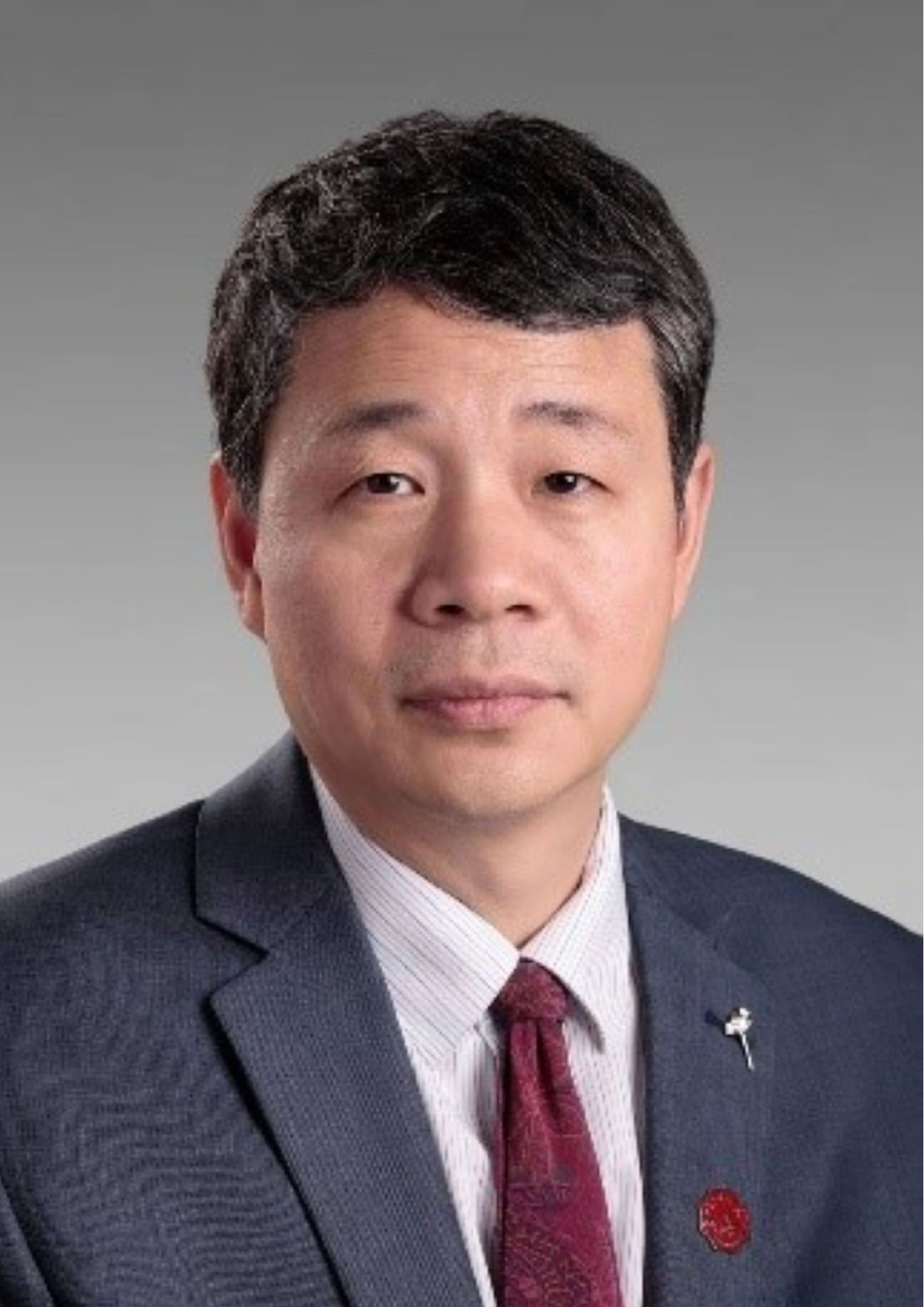}}]{Guangming Shi} (Fellow, IEEE) 
is the Vice Dean of Peng Cheng Laboratory and a Professor with the School of Artificial Intelligence, Xidian University. He is an IEEE Fellow, the chair of IEEE CASS Xi’an Chapter, a senior member of ACM and CCF, Fellow of the Chinese Institute of Electronics, and Fellow of IET. 
His research interests include Artificial Intelligence, Semantic Communications, and Human-Computer Interaction.
\end{IEEEbiography}
\begin{IEEEbiography}[{\includegraphics[width=1.1in,height=1.3in,clip,keepaspectratio]{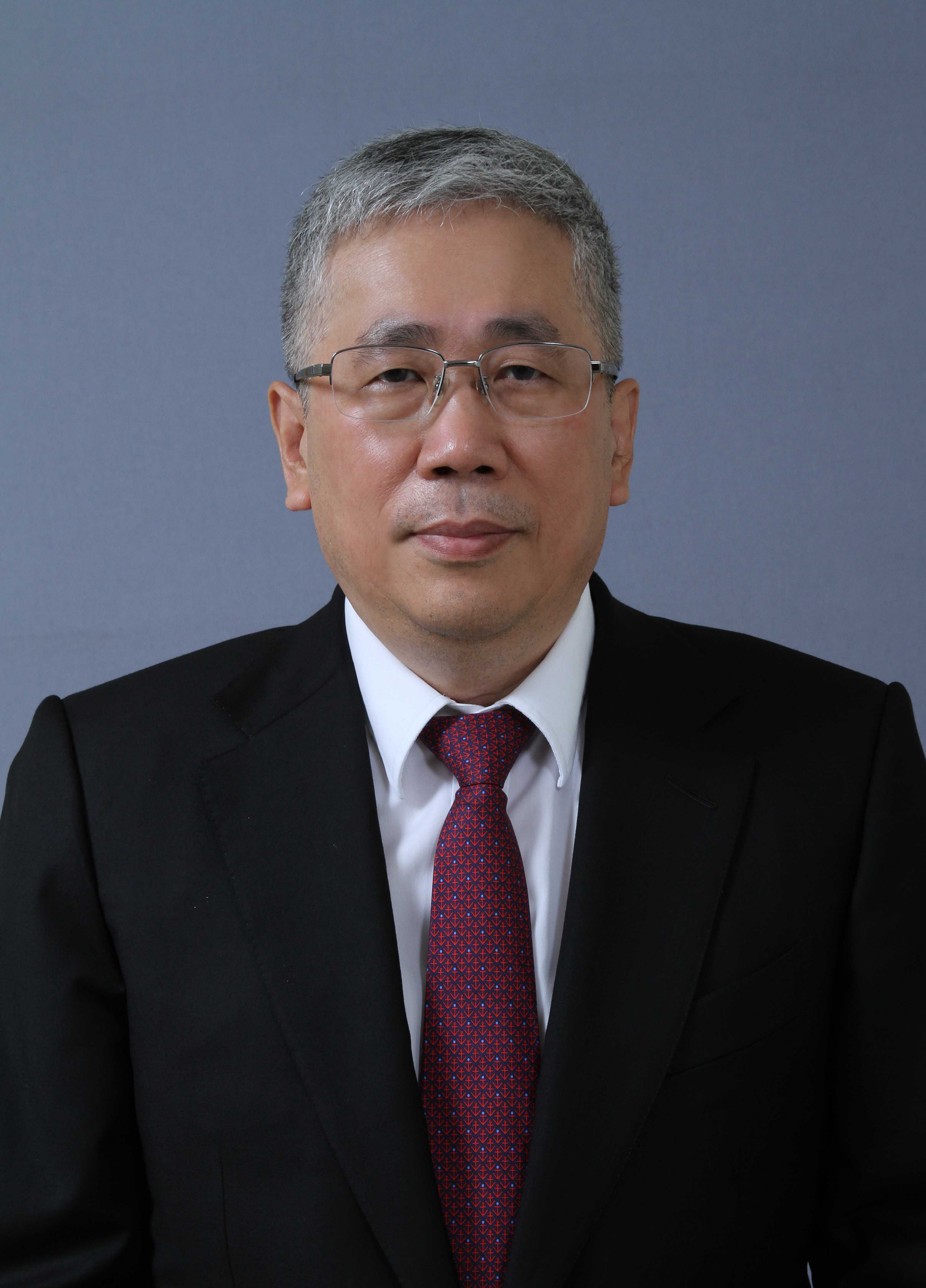}}]{Ping Zhang} (Fellow, IEEE) is a professor in the School of Information and Communication Engineering at the Beijing University of Posts and Telecommunications. He is a member of the National Academy of Engineering of China. He is currently the director of the State Key Laboratory of Networking and Switching Technology, a member of IMT-2020 (5G) Experts Panel, and a member of the Experts Panel for China's 6G development. 
His research is in the broad area of wireless communications with emphasis on novel coding design and model-driven approaches for semantic communications. 
\end{IEEEbiography}
\begin{IEEEbiography}[{\includegraphics[width=1.1in,height=1.3in,clip,keepaspectratio]{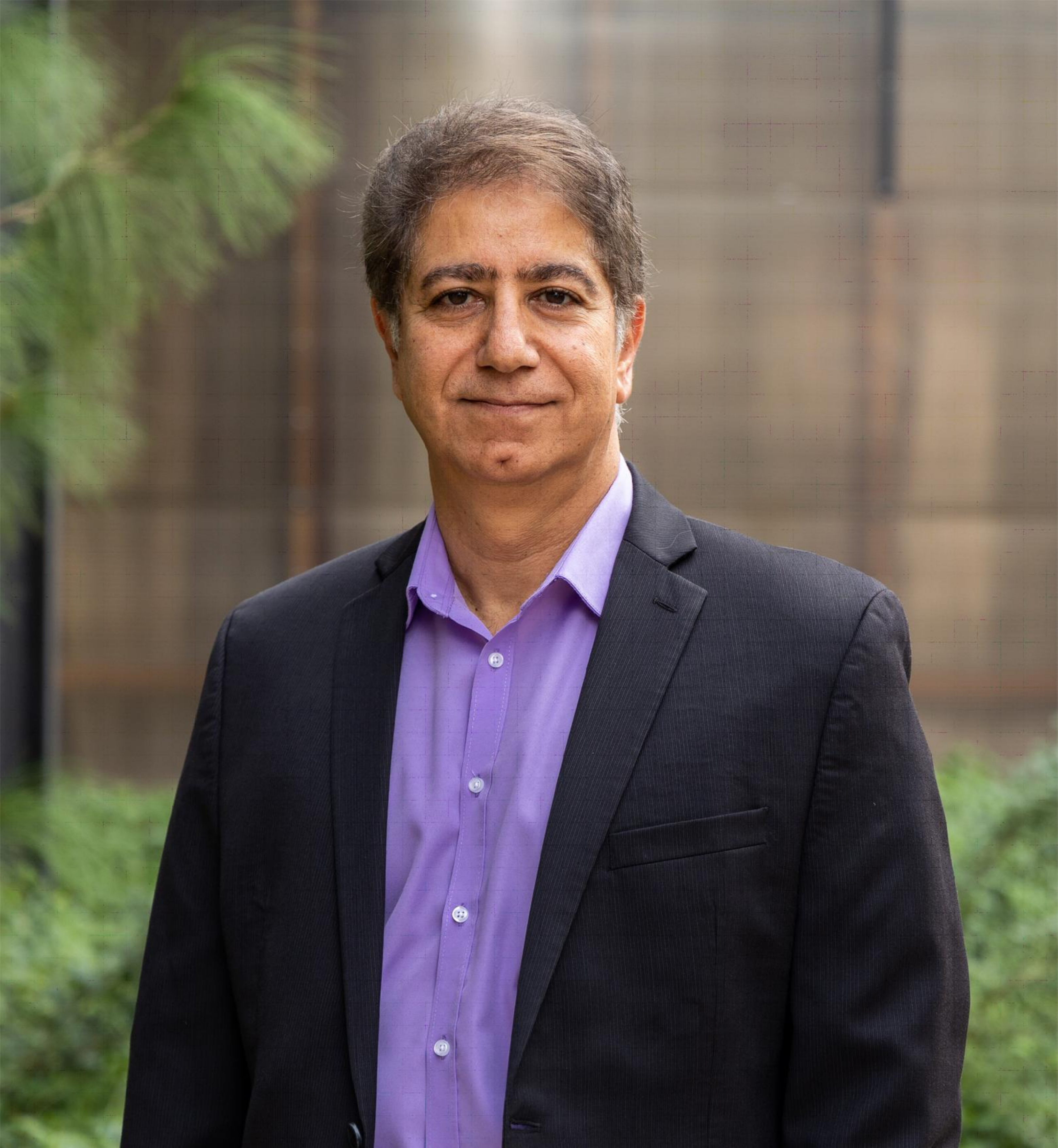}}]{Marwan Krunz}(Fellow, IEEE) is a Regents Professor in ECE at the University of Arizona (UA) and the Edward \& Maria Keonjian Endowed Chair in Electrical and Computer Engineering. He also holds a joint appointment as a Professor of Computer Science and is a member of the UA Cancer Center. He is the Deputy Center Director and Site Director of WISPER, an NSF/industry-funded consortium 
that aims to provide solutions for secure and AI-enabled NextGen wireless systems. Previously, Dr. Krunz directed two NSF/industry centers related to wireless technologies. Dr. Krunz's research is in the fields of wireless communications, networking, and security, with recent focus on applying AI and machine learning techniques for protocol adaptation, resource management, and signal intelligence. 
\end{IEEEbiography}

\newpage

\appendices
\section{Proofs of Theorem \ref{Theorem_UpperBoundOError_without CA} and Corollary \ref{Theorem_UpperBoundOError}}

\subsection{Auxiliary Lemmas}
\begin{lemma}\label{Opt_Lemma1}
    Suppose Assumption \ref{smoothness} holds. Consider the sequence $\{ \bOmega_{m, t} \}_{t=1}^{T}$ generated by static-weighting-based solution. Define
    \begin{flalign}
        &A_1 = \frac{1}{T} \mathbb{E} [ \sum_{i\in\{a, p, n\}} \gamma^i_{m, 0} ( \cL^i_m (\bOmega_{m, 0}, \cD^i) - \cL^i_m (\bOmega_{m, T}, \cD^i) ) ], \nonumber\\
        &A_2 = \frac{1}{T} \sum_{t=0}^{T-1} \mathbb{E} [ \| \sum_{i\in\{a, p, n\}} \gamma^i_{m, t} \nabla \ell^i_m (\bOmega_{m, t}, d^i_t) \|^2 ]. \nonumber
    \end{flalign}
    Then, it holds that
    \begin{eqnarray}
        \frac{1}{T} \sum_{t=0}^{T-1} \mathbb{E} [ \| \!\!\!\!\! \sum_{i\in\{a, p, n\}} \!\!\!\!\! \gamma^i_m \nabla \cL^i_m (\bOmega_{m, t}, \cD^i) \|^2 ] \le \frac{1}{\beta} A_1 + \frac{\beta \mu_g}{2} A_2. \nonumber
    \end{eqnarray}
\end{lemma}

\begin{IEEEproof}
    We first establish that, for any given static weight $\gamma^i_{m, 0}$, there exists an upper bound for the gap between the global losses in two consecutive coordination rounds. By leveraging the $\mu_g$-smoothness of $\ell_m^i (\cdot)$, we have
    \begin{flalign}
        & \sum_{i\in\{a, p, n\}} \gamma^i_{m, 0} (\cL^i_m (\bOmega_{m, t+1}, \cD^i) - \cL^i_m (\bOmega_{m, t}, \cD^i)) \nonumber\\
        \le& \langle \sum_{i\in\{a, p, n\}}\!\!\! \gamma^i_{m, 0} \nabla \cL^i_m (\bOmega_{m, t}, \cD^i),  \bOmega_{m, t+1} - \bOmega_{m, t}\rangle \nonumber\\
        & + \frac{\mu_g}{2}\| \bOmega_{m, t+1} - \bOmega_{m, t} \|^2 \nonumber\\
        =& - \beta_t \langle \!\!\!\!\!\sum_{i\in\{a, p, n\}}\!\!\!\!\!\! \gamma^i_{m, 0} \nabla \cL^i_m (\bOmega_{m, t}, \cD^i), \!\!\!\!\!\! \sum_{i\in\{a, p, n\}}\!\!\!\!\!\! \gamma^i_{m, t} \nabla \ell^i_m (\bOmega_{m, t}, d^i_t) \rangle \nonumber \\
        & + \frac{\mu_g}{2} \beta_t^2 \| \!\!\! \sum_{i\in\{a, p, n\}}\!\!\! \gamma^i_{m, t} \nabla \ell^i_m (\bOmega_{m, t}, d^i_t) \|^2. \label{lemma1_eq1}
    \end{flalign}

    Note that for static-weighting scheme, the weight $\gamma^i_{m, t}$ for each agent $i$ remain constant across coordination rounds, i.e., $\gamma^i_{m, t} = \gamma^i_{m, 0}$ for all $t$. Taking expectation over the random sample $d^i_t$ on both sides of the above inequality, and setting the step size $\beta_t = \beta \ge 0$, we have
    \begin{flalign}
        & \mathbb{E} [\sum_{i\in\{a, p, n\}} \gamma^i_{m, 0} (\cL^i_m (\bOmega_{m, t+1}, \cD^i) - \cL^i_m (\bOmega_{m, t}, \cD^i))] \nonumber\\
        \le& - \beta \mathbb{E} \| \sum_{i\in\{a, p, n\}}\gamma^i_{m, t} \nabla \cL^i_m (\bOmega_{m, t}, \cD^i) \|^2 \nonumber\\
        & + \frac{\mu_g}{2} \beta^2 \mathbb{E} \| \sum_{i\in\{a, p, n\}} \gamma^i_{m, t} \nabla \ell^i_m (\bOmega_{m, t}, d^i_t) \|^2.
    \end{flalign}

    Summing the the above inequality across all coordination rounds $t=0$ to $T-1$ and rearranging terms, we obtain
    \begin{flalign}
        &\frac{1}{T} \sum_{t=0}^{T-1} \mathbb{E}[ \| \sum_{i\in\{a, p, n\}} \gamma^i_m \nabla \cL^i_m (\bOmega_{m, t}, \cD^i) \|^2 ] \nonumber\\
        \le& \frac{1}{\beta T} \mathbb{E} [ \! \sum_{i\in\{a, p, n\}} \! \gamma^i_m ( \cL^i_m (\bOmega_{m, 0}, \cD^i) - \cL^i_m (\bOmega_{m, T}, \cD^i) )] \nonumber\\
        & + \frac{\beta \mu_g }{2} \mathbb{E} [\| \sum_{i\in\{a, p, n\}} \gamma^i_m \nabla \ell^i_m (\bOmega_{m, t}, d^i_t) \|^2].
    \end{flalign}
    This concludes the proof.
\end{IEEEproof}

\begin{lemma}[Lemma 19 \cite{chen2023three}]\label{Opt_Lemma2}
    Suppose Assumption \ref{smoothness} holds. Consider the sequence $\{ \bOmega_{m, t} \}_{t=1}^{T}$ generated by dynamic-weighting scheme. Further define
    \begin{flalign}
        &A_3 =  \nonumber\\
        &\frac{1}{T} \!\sum_{t=0}^{T-1}\mathbb{E} [ \| \!\!\!\!\!\sum_{i\in\{a, p, n\}}\!\!\!\!\!\! \gamma^i_{m, t} \nabla \ell^i_m (\bOmega_{m, t}, d^i_{t(1)})^{\top}\! \nabla \ell^i_m (\bOmega_{m, t}, d^i_{t(2)}) \|^2 ], \nonumber\\
        &A_4 = \nonumber\\
        &\frac{1}{T} \sum_{t=0}^{T-1}\mathbb{E} \big[ \| \!\!\!\!\!\sum_{i\in\{a, p, n\}}\!\!\!\!\! \gamma^i_{m, t} \nabla \ell^i_m (\bOmega_{m, t}, d^i_{t(1)})^{\top} \nabla \ell^i_m (\bOmega_{m, t}, d^i_{t(2)}) \|\nonumber\\
        &\ \ \ \ \ \ \ \ \ \ \ \ \| \!\!\!\!\!\sum_{i\in\{a, p, n\}}\!\!\!\!\! \gamma^i_{m, 0} \nabla \cL^i_{m} (\bOmega_{m, t}, \cD^i)^{\top} \nabla \cL^i_m (\bOmega_{m, t}, \cD^i) \| \big], \nonumber
    \end{flalign}
    Then, it holds that
    \begin{flalign}
        &\blu{\frac{1}{T} \sum_{t=0}^{T-1} \mathbb{E}_{\cA}[ \| \sum_{i\in\{a, p, n\}} \gamma^i_{m, t} \ell^i_m (\bOmega_{m, t}, \cD^i) \|^2 ]} \nonumber\\
        \le \ & \frac{1}{\beta} A_1 + \frac{\beta \mu_g}{2} A_2 + \frac{\eta}{2} A_3 + \eta A_4.
    \end{flalign}
\end{lemma}

\subsection{Proof of Theorem \ref{Theorem_UpperBoundOError_without CA}}\label{proof of O-error without CA}
According to Lemma \ref{Opt_Lemma1}, under Assumption \ref{smoothness}-\ref{initial model}, we have
\begin{flalign}
    &\blu{\frac{1}{T} \sum_{t=0}^{T-1} \mathbb{E} [ \| \sum_{i\in\{a, p, n\}} \gamma^i_m \nabla \cL^i_m (\bOmega_{m, t}, \cD^i) \|^2 ]} \nonumber\\
    \le \ &  \frac{1}{\beta} A_1 + \frac{\beta \mu_g}{2} A_2 \le \frac{c_I}{\beta T} + \frac{\beta \mu_g \mu_l^2}{2},
\end{flalign}
where the last inequality follows from the Lipschitz continuity of $\ell_m^i (\cdot)$ in Assumption \ref{continuous}, which ensures $A_2 \leq \mu_l^2$.

Then, by the Jensen's inequality and the convexity of the square function, as well as the sub-additivity of the square root function, it holds that
\begin{eqnarray}
    && \blu{\frac{1}{T} \sum_{t=0}^{T-1} \mathbb{E} [ \| \sum_{i\in\{a, p, n\}} \gamma^i_m \nabla \cL^i_m (\bOmega_{m, t}, \cD^i) \| ]} \nonumber\\
    &\blu{\le}& \blu{\left( \frac{1}{T} \sum_{t=0}^{T-1} \mathbb{E}[ \| \sum_{i\in\{a, p, n\}} \gamma^i_m \nabla \cL^i_m (\bOmega_{m, t}, \cD^i) \|^2 ] \right)^{\frac{1}{2}}} \nonumber\\
    &\le& \sqrt{\frac{c_I}{\beta T}} + \sqrt{\frac{\beta \mu_g \mu_l^2}{2}} = \mathcal{O}(\beta^{-\frac{1}{2}} T^{-\frac{1}{2}} + \beta^{\frac{1}{2}}).
\end{eqnarray}
This completes the proof.

\subsection{Proof of Corollary \ref{Theorem_UpperBoundOError}}\label{proof of O-error}
The proof follows a similar structure to that of Theorem \ref{Theorem_UpperBoundOError_without CA}.
According to Lemma \ref{Opt_Lemma2}, under Assumption \ref{smoothness}-\ref{initial model}, it holds that
\begin{eqnarray}
    &&\blu{\frac{1}{T} \sum_{t=0}^{T-1} \mathbb{E}[ \| \sum_{i\in\{a, p, n\}} \gamma^i_{m, t} \nabla \cL^i_m (\bOmega_{m, t}, \cD^i) \|^2 ]} \nonumber\\
    &\le& \frac{1}{\beta} A_1 + \frac{\beta \mu_g}{2} A_2 + \frac{\eta}{2} A_3 + \eta A_4 \nonumber\\
    &\le& \frac{c_I}{\beta T} + \frac{9}{2} \eta \mu_l^4 + \frac{1}{2} \beta \mu_g \mu_l^2,
\end{eqnarray}
where the last inequality comes from the fact that 1) $A_3 \le 3 \mu_l^4$ and 2) $A_4 \le 3 \mu_l^4$, which results from the smoothness in Assumption \ref{smoothness}.

Then, following the same reasoning as in the proof of Theorem \ref{Theorem_UpperBoundOError_without CA},
we apply Jensen's inequality and the subadditivity of the square root to derive:
\begin{eqnarray}
    &&\blu{\frac{1}{T} \sum_{t=0}^{T-1} \mathbb{E} [ \| \sum_{i\in\{a, p, n\}} \gamma^i_{m, t} \nabla \cL^i_m (\bOmega_{m, t}, \cD^i) \| ]} \nonumber\\
    &\blu{\le}& \blu{\left( \frac{1}{T} \sum_{t=0}^{T-1} \mathbb{E} [ \| \sum_{i\in\{a, p, n\}} \gamma^i_{m, t} \nabla \cL^i_m (\bOmega_{m, t}, \cD^i) \|^2 ] \right)^{\frac{1}{2}}} \nonumber\\
    &\le& \sqrt{\frac{c_I}{\beta T}} + 3\sqrt{\frac{\eta \mu_l^4}{2}} + \sqrt{\frac{\beta \mu_g \mu_l^2}{2}} \nonumber\\
    &=& \mathcal{O}(\beta^{-\frac{1}{2}} T^{-\frac{1}{2}} + \eta^{\frac{1}{2}} + \beta^{\frac{1}{2}}).
\end{eqnarray}
This concludes the proof.

\section{Proofs of Theorem \ref{Theorem_UpperBoundGError_without CA} and Corollary \ref{Theorem_UpperBoundGError}}

\subsection{Auxiliary Lemmas}\label{connection between stability and generalization}
\begin{lemma}[Theorem 16 \cite{lei2023stability}]\label{lemma 1}
    Let $\cD = \bigcup_{i\in \{a, p, n\}} {\cD^i}$ and $\tilde{\cD} = \bigcup_{i\in \{a, p, n\}} \tilde{\cD^i}$ be two identical training datasets except that only the $k$th data sample $d^j_k$ in dataset $\cD^j$ is different. Let $\cI_{\cD^i}(\cA_m) = \{ s^i_{m,t} \}_{t=1}^T$ represent the random index sequence of samples in dataset $\cD^i$ to compute stochastic gradients during each training round. If for any data sample $d^i \in \cD^i$, 
    \begin{equation}
        \sup_{d^i} \mathbb{E}[ \| \!\!\! \sum_{i\in\{a, p, n\}} \!\!\! \gamma^i_{m} \nabla \ell^i_m (\cA_m(\cD), d^i) \|^2 | k \in \cI_{\cD^j}(\cA_m) ] \le G^2. \nonumber
    \end{equation}
    Then, we have
    \begin{flalign}
        &\sup\limits_{d^i} \mathbb{E}[ \| \!\!\!\! \sum\limits_{i\in\{a, p, n\}} \!\!\!\! \gamma^i_m (\nabla \ell^i_m (\cA_m(\cD), d^i) - \nabla \ell^i_m (\cA_m(\tilde{\cD}), d^i)) \|^2 ] \nonumber\\
        &\le 4 G^2 \cdot \mathbb{P}(k \in \cI_{\cD^j}(\cA_m)).
    \end{flalign}
\end{lemma}

\begin{lemma}[Theorem 6 \cite{lei2023stability}]\label{connection between generalization and stability}
    Suppose for any data sample $d^i \in \cD^i$, the loss function $\ell^i_m (\bOmega_m, d^i)$ is differentiable. If a randomized algorithm $\cA$ is uniformly stable with $\epsilon$, then
    \begin{flalign}
        &\mathbb{E}_{\cA, \cD}[ \| \sum_{i\in\{a, p, n\}} \gamma^i_m (\nabla \ell^i_m (\cA_m(\cD), \cD^i) - \nabla {\tilde \ell}^{i}_m (\cA_m(\tilde{\cD})) \|^2 ] \nonumber\\
        &\le 4 \epsilon + \sqrt{\frac{V}{D}},
    \end{flalign}
    where 
    \begin{equation}
        V = \mathbb{E}[ \| \!\!\!\!\!\! \sum_{i\in\{a, p, n\}} \!\!\!\!\!\! \gamma^i_m (\nabla \ell^i_m (\cA_m(\cD), d^i) \!-\! \mathbb{E}[\nabla \ell^i_m (\cA_m(\cD), d^i)]) \|^2 ] \nonumber
    \end{equation}
    denotes the variance of gradients.
\end{lemma}

\subsection{Proof of Theorem \ref{Theorem_UpperBoundGError_without CA}}\label{proof of G-error without CA}
Based on Lemma \ref{lemma 1}, the uniform stability parameter in Definition \ref{def_of_stability} can be bounded by
\begin{equation}\label{Thm1_1}
    \epsilon^2 \le 4 G^2 \cdot \mathbb{P}(k \in \cI_{\cD^j}(\cA_m)).
\end{equation}

Let $s^i_t$ be the index of sample in dataset $\cD^i$ selected by algorithm $\cA$ at $t$th coordination, $\cI_{\cD^i, t}(\cA_m)$ be the indices of samples in dataset $\cD^i$ selected by algorithm $\cA$ up to the $t$-th iteration. Then, 
\begin{eqnarray}\label{Thm1_2}
    \mathbb{P}(j \in \cI_{\cD^i, t}(\cA_m)) &=& \mathbb{P}(\bigcup_{\tau=1}^t \{ s_\tau^i = j \} ) \nonumber\\
    &\le& \sum_{\tau =1}^t \mathbb{P}\{ s_\tau^i = j \} = \frac{t}{D}.
\end{eqnarray}

Substituting (\ref{Thm1_2}) into (\ref{Thm1_1}), the uniform stability parameter of the $t$-th iteration is upper bounded by 
\begin{equation}\label{}
    \epsilon^2 \le \frac{4 G^2 t}{D} \le \frac{4 G^2 T}{D}.
\end{equation}

Then, based on Lemma \ref{connection between generalization and stability}, the G-error is upper bounded by
\begin{eqnarray}
    \mathbb{E}_{\cA, \cD} [\cE_G]
    &\le& 4 \epsilon + \sqrt{\frac{V}{D}} \nonumber\\ 
    &\le& 8G \sqrt{\frac{T}{D}} + \sqrt{\frac{V}{D}} = \mathcal{O}(T^{\frac{1}{2}} D^{-\frac{1}{2}}).
\end{eqnarray}
This concludes the proof.

\subsection{Proof of Corollary \ref{Theorem_UpperBoundGError}}\label{proof of G-error}
The proof of Corollary \ref{Theorem_UpperBoundGError_without CA} reveals that the algorithm's stochasticity originates solely from the random sampling process at each iteration. As the conflict-resolving mechanism preserves this sampling distribution, we conclude that the upper bound on the G-error remains unchanged. Refer to Appendix \ref{proof of G-error without CA} for a detailed proof.

\section{Proofs of Theorem \ref{Theorem_UpperBoundCError_without CA} and Theorem \ref{Theorem_UpperBoundCError}} 

\subsection{Auxiliary Lemmas}
\begin{lemma}[Lemma 18 \cite{chen2023three}] \label{conflict_lemma1}
    Suppose Assumption \ref{smoothness} holds. Consider the sequence $\{ \bOmega_{m, t} \}_{t=1}^T$ and $\{ \gamma_{m, t} \}_{t=1}^T$ generated by dynamic weighting algorithm with step size $\beta_t = \beta \ge 0, \eta_t = \eta \ge 0$. Follow the definition of $A_3$ in Lemma \ref{Opt_Lemma2} and define 
    \begin{flalign}
        &A_5 \!=\! \frac{1}{T} \sum_{t=0}^{T-1} \!\mathbb{E} \{ \| \!\!\!\!\!\! \sum_{i\in\{a, p, n\}} \!\!\!\!\![\nabla \cL^i_m (\bOmega_{m, t}, \cD^i) \!+\! \nabla \cL^i_m (\bOmega_{m, t+1}, \cD^i)] \| \nonumber\\
        &\quad\quad\quad\quad\quad\quad\quad \| \sum_{i\in\{a, p, n\}} \gamma^i_{m, t+1} \nabla \ell^i_m (\bOmega_{m, t}, d^i)  \| \}. \nonumber
    \end{flalign}
    Then it holds that
    \begin{eqnarray}
    &&\frac{1}{T} \sum_{t=0}^{T-1} \mathbb{E} [\| \sum_{i\in\{a, p, n\}} \gamma^i_{m, t} \nabla \cL^i_m (\bOmega_m, \cD^i) \|^2 \nonumber\\
    && \quad\quad\quad\quad\quad - \| \sum_{i\in\{a, p, n\}} \gamma^{i*}_{m, t} \nabla \cL^i_m (\bOmega_m, \cD^i) \|^2] \nonumber\\
    &\le& \rho + \frac{4}{\eta T}(1 + \sqrt{3} \rho^{-1} \beta T \mu_g A_5) + \eta A_3,
    \end{eqnarray}
    where $\rho$ is a strictly positive parameter and is used for analysis but not for algorithm update such that 
    \begin{flalign}
        &\bgamma^{*}_{m, t} = \langle \gamma^{a*}_{m, t}, \gamma^{p*}_{m, t}, \gamma^{n*}_{m, t} \rangle = \nonumber\\
        &\arg \min_{\bgamma} \ \frac{1}{2}\| \sum_{i\in\{a, p, n\}} \gamma^{i*}_{m, t} \nabla \cL^i_m (\bOmega_m, \cD^i) \|^2 + \frac{\rho}{2}\| \bgamma \|^2.
    \end{flalign}
\end{lemma}

\subsection{Proof of Theorem \ref{Theorem_UpperBoundCError_without CA}}\label{proof of C-error without CA}
We will present the upper and lower bounds of the conflict error under the static weighting algorithm, respectively. Let $\cE_C(\bOmega_{m, t}, \bgamma_{m, t})$ denote the conflict error associated with model parameters $\bOmega_{m, t}$ and weight vector $\bgamma_{m, t} = \langle \gamma^{a}_{m, t}, \gamma^{p}_{m, t}, \gamma^{n}_{m, t} \rangle $.

First, for the upper bound, it holds that
\begin{flalign}\label{Thm3_upper_bound}
    &\cE_C(\bOmega_{m, T}, \bgamma_{m, T}) \\
    = \ & \mathbb{E} [ \| \sum_{i\in\{a, p, n\}}(\gamma^i_{m, T}-\gamma^{i*}_{m, T}) \nabla \cL^i_m (\bOmega_{m, T}, \cD^i) \|^2] \le 4G^2, \nonumber
\end{flalign}
where the inequality comes from the Assumption \ref{Bounded_Gradient}.

Second, by defining $\tilde{i} = \arg \min_{i\in\{a, p, n\}} \mathbb{E} [\gamma_{m, t}^{i*} (\bOmega_{m, t})]$, we can prove the lower bound as follows. By the pigeonhole principle, we can easily have $\mathbb{E} [\gamma_{m, t}^{\tilde{i}} (\bOmega_{m,t})] \le \frac{1}{3}$ for any given $t$, then there exists $\gamma^{\tilde{i}}_{m, T} = 1$ such that
\begin{flalign}\label{Thm3_lower_bound}
    &\cE_C(\bOmega_{m, T}, \bgamma_{m, T}) \nonumber\\
    = \ & \mathbb{E} [ \| \sum_{i\in\{a, p, n\}}(\gamma^{\tilde{i}}_{m, T}-\gamma^{\tilde{i}*}_{m, T}) \nabla \cL^i_m (\bOmega_{m, T}, \cD^i) \|^2] \nonumber\\
    \ge \ & (1 - \frac{1}{3})^2 G^2 \ge \frac{4}{9} G^2.
\end{flalign}

Combining the upper and lower bounds in (\ref{Thm3_upper_bound}) and (\ref{Thm3_lower_bound}) yields the result.

\subsection{Proof of Theorem \ref{Theorem_UpperBoundCError}}\label{proof of C-error}

First, building on the result in Lemma \ref{conflict_lemma1}, we have
\begin{eqnarray}
    && \frac{1}{T} \sum_{t=0}^{T-1} \mathbb{E} [\| \sum_{i\in\{a, p, n\}}(\gamma^i_{m, t}-\gamma^{i*}_{m, t}) \nabla \cL^i_m (\bOmega_m, \cD^i) \|^2] \nonumber\\
    &\le& \frac{1}{T} \sum_{t=0}^{T-1} \mathbb{E} [\| \sum_{i\in\{a, p, n\}} \gamma^i_{m, t} \nabla \cL^i_m (\bOmega_m, \cD^i) \|^2 \nonumber\\
    && \quad\quad\quad\quad\quad - \| \sum_{i\in\{a, p, n\}} \gamma^{i*}_{m, t} \nabla \cL^i_m (\bOmega_m, \cD^i) \|^2] \nonumber\\
    &\le& \rho + \frac{4}{\eta T}(1 + \sqrt{3} \rho^{-1} \beta T \ell_{f} A_5) + \eta A_3,
\end{eqnarray}
where the first inequality comes from the triangle inequality.

Next, we derive the C-error of $t$th coordination as follows
\begin{flalign}
    &\cE_C(\bOmega_{m, t}, \bgamma_{m, t}) \nonumber\\
    = \ & \| \sum_{i\in\{a, p, n\}}(\gamma^i_{m, t}-\gamma^{i*}_{m, t}) \nabla \cL^i_m (\bOmega_{m, t}, \cD^i) \|^2 \nonumber\\
    = \ & \| \sum_{i\in\{a, p, n\}} \gamma^i_{m, t}\! \nabla \cL^i_m (\bOmega_{m, t}, \cD^i) \|^2 \nonumber\\
    &\ + \|  \sum_{i\in\{a, p, n\}} \gamma^{i*}_{m, t}\! \nabla \cL^i_m (\bOmega_{m, t}, \cD^i) \|^2 \nonumber\\
    &\ - 2  \langle \!\!\!\!\sum_{i\in\{a, p, n\}} \!\!\!\!\!\!\! \gamma^i_{m, t} \nabla \cL^i_m (\bOmega_{m, t}, \cD^i),\!\!\!\!\!\!\! \sum_{i\in\{a, p, n\}} \!\!\!\!\! \gamma^{i*}_{m, t} \nabla \cL^i_m (\bOmega_{m, t}, \cD^i) \rangle \nonumber\\
    \le \ & \| \sum_{i\in\{a, p, n\}} \gamma^i_{m, t} \nabla \cL^i_m (\bOmega_{m, t}, \cD^i) \|^2 \nonumber\\
    &\ - \| \sum_{i\in\{a, p, n\}} \gamma^{i*}_{m, t} \nabla \cL^i_m (\bOmega_{m, t}, \cD^i) \|^2, \nonumber
\end{flalign} 
where the inequality comes from the fact that
\begin{equation}
    \| \!\!\!\!\!\!\sum_{i\in\{a, p, n\}} \!\!\!\!\!\!\gamma^i_{m, t}\! \nabla \cL^i_m (\bOmega_{m, t}, \cD^i) \|^2 \ge \|\!\!\!\!\!\! \sum_{i\in\{a, p, n\}}\!\!\!\!\!\!\gamma^{i*}_{m, t}\! \nabla \cL^i_m (\bOmega_{m, t}, \cD^i) \|^2. \nonumber
\end{equation}


By averaging the results of each coordination and choosing $\rho = 2 (3 \beta \mu_g \mu_l^2/\eta)^{\frac{1}{2}}$, we have 
\begin{flalign}
    &\frac{1}{T} \sum_{t=0}^{T-1} \mathbb{E}[\cE_C(\bOmega_{m, t}, \bgamma_{m, t})] \nonumber\\
    \le \ & \frac{1}{T} \sum_{t=0}^{T-1} \mathbb{E} [\| \!\!\!\!\! \sum_{i\in\{a, p, n\}}\!\!\!\!\!(\gamma^i_{m, t}-\gamma^{i*}_{m, t}) \nabla \cL^i_m (\bOmega_{m, t}, \cD^i) \|^2] \nonumber\\
    \le \ & \rho + \frac{4}{\eta T}(1 + 6 \rho^{-1}  \beta T \mu_g \mu_l^2 ) + \eta \cdot 3\mu_l^4 \nonumber\\
    = \ & \frac{4}{\eta T} + 6\sqrt{3\mu_g \mu_l^2 \frac{\beta}{\eta}} + 3\eta \mu_l^4,
\end{flalign}
where the last inequality comes from that $A_3 \le 3 \mu_l^4$ and $A_5 \le 2 \sqrt{3} \mu_l^2$.
This completes the proof.

\end{document}